\documentclass[letterpaper]{article} % DO NOT CHANGE THIS
\usepackage{aaai2026}  % DO NOT CHANGE THIS
\usepackage{times}  % DO NOT CHANGE THIS
\usepackage{helvet}  % DO NOT CHANGE THIS
\usepackage{courier}  % DO NOT CHANGE THIS
\usepackage[hyphens]{url}  % DO NOT CHANGE THIS
\usepackage{graphicx} % DO NOT CHANGE THIS
\usepackage{natbib}  % DO NOT CHANGE THIS AND DO NOT ADD ANY OPTIONS TO IT
\usepackage{caption} % DO NOT CHANGE THIS AND DO NOT ADD ANY OPTIONS TO IT
\usepackage{amsmath}
\usepackage{array}
\usepackage{amssymb}
\usepackage{multirow}  
\usepackage{booktabs} 
\usepackage{algorithm}
\usepackage{algorithmic}

\usepackage{newfloat}
\usepackage{listings}
\DeclareCaptionStyle{ruled}{labelfont=normalfont,labelsep=colon,strut=off} % DO NOT CHANGE THIS
\floatstyle{ruled}
\newfloat{listing}{tb}{lst}{}
\floatname{listing}{Listing}
\title{Rethinking Rainy 3D Scene Reconstruction via Perspective Transforming and Brightness Tuning}
\author{
    Qianfeng Yang\equalcontrib\textsuperscript{\rm 1}, 
    Xiang Chen\equalcontrib\textsuperscript{\rm 2}, 
    Pengpeng Li\textsuperscript{\rm 2}, 
    Qiyuan Guan\textsuperscript{\rm 1}, 
    Guiyue Jin\textsuperscript{\rm 1}\thanks{Corresponding author.}, 
    Jiyu Jin\textsuperscript{\rm 1}\thanks{Corresponding author.}
}

\affiliations{
    \textsuperscript{\rm 1}Dalian Polytechnic University\\
    \textsuperscript{\rm 2}Nanjing University of Science and Technology\\
}
\usepackage{bibentry}
\begin{document}

\maketitle

\begin{abstract}
Rain degrades the visual quality of multi-view images, which are essential for 3D scene reconstruction, resulting in inaccurate and incomplete reconstruction results. Existing datasets often overlook two critical characteristics of real rainy 3D scenes: the viewpoint-dependent variation in the appearance of rain streaks caused by their projection onto 2D images, and the reduction in ambient brightness resulting from cloud coverage during rainfall. To improve data realism, we construct a new dataset named OmniRain3D that incorporates perspective heterogeneity and brightness dynamicity, enabling more faithful simulation of rain degradation in 3D scenes. 
Based on this dataset, we propose an end-to-end reconstruction framework named REVR-GSNet (\textbf{R}ain \textbf{E}limination and \textbf{V}isibility \textbf{R}ecovery for 3D \textbf{G}aussian \textbf{S}platting).
Specifically, REVR-GSNet integrates recursive brightness enhancement, Gaussian primitive optimization, and GS-guided rain elimination into a unified architecture through joint alternating optimization, achieving high-fidelity reconstruction of clean 3D scenes from rain-degraded inputs.
Extensive experiments show the effectiveness of our dataset and method. Our dataset and method provide a foundation for future research on multi-view image deraining and rainy 3D scene reconstruction.
\end{abstract}

\begin{links}
    \link{Code}{https://github.com/ncfjd/REVR-GSNet}
\end{links}

\section{Introduction}

Three-dimensional (3D) scene reconstruction recovers the spatial structure of real-world environments from image sequences and plays a critical role in vision-centric applications such as autonomous driving~\cite{lu2025dual,hu2024daldet} and robotics~\cite{liang2025diffusion}. However, adverse weather conditions, particularly rainy scenes, introduce visual degradations such as reduced visibility, which disrupt multi-view consistency and degrade reconstruction. Therefore, developing robust reconstruction techniques specifically designed for rainy scenarios is essential to ensure the reliability and precision of downstream tasks under challenging conditions.

\begin{figure*}[t]
\centering
\includegraphics[width=1.0\textwidth]{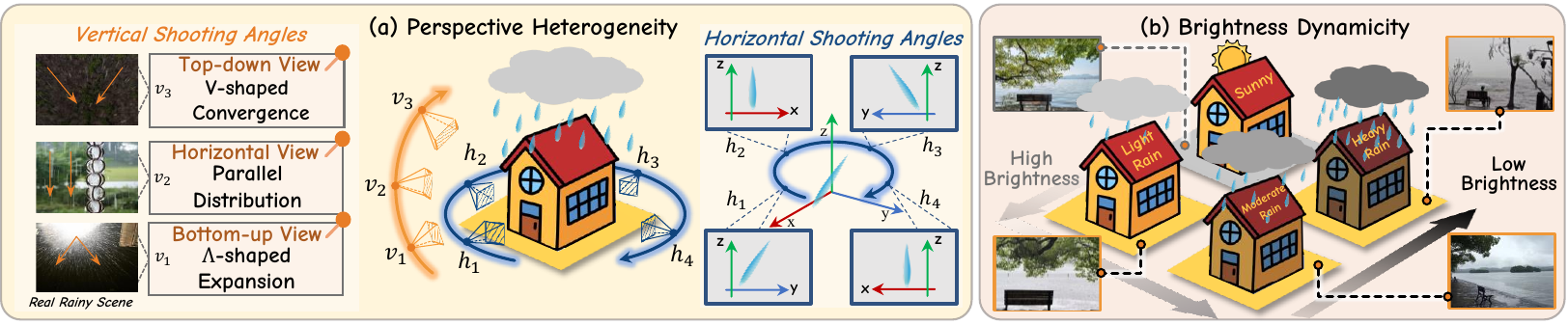}
\vspace{-4mm}
\caption{Illustration of key characteristics in real rainy scenes. (a) Perspective Heterogeneity: Rain streaks vary in appearance across both vertical and horizontal directions, as shown in real observations. (b) Brightness Dynamicity: An increase in rainfall is often accompanied by a decrease in ambient brightness, as confirmed by images ranging from sunny to heavy rain conditions.
}
\label{fig_motivation}
\vspace{-3mm}
\end{figure*}

In the real world, rain occurs within a 3D scene, and rain streaks therefore exhibit certain characteristics unique to 3D spatial structure, as illustrated in Figure~\ref{fig_motivation}. First, rain in a 3D scene is a volumetric phenomenon with depth, where rain streaks are distributed between the camera and the objects. As the viewpoint changes, the appearance of rain in the image also varies, such as changes in the angle and length of rain streaks~\cite{farber1978geometric}. We refer to this viewpoint-dependent variability as \textbf{Perspective Heterogeneity}. Second, in real-world rainy scenarios, rain clouds often reduce ambient brightness, and there is a correlation between rain and scene brightness~\cite{changbo2008real,tremblay2021rain}, a phenomenon we define as \textbf{Brightness Dynamicity}. These coupled effects introduce geometric distortions and texture artifacts, ultimately degrading image quality and affecting reconstruction accuracy.
Several prior works have attempted to construct 3D rainy scene datasets to support reconstruction under rainy conditions. HydroViews~\cite{liu2024deraings} generates rain-degraded images by linearly overlaying 2D rain masks onto background images captured from different viewpoints. However, this approach remains limited to two-dimensional degradation and lacks physical consistency with 3D rain behavior. RainyScape~\cite{lyu2024rainyscape} introduces a more realistic degradation by simulating rain effects in 3D space, which improves visual realism. Nevertheless, it neglects the impact of rainfall on ambient brightness, resulting in a significant domain gap from real rainy scenes.

In light of these limitations, we introduce a new data construction pipeline and build a dataset named OmniRain3D to address the challenges of 3D scene reconstruction under rainy conditions, as illustrated in Figure~\ref{fig_pipeline}. 
Specifically, we first extract multi-view camera poses from background images using COLMAP.
Blender then renders dynamic rain streaks from corresponding viewpoints based on these poses.
Finally, we adjust background brightness via an exponential decay model and overlay rain streaks under varying intensities to produce realistic rainy scenes.

%----------------------------------------------------------------------------
%----------------------------------------------------------------------------
Based on this dataset, developing a robust model for high-quality 3D reconstruction remains an open and valuable research direction.
Existing methods~\cite{li2024derainnerf, liu2024deraings} for 3D scene reconstruction under rainy conditions typically rely on pre-trained models to remove rain-induced degradations. 
However, this paradigm separates rain removal and 3D reconstruction into two independent stages, which may lead to overfitting to specific rain patterns during pre-training and consequently limit the model’s generalization ability in real-world complex rainy scenarios.
Furthermore, existing methods lack a brightness adjustment mechanism, making it difficult to adapt to changes in brightness in real-world rainy environments.

These challenges motivate us to develop REVR-GSNet, an end-to-end framework that simultaneously addresses rain degradation and low brightness conditions to achieve high-quality 3D scene reconstruction. 
Specifically, the Recursive Brightness Enhancement (RBE) module progressively refines the brightness enhancement curve through recursive enhancement, gradually improving the brightness of the input images. The Gaussian Primitives Optimization (GPO) module further utilizes the enhanced images along with camera poses to construct and optimize a 3D Gaussian representation. The GS-guided Rain Elimination (GRE) module integrates the rendered images and enhanced views, and employs a residual recurrent network with recurrent refinement to remove rain streaks while simultaneously feeding back to improve the 3D reconstruction. These components collectively form a joint alternating optimization system that enhances image brightness and removes rain streaks, thereby improving the quality of 3D scene reconstruction.

To summarize, our key contributions include:
\begin{itemize}
\item We contribute OmniRain3D, a high-quality dataset for rainy 3D scene reconstruction, which simulates physically consistent rain degradation in 3D space and reduces the domain gap to real-world rainy scenes.

\item 
We propose REVR-GSNet, an end-to-end framework for high-fidelity 3D scene reconstruction through rain elimination and visibility recovery in 3D Gaussian Splatting.

\item We demonstrate the effectiveness of our dataset, and show that our proposed method achieves favorable reconstruction performance against state-of-the-art ones.
\end{itemize}

\section{Related Work}
\subsection{Image / Video Deraining}
Image deraining primarily relies on the spatial information of adjacent pixels within a single image, as well as the visual features of rain streaks and background textures.
Prior-based deraining methods utilize algorithms such as morphological component analysis~\cite{kang2011automatic} and layer priors~\cite{li2016rain} to separate rain streaks from the clean background through iterative optimization.
Techniques such as multi-scale design~\cite{wang2020dcsfn,fu2019lightweight,wang2023multi}, attention mechanisms~\cite{chen2023learning,wang2019spatial,song2024exploringanefficient,jiang2023dawn}, and multi-stage processing~\cite{chen2024bidirectional,zamir2021multi,song2024exploring,wang2024progressive} enhance the performance of single image deraining, driving progress in this field.
For video deraining, the primary goal is to separate rain and background from a sequence of frames containing temporal information while ensuring temporal consistency and preventing jitter or artifacts between consecutive frames.
% %
The initial attempts primarily rely on the linear space-time correlation~\cite{garg2004detection, garg2007vision} and physical attributes~\cite{zhang2006rain, liu2009pixel} of rain streaks. Subsequent studies adopt composite approaches integrating Fourier domain analysis~\cite{barnum2010analysis}, low-rank modeling~\cite{chen2013generalized}, and prior constraints~\cite{brewer2008using,bossu2011rain}. More recently, learning-based methods, such as Gaussian mixture model~\cite{chen2013rain, wei2017should} and deep residual network~\cite{zhang2022enhanced}, have improve modeling capabilities and advance the field.
Different from this, we use multi-view rainy images with geometric consistency for rainy 3D scene reconstruction.

\subsection{Rainy 3D Scene Reconstruction}
Some studies~\cite{lyu2024rainyscape, liu2024deraings} have begun exploring rainy 3D scene reconstruction from a dataset perspective.
However, most of these datasets adopt a 2D rain degradation synthesis approach by directly and linearly overlaying rain masks onto clean background images.
This linear composition fails to capture the physical consistency and dynamic viewpoint variations inherent in real-world rainy scenes, resulting in a significant domain gap.

To reconstruct 3D scenes under rain, 
DerainNeRF~\cite{li2024derainnerf} integrates a pre-trained deraining network~\cite{shao2021selective} with NeRF, using hard-coded masks to remove raindrops before reconstruction.
Building upon this, 
DerainGS~\cite{liu2024deraings} proposes a two-stage pipeline that first applies an pre-trained deraining network for rain removal and then performs 3D scene reconstruction using 3D Gaussian Splatting (3DGS), guided by learned occlusion masks.
However, these methods typically adopt a multi-stage processing pipeline, which may lead to overfitting to specific degradation patterns and limit the model’s generalization ability in complex scenes. 
In this work, we present a dataset modeling perspective heterogeneity and brightness dynamics for realistic rain degradation, and a unified model that removes rain and restores brightness to reconstruct clean, high-quality 3D scenes.

%----------------------------------------------------------------------------
\begin{figure}[t]
  \centering
  \includegraphics[width=\linewidth]{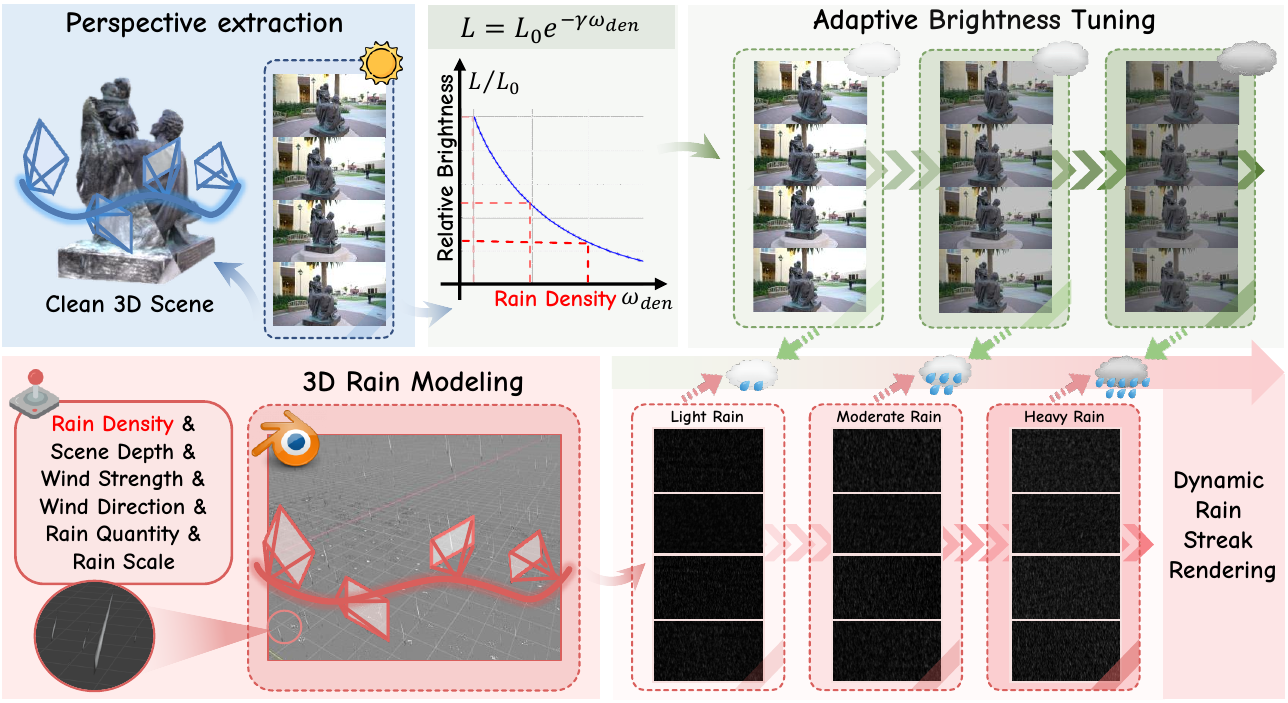}
  \vspace{-4mm}
  \caption{Overview of the data construction pipeline. Camera viewpoint extraction from clean background images.
  Rain masks are generated by rendering the 3D rain model from the same viewpoint as the background image.  The 3D rain model is controlled by parameters such as rain density.
  A brightness–rain density mapping function adjusts the brightness of background images under different rainfall levels. Finally, brightness-adjusted backgrounds are combined with rain masks to synthesize dynamically rain-degraded images, reducing the domain gap to real rainy scenes.
  }
  \label{fig_pipeline}
\vspace{-3mm}
\end{figure}
%----------------------------------------------------------------------------

\section{Proposed Dataset}
Existing rainy 3D reconstruction datasets~\cite{ liu2024deraings} overlay 2D rain streaks onto static backgrounds, ignoring the real appearance and dynamic behavior of rain.
In this paper, we develop an effective method to simulate complex dynamic rain in real-world environments, enabling the acquisition of more realistic 3D rainy scenes.

\subsection{Spatio-Temporal Rain Model}
% %
Prior works~\cite{jiang2020multi,shi2024nitedr,wang2024cascaded,wang2020dcsfn,chen2023towards} typically synthesize rainy images using a linear model:
\begin{equation}
O_t = B_t + R_t, \quad t=1,2,\dots,N,
\label{eq2}
\end{equation}
where $B_t$, $R_t$, and $O_t$ are the clean image, rain streak, and rainy image at time $t$, respectively.
The rain streaks $R_t$ are often assumed to be independent and identically distributed, with uncorrelated positions across frames~\cite{liu2018erase}.

In this work, we rethink the formation of rainy images in 3D scenes, focusing on the perspective heterogeneity of rain streak morphology and the brightness dynamicity of rainy scenes.
For perspective heterogeneity, the orientation of rain streaks varies significantly with the camera's shooting angles, as shown in Figure~\ref{fig_motivation}. 
Specifically, when the camera captures images from varying angles, {\itshape e.g.}, bottom-up view, horizontal view, and top-down view, the rain streaks in the resulting pictures exhibit distribution characteristics of $\Lambda $-shaped expansion, parallel distribution, and v-shaped convergence.
Furthermore, when the camera's shooting direction deviates from the rainfall direction, the rain streaks exhibit a distinct tilt angle determined by the relative orientation between the camera's motion trajectory and the rain's falling trajectory.
During panoramic shooting, the continuous change in camera direction causes the tilt angle of the rain streaks to dynamically adjust, creating a continuously varying visual effect.
Regarding brightness dynamicity, cloud coverage caused by rainfall typically leads to a gradual decrease in brightness. Light rain usually causes a slight reduction, while heavy rain leads to a more significant attenuation. This pattern generally holds true in most cases.

Based on these observations, we propose a spatio-tempora rain model for realistic rain imaging in 3D scene that simultaneously incorporates both temporal and spatial dimensions, defined as follows:
\begin{equation}
O_t(\theta_i,\phi_j)=L\odot (B_t(\theta_i,\phi_j) + R_t(\theta_i,\phi_j)), \label{eq3}
\end{equation}
where $L$ represents the ambient brightness under rainy conditions, $\odot$ denotes the Hadamard product, and $\theta_i$ and $\phi_j$ are the elevation and azimuthal angles, respectively.

\subsection{Data Construction Pipeline}
We present the data construction pipeline of OmniRain3D in Figure~\ref{fig_pipeline}.
Compared to previous synthesis methods~\cite{zhang2019image,zhang2018density,ran2024rainmer}, our approach generates rain images that are closer to real-world rainy 3D scenes in terms of visual realism.

{\flushleft\textbf{Perspective extraction}.}~To fully account for the characteristics of rainy images in 3D scenes, we first perform perspective extraction from the background. 
Specifically, we utilize COLMAP~\cite{schoenberger2016sfm} to estimate the extrinsic parameters of all cameras in the scene and extract their corresponding elevation and azimuth angles.
% Specifically, we employ COLMAP~\cite{schoenberger2016sfm} to visualize the camera poses and categorize the scene based on the camera's elevation angles. 
%
Each viewpoint is defined by a unique pair of elevation $\theta$ and azimuth $\phi$ angles, representing the vertical and horizontal positioning of the camera relative to the target.
We discretize the viewing space into a grid of $W$ elevation angles and $U$ azimuth angles, constructing a viewpoint matrix as:
\begin{equation}
\mathbf{Z} = \{ (\theta_i, \phi_j) \mid i = 1, \ldots, W;\, j = 1, \ldots, U \},
\end{equation}
where each pair $(\theta_i, \phi_j)$ defines a camera pose and corresponds to one view in the 3D multi-view image set.

{\flushleft\textbf{Dynamic rain streak rendering}.}
To realistically simulate rain appearance across multiple viewpoints, we perform 3D rain modeling with multi-dimensional meteorological features using Blender~\cite{hess2013blender}:
\begin{equation}
S =\{ \omega_{den},\omega_{dep}, \omega_{str}, \omega_{dir}, \omega_{qty}, \omega_{scl} \},
\end{equation}
where $\omega_{den}$, $\omega_{dep}$, $ \omega_{str}$, $\omega_{dir}$, $\omega_{qty}$, and $\omega_{scl}$ represent rain density, scene depth, wind strength, wind direction, rain quantity, and rain scale, respectively.
Subsequently, for each camera pose $(\theta_i, \phi_j)$, we perform viewpoint-synchronized rendering of the rain streak model $S$ using the corresponding camera parameters from the viewpoint set $\mathbf{Z}$, thereby ensuring that the morphology and visual effects of rain streaks under different viewpoints are consistent with real-world scenes.
The rendering process is defined as follows:
\begin{equation}
R_t(\theta _i,\phi _j) =\mathcal{R}(S,(\theta _i,\phi _j)),
\end{equation}
where $R_t(\theta _i,\phi _j)$ represents the result of rain streak rendering under viewpoint $(\theta _i,\phi _j)$, $\mathcal{R}( \cdot )$ is rendering process.

{\flushleft\textbf{Adaptive brightness tuning}.}
In the real world, an increase in rainfall commonly leads to a decrease in ambient brightness, primarily due to cloud cover and atmospheric scattering effects.
Inspired by atmospheric attenuation theory, we model the relationship between rainfall intensity and scene brightness using an exponential decay formulation. Specifically, based on the Beer–Lambert law~\cite{swinehart1962beer} and empirical rain attenuation models~\cite{nia2025exploring}, the observed brightness $L$ under rain density $\omega_{den}$ is given by:
\begin{equation}
L =L_0e^{-\gamma \omega_{den}} ,
\label{eq1}
\end{equation}
where $L_0$ represents the baseline ambient brightness under rain-free conditions, $\omega_{den}$ denotes the rain density (a physically-based parameter adjustable in Blender), and $\gamma$ quantifies the atmospheric attenuation coefficient.

To enhance the diversity and realism of our dataset, we simulate varying rainy conditions by setting three distinct levels of rain density $\omega_{den}$, corresponding to light rain, moderate rain, and heavy rain.
For each level, we compute the corresponding ambient brightness using Eq.~\ref{eq1}, resulting in three luminance conditions that reflect typical visibility degradation observed in real rainy scenarios.
Finally, the rain streaks generated under different densities are spatially aligned and superimposed on backgrounds with their corresponding brightness levels, producing the final rainy scenes with physically consistent degradation.

\section{Proposed Method}
\begin{figure}[t]
  \centering
  \includegraphics[width=\linewidth]{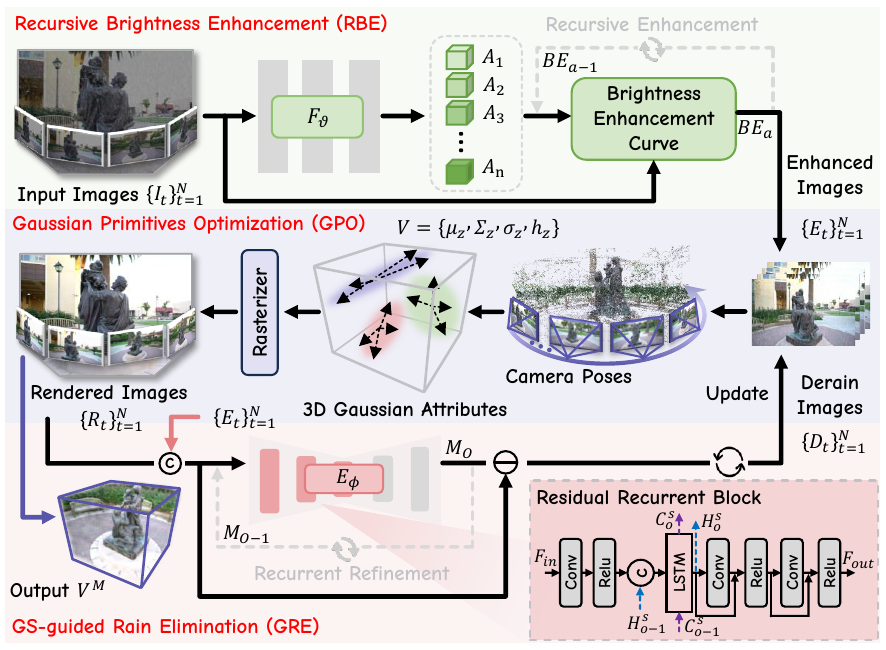}
  \vspace{-6mm}
  \caption{Overview of the REVR-GSNet framework. The RBE progressively improves the brightness of multi-view rainy images. The GPO constructs and optimizes a 3D Gaussian representation using enhanced images and camera poses via differentiable rendering. The GRE fuses rendered and enhanced images to remove rain streaks through a residual recursive network, while also feeding back to refine 3D reconstruction. These components employ a joint alternating optimization strategy that enhances brightness, removes rain streaks, and improves 3D scene reconstruction quality.
  }
  \label{pipline}
\vspace{-3mm}
\end{figure}

To reconstruct a rain-free and brightness-corrected 3D Gaussian scene representation $V^M$, we employ the proposed REVR-GSNet to process a set of $N$ multi-view rainy images $\left \{ I_t\right \}_{t=1}^N$, as illustrated in Figure~\ref{pipline}.
Specifically, our framework adopts a joint alternating optimization strategy.
Initially, REVR-GSNet performs joint optimization of the Recursive Brightness Enhancement (RBE) and Gaussian Primitive Optimization (GPO) modules.
In each iteration, the rainy inputs $I_t$ are first processed by RBE for brightness correction. 
The enhanced images $E_t$ are then passed to the GPO module, where differentiable 3D Gaussian splatting is applied to refine the representation of the radiance field $V$.
As the iterations proceed, both the brightness enhancement curves and the 3D Gaussian attributes are progressively optimized, leading to improved brightness and structural fidelity of the scene.
Once the brightness information has been effectively embedded into the Gaussian representation, the RBE step is removed from subsequent iterations.
At this stage, the framework focuses on the joint optimization of GPO and Gaussian-guided Rain Elimination (GRE).
The brightness-corrected images $E_t$ and the rendered images $R_t$ jointly guide the GRE module, which progressively removes rain artifacts through a recurrent refinement process.
The derained outputs $D_t$ are then fed back into GPO as updated inputs, enabling further refinement of the 3D Gaussian attributes and continued improvement in both 3D reconstruction quality and visual clarity.
In the final iteration, only the GPO branch is retained to produce the clean and complete radiance field $V^M$.
The following sections provide detailed descriptions of each component within REVR-GSNet.

% 亮场景
\begin{table*}[t]
\renewcommand\arraystretch{1.1}
\vspace{-2mm}
\resizebox{\linewidth}{!}{
\begin{tabular}{c|ccccccccccccccc|ccc}
\toprule
\multirow{2}{*}{Scene} & \multicolumn{3}{c}{DRSformer*} & \multicolumn{3}{c}{NeRD-Rain*} & \multicolumn{3}{c}{DerainNeRF} & \multicolumn{3}{c}{RainyScape} & \multicolumn{3}{c|}{DerainGS} & \multicolumn{3}{c}{REVR-GSNet (Ours)} \\
 \cmidrule(r){2-4}\cmidrule(lr){5-7}\cmidrule(lr){8-10}\cmidrule(lr){11-13}\cmidrule(lr){14-16}\cmidrule(lr){17-19}
                       & PSNR$\uparrow$     & SSIM$\uparrow$     & LPIPS$\downarrow$   & PSNR$\uparrow$     & SSIM$\uparrow$     & LPIPS$\downarrow$   & PSNR$\uparrow$     & SSIM$\uparrow$     & LPIPS$\downarrow$    & PSNR$\uparrow$     & SSIM$\uparrow$     & LPIPS$\downarrow$    & PSNR$\uparrow$     & SSIM$\uparrow$     & LPIPS$\downarrow$   & PSNR$\uparrow$   & SSIM$\uparrow$   & LPIPS$\downarrow$  \\ \hline
Francis                & \underline{24.44}    & 0.773    & \textbf{0.330}   & 24.34    & 0.769    & \underline{0.332}  & 16.17    & 0.599    & 0.577    & 22.99    & 0.749    & 0.362    & 23.40    & \underline{0.779}    & 0.354   & \textbf{24.56}  & \textbf{0.785}  & 0.336  \\
Lgnatius               & 18.92    & 0.543    & \textbf{0.386}   & 18.74    & 0.528    & \underline{0.399}  & 13.22    & 0.293    & 0.670    & 18.50    & 0.505    & 0.412    & \underline{18.98}  & \textbf{0.567}    & 0.400   & \textbf{19.44}  & \underline{0.560}  & 0.401  \\
Caterpillar            & \underline{21.16}   & \underline{0.605}   &  \textbf{0.381}   & 21.07    & 0.596    & \underline{0.385}  & 13.99    & 0.338    & 0.664    & 19.90    & 0.562    & 0.407    & 20.26    & 0.601    & 0.423   & \textbf{21.48 } & \textbf{0.617} & 0.394  \\
Garden                 & \textbf{25.39}   & 0.751    & 0.230   & 24.97    & 0.740    & 0.241   & 21.74    & 0.619    & 0.320    & 22.58    & 0.712    & 0.241    & 25.30    & \textbf{0.793}   & \underline{0.200}  & \underline{25.35} & \underline{0.790} & \textbf{0.184} \\
\bottomrule
\end{tabular}
}
\caption{Performance comparison of different methods on rain streak scenes from OmniRain3D dataset.
The best and second-best values are blod and underlined.
% The best performance is shown in red color, and the second-best performance is blue color. 
Methods marked with $*$ indicate preprocessing.}
\label{Tab1_light}
\end{table*}

% 暗场景
\begin{table*}[t]
\renewcommand\arraystretch{1.1}
\vspace{-2mm}
\resizebox{\linewidth}{!}{
\begin{tabular}{c|ccccccccccccccc|ccc}
\toprule
\multirow{2}{*}{Scene} & \multicolumn{3}{c}{DRSformer* $\dagger$} & \multicolumn{3}{c}{NeRD-Rain* $\dagger$} & \multicolumn{3}{c}{DerainNeRF $\dagger$} & \multicolumn{3}{c}{RainyScape $\dagger$} & \multicolumn{3}{c|}{DerainGS $\dagger$} & \multicolumn{3}{c}{REVR-GSNet (Ours)} \\
 \cmidrule(r){2-4}\cmidrule(lr){5-7}\cmidrule(lr){8-10}\cmidrule(lr){11-13}\cmidrule(lr){14-16}\cmidrule(lr){17-19}
                       & PSNR$\uparrow$     & SSIM$\uparrow$     & LPIPS$\downarrow$   & PSNR$\uparrow$     & SSIM$\uparrow$     & LPIPS$\downarrow$   & PSNR$\uparrow$     & SSIM$\uparrow$     & LPIPS$\downarrow$    & PSNR$\uparrow$     & SSIM$\uparrow$     & LPIPS$\downarrow$    & PSNR$\uparrow$     & SSIM$\uparrow$     & LPIPS$\downarrow$   & PSNR$\uparrow$   & SSIM$\uparrow$   & LPIPS$\downarrow$  \\ \hline
Flowers                & 14.37    & 0.387    & \textbf{0.475}  & 14.20    & 0.385    & 0.483   & 15.00    & 0.310    & 0.593    & 14.81    & \underline{0.397}    & 0.499    & \underline{15.24}    & 0.387    & \underline{0.476 }  & \textbf{15.36} & \textbf{0.398}  & 0.490  \\
Stump                  & 17.66    & 0.417    & \underline{0.493}  & 17.73    & 0.415    & 0.504   & \underline{18.63}   & 0.321    & 0.599    & 18.37    & \underline{0.432}    & 0.512    & 18.53    & \textbf{0.475}   & \textbf{0.480}  & \textbf{18.78} & 0.418  & 0.506  \\
Bicycle                & 18.31    & 0.473    & \underline{0.401}  & 18.44    & 0.477    & 0.419   & 18.13    & 0.349    & 0.573    & 18.63    & \underline{0.548}   & 0.431    & \underline{18.88}   & 0.497    & 0.406  & \textbf{19.06}  & \textbf{0.554}  & \textbf{0.397} \\
Family                 & 17.49     & 0.644    & \underline{0.458}   & 17.51     & 0.641    & 0.463   & 17.05     & 0.593    & 0.595    & 16.92     & \underline{0.645}   & 0.497    & \underline{17.78}     & 0.633    & 0.461   & \textbf{17.83} & \textbf{0.657} & \textbf{0.440} \\
\bottomrule
\end{tabular}
}
\caption{Performance comparison of different methods on rain streak scenes with varying brightness from the OmniRain3D dataset.
Methods marked with $\dagger$ indicate the use of low brightness enhancement preprocessing. 
}
\label{Tab1_dim_light}
\vspace{-3mm}
\end{table*}

\subsection{Recursive Brightness Enhancement}
To correct brightness degradation caused by low-brightness rainy conditions, we propose a recursive brightness enhancement that progressively enhances image brightness via a learnable curve-based mapping.
Specifically, we adopt a lightweight CNN-based Curve Parameter Estimation Network (CPEN) composed of seven convolutional layers with symmetric skip connections to estimate brightness adjustment parameters from the input image $I_t$:
\begin{equation}
\left \{ A_a\right \}_{a=1}^n= F_ \vartheta (I_t),
\end{equation}
where $A_a\in \mathbb{R} ^{1\times 3}$ denotes the curve parameter for the  $a$-th  recursive step, and $F_ \vartheta $ is the CPEN.
Given the predicted parameters $A_a$, we define a quadratic brightness enhancement curve (BE-curve)~\cite{guo2020zero} as:
\begin{equation}
\mathbf{BE}(I_t,A_1)= I_t+A_1 I_t(1-I_t), 
\end{equation}
where $\mathbf{BE}(I_t,A_1)$ denotes the enhanced version of $I_t$.
To improve brightness progressively, the BE-curve is applied recursively using brightness adjustment parameters in each iteration. The recursive enhancement is expressed as:
\begin{equation}
BE_a = \mathbf{BE}(BE_{a-1},A_a), 
\label{eq10}
\end{equation}
where $BE_{0}=I_t$, and the final enhanced image $E_t$ is obtained after $n$ recursive steps.

\subsection{Gaussian Primitives Optimization}
As the iterative process advances, the brightness of the enhanced images generated by RBE progressively improves.  
We estimate camera poses from the enhanced image set $\{E^i_t\}_{t=1}^N$ using COLMAP~\cite{schoenberger2016sfm}.  
%
% , where $i \in [1, M]$ denotes the current iteration index
%
With the estimated poses and the corresponding enhanced images, we aggregate multi-view information to construct a 3D Gaussian scene representation:
\begin{equation}
V = \{\mu_z, \Sigma_z, \sigma_z, h_z\},
\end{equation}
where $\mu_z$ represents the spatial location, $\Sigma_z$ the covariance matrix, $\sigma_z$ the opacity, and $h_z$ the spherical harmonic coefficients for appearance modeling.
After successfully encoding brightness information into the Gaussian attributes, the RBE step is no longer required in later iterations.

It is worth noting that the enhanced images still contain rain streaks, which could introduce noise during scene aggregation.
However, by leveraging cross-view consistency and spatial correlation, the radiance field optimization process can effectively suppress these artifacts, producing higher-fidelity reconstructions than single image alone.

\subsection{GS-guided Rain Elimination}
To further eliminate rain artifacts, we introduce the GS-guided rain elimination that leverages the current 3D Gaussian scene representation $V$ to guide the deraining process.
For each camera poses, we render the reference image $R_t$ from the Gaussian scene through differentiable rasterization.
This rendered image preserves consistent scene content while exhibiting fewer rain artifacts and sharper structures compared to the original enhanced image~\cite{lyu2024rainyscape, choi2025exploiting}.

Motivated by this, we adopt a Recurrent Rain Estimation Network (RREN) that estimates a rain streak map $M_l$  from the rendered images and enhanced images, and derives the derained image $D_t$ via residual subtraction.
The process is defined as follows:
\begin{equation}
D_t = Cat(R_t, E_t)-E_\phi(Cat(R_t, E_t)),
\end{equation}
where $E_\phi$ is the RREN and $Cat$ denote concatenation operations. 
To enable progressive refinement, $E_\phi$ employ a recurrent U-Net architecture integrated with LSTM units~\cite{li2022single}. At each recurrent step  $o \in [1,l]$, the decoder generates output $M_o$ based on both the current input $Cat(R_t, E_t)$ and previous output $M_{o-1}$, forming the recurrence relation:
\begin{equation}
M_o = \mathbf{RREN}(M_{o-1}, Cat(R_t, E_t)).
\end{equation}

To enhance spatial-temporal feature modeling, we introduce Residual Recurrent Blocks (RRBs), each embedding a convolutional LSTM within a standard residual block. As shown in Figure~\ref{pipline}, each RRB takes three inputs: the previous hidden state $H_{o-1}^s$, cell state $C_{o-1}^s$, and the feature map from the previous ReLU layer, where $s$ denotes the scale level.
Through gated memory mechanisms, the LSTM units selectively retain rain-relevant information while filtering out irrelevant content, thereby producing updated hidden and cell states $H_{o}^s$ and  $C_{o}^s$ for the next recurrent step.
Once the updated derained images $D_t$ are obtained, we return to the radiance field construction stage and initiate a new iteration. This process progressively improves both the visual quality of the images and the fidelity of the reconstructed 3D scene.

%----------------------------------------------------------------------------
% %%%%%%%%%%%%%%%%%%%%%%%%%%%%%%%%%%%%%%%%%%%%%%%%%%%%%%%%%%%%%
\begin{figure*}[t]
    \centering
    \setlength{\tabcolsep}{0.5pt} % 设置列间距
    \begin{tabular}{ 
        >{\centering\arraybackslash}m{0.14\textwidth} 
        >{\centering\arraybackslash}m{0.14\textwidth} 
        >{\centering\arraybackslash}m{0.14\textwidth} 
        >{\centering\arraybackslash}m{0.14\textwidth} 
        >{\centering\arraybackslash}m{0.14\textwidth} 
        >{\centering\arraybackslash}m{0.14\textwidth}
        >{\centering\arraybackslash}m{0.14\textwidth} 
    }

      % 第一行：图像
    \includegraphics[width=\linewidth]{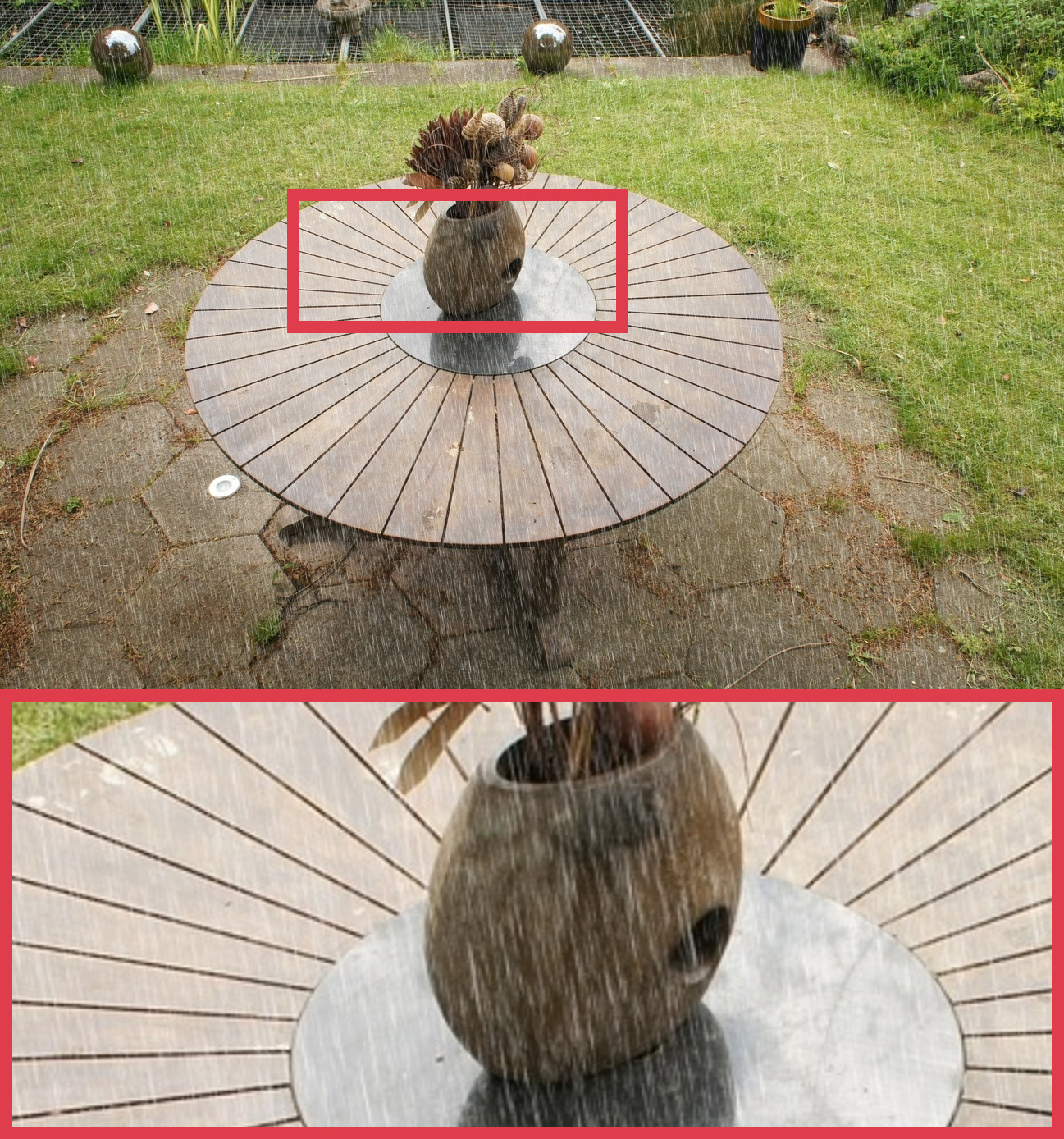} &
    \includegraphics[width=\linewidth]{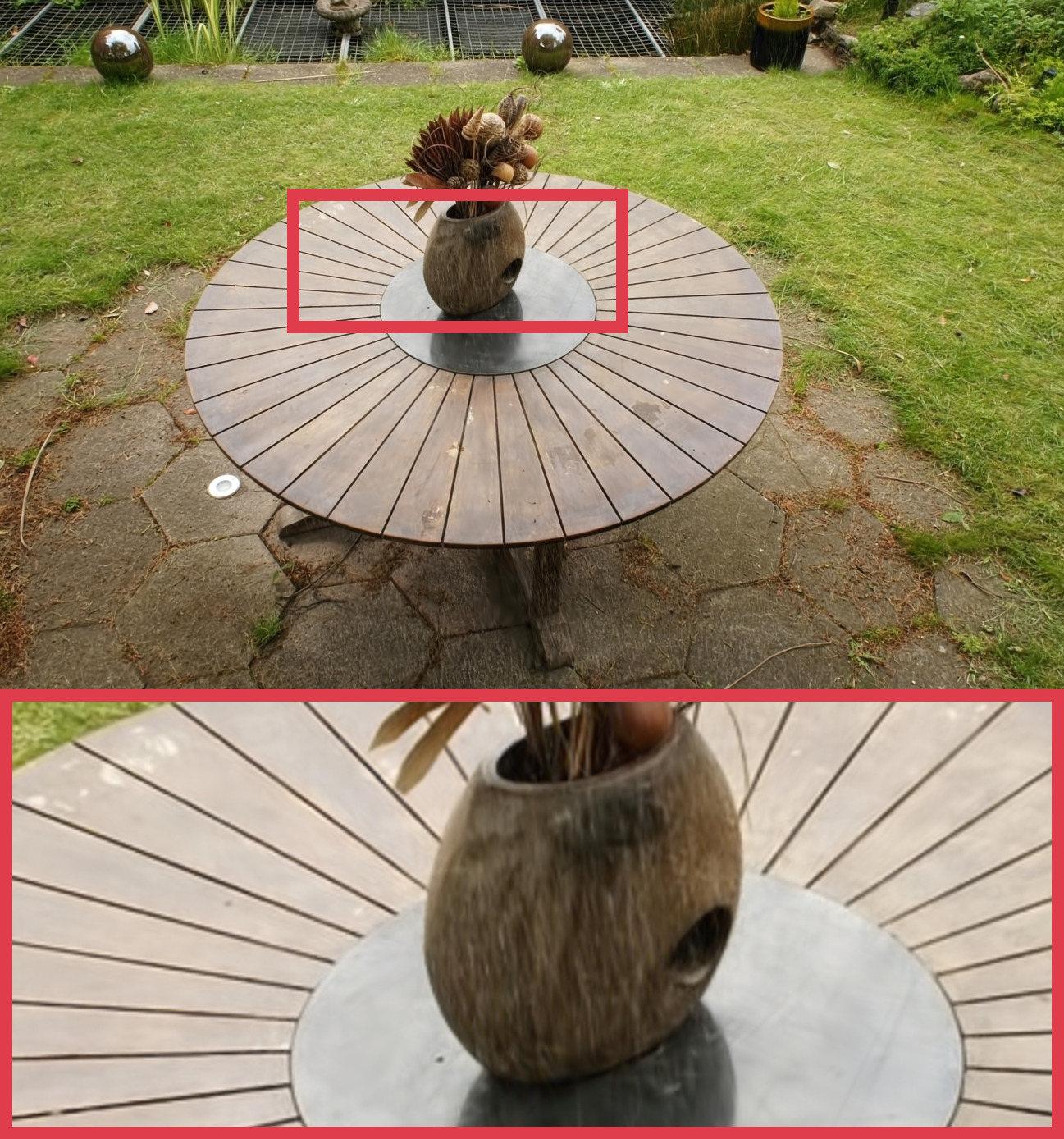} &
    \includegraphics[width=\linewidth]{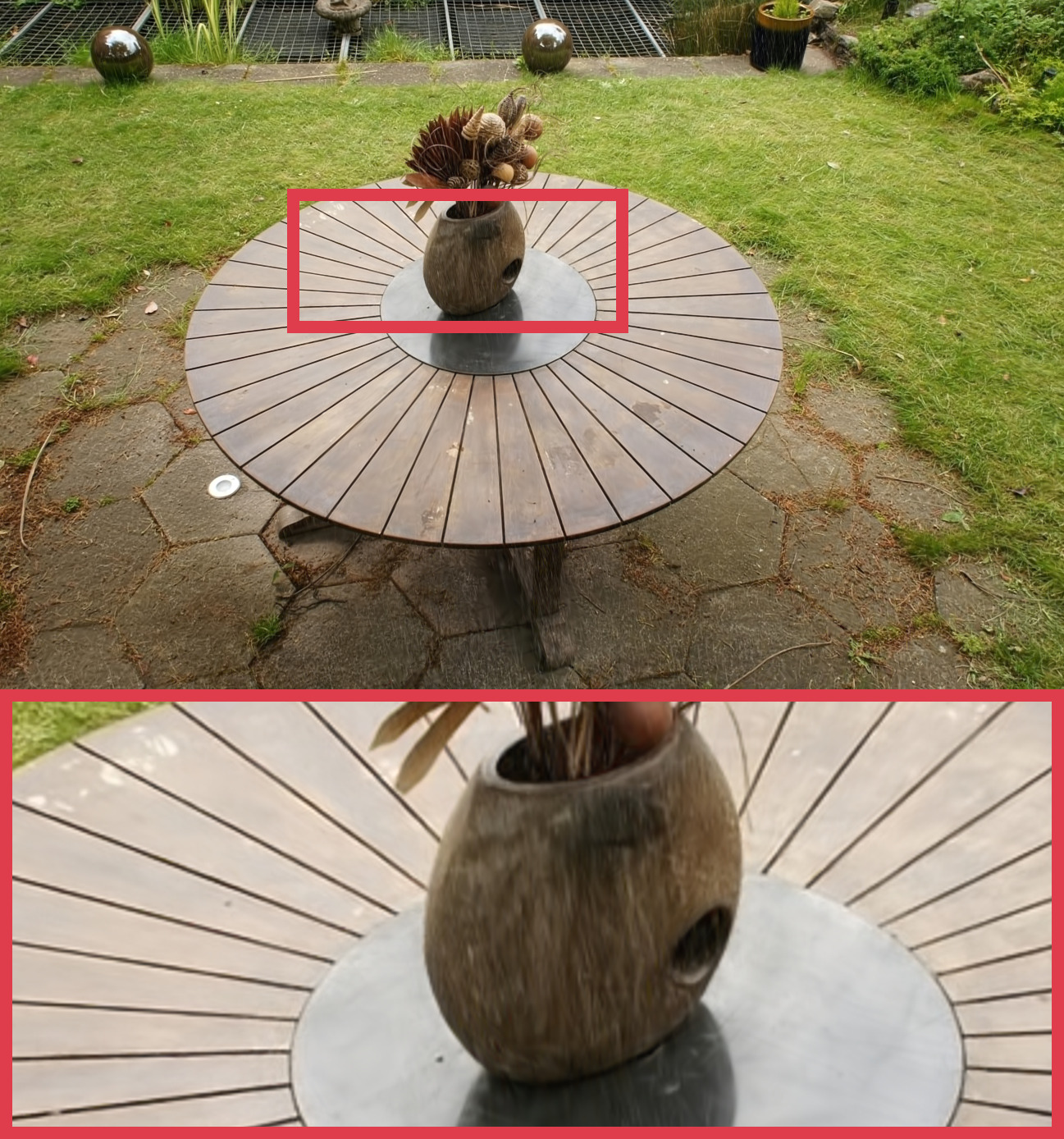} &
    \includegraphics[width=\linewidth]{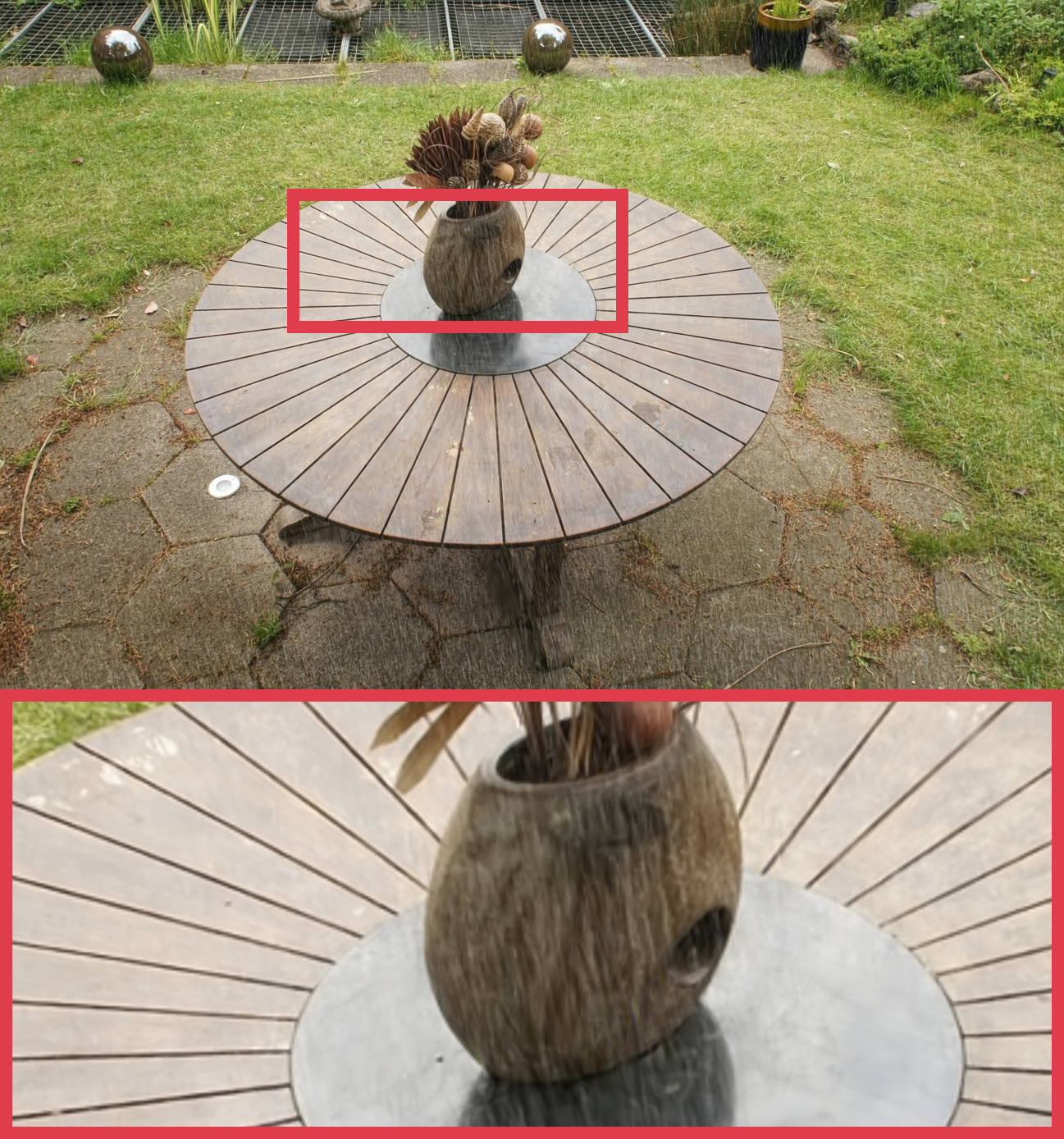} &
    \includegraphics[width=\linewidth]{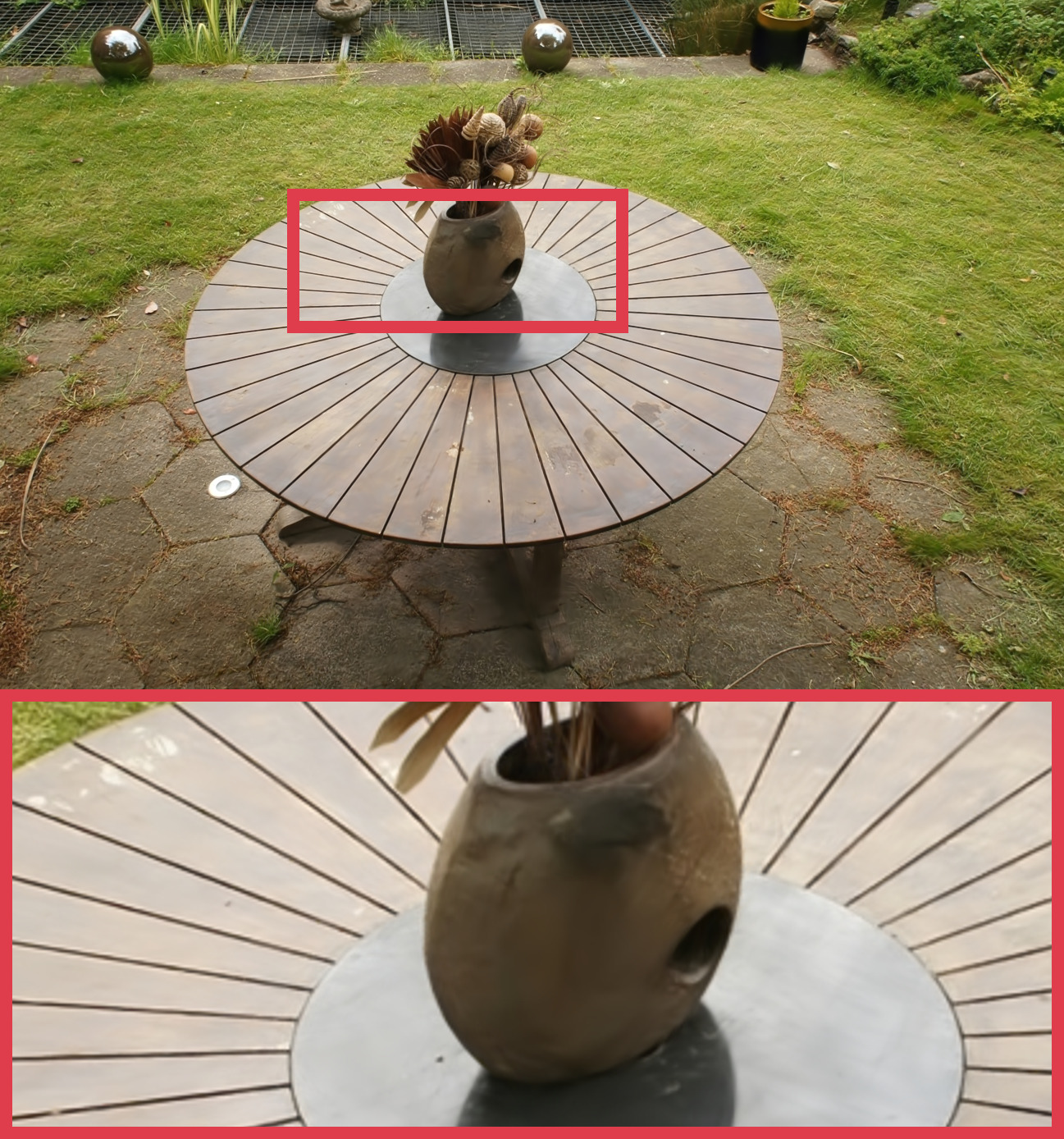} &
    \includegraphics[width=\linewidth]{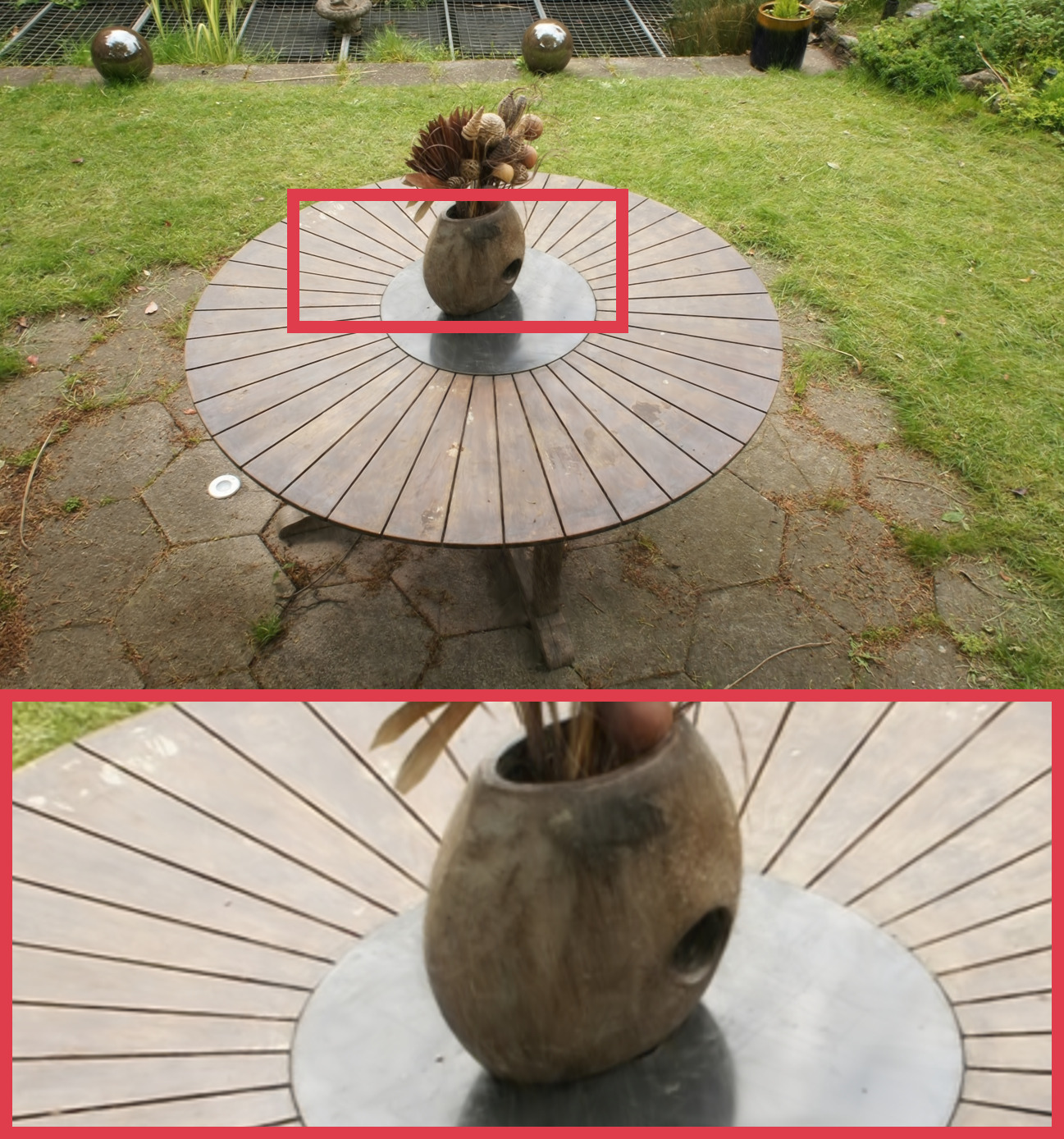} &
    \includegraphics[width=\linewidth]{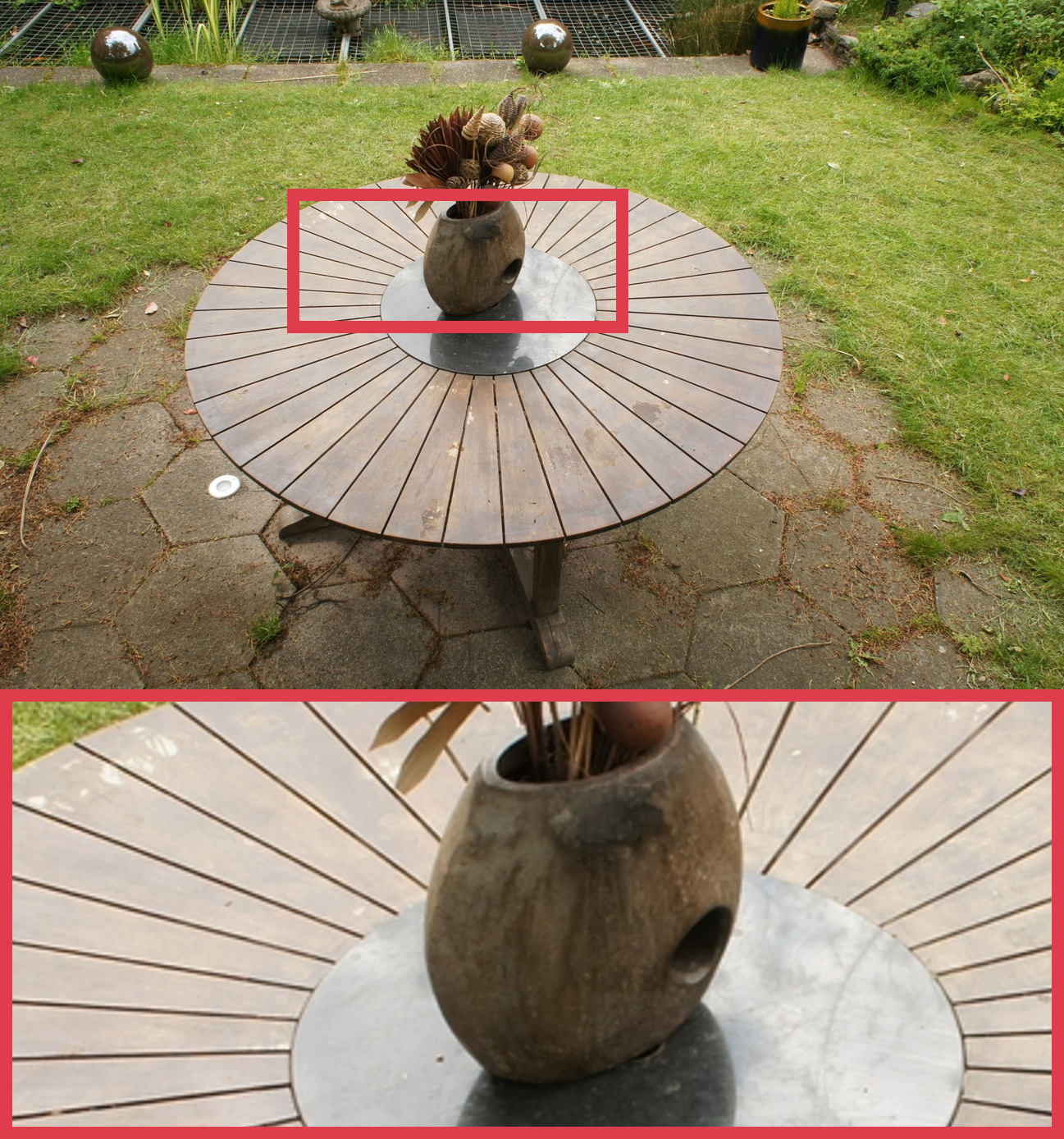}\\[10pt] 
    % 第一行：图像
    Input & 
    DRSformer*& 
    NeRD-Rain*& 
    RainyScape & 
    DerainGS & 
    Ours & 
    GT \\
    % 第一行：图像
    \includegraphics[width=\linewidth]{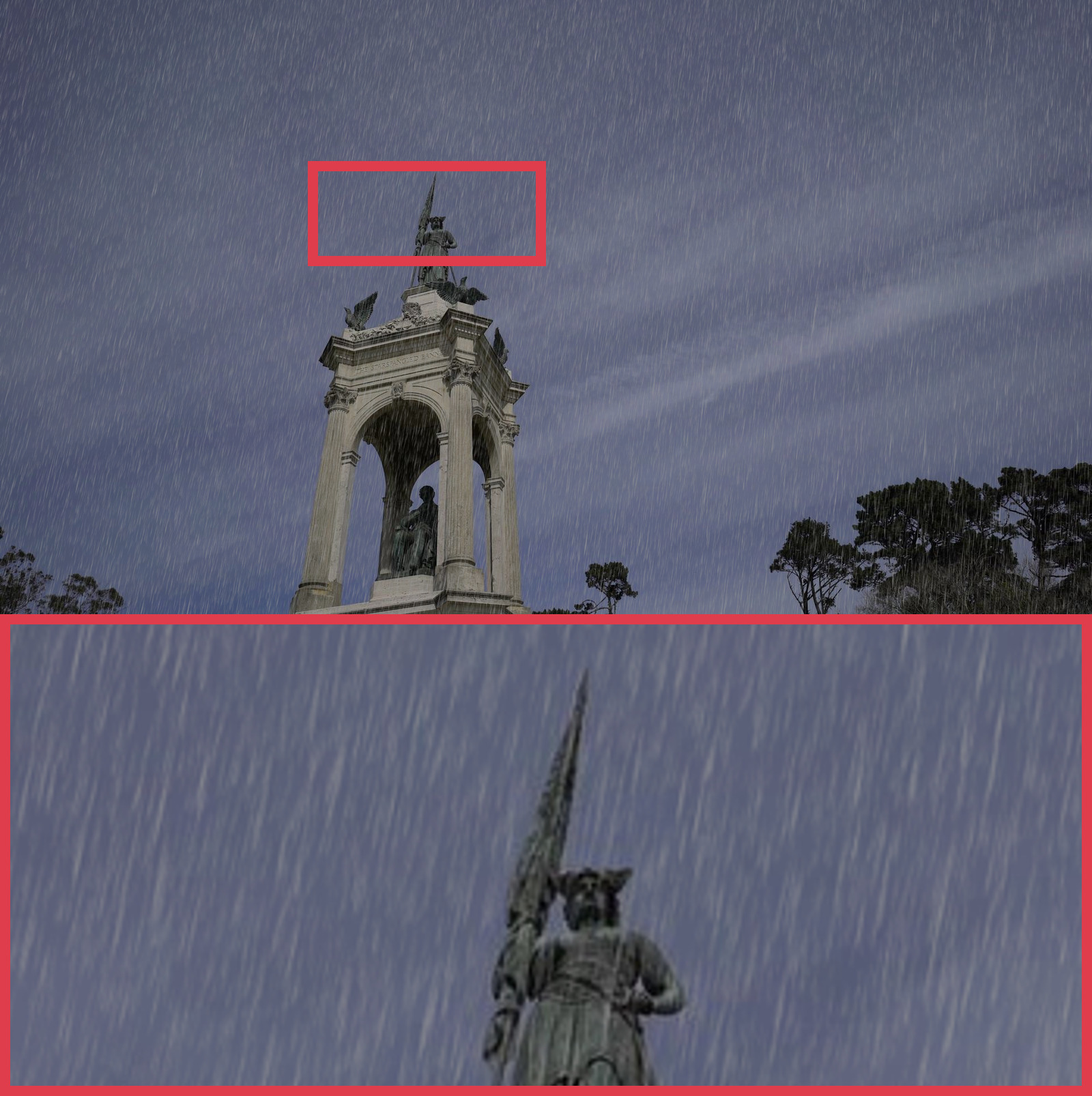} &
    \includegraphics[width=\linewidth]{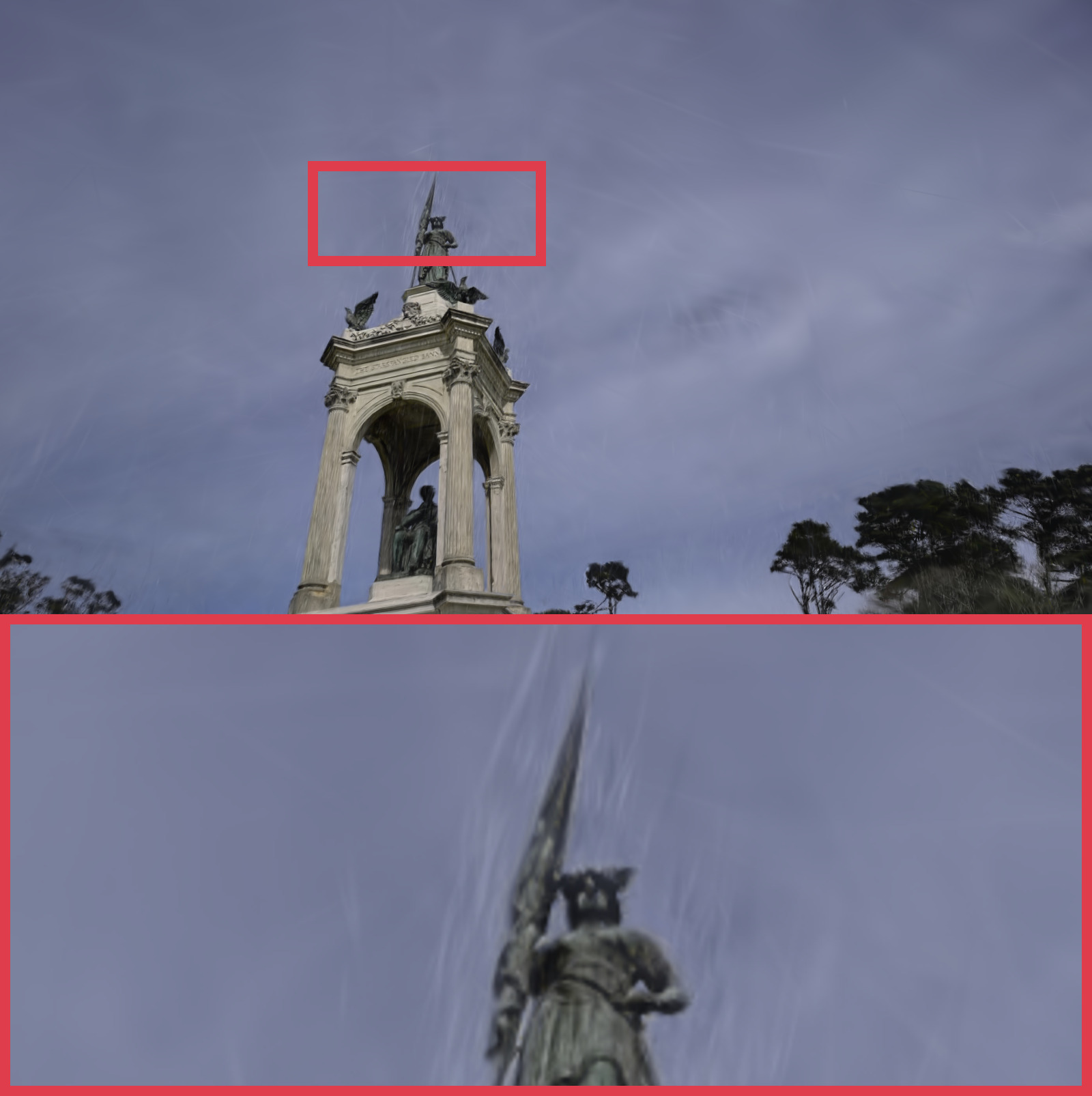} &
    \includegraphics[width=\linewidth]{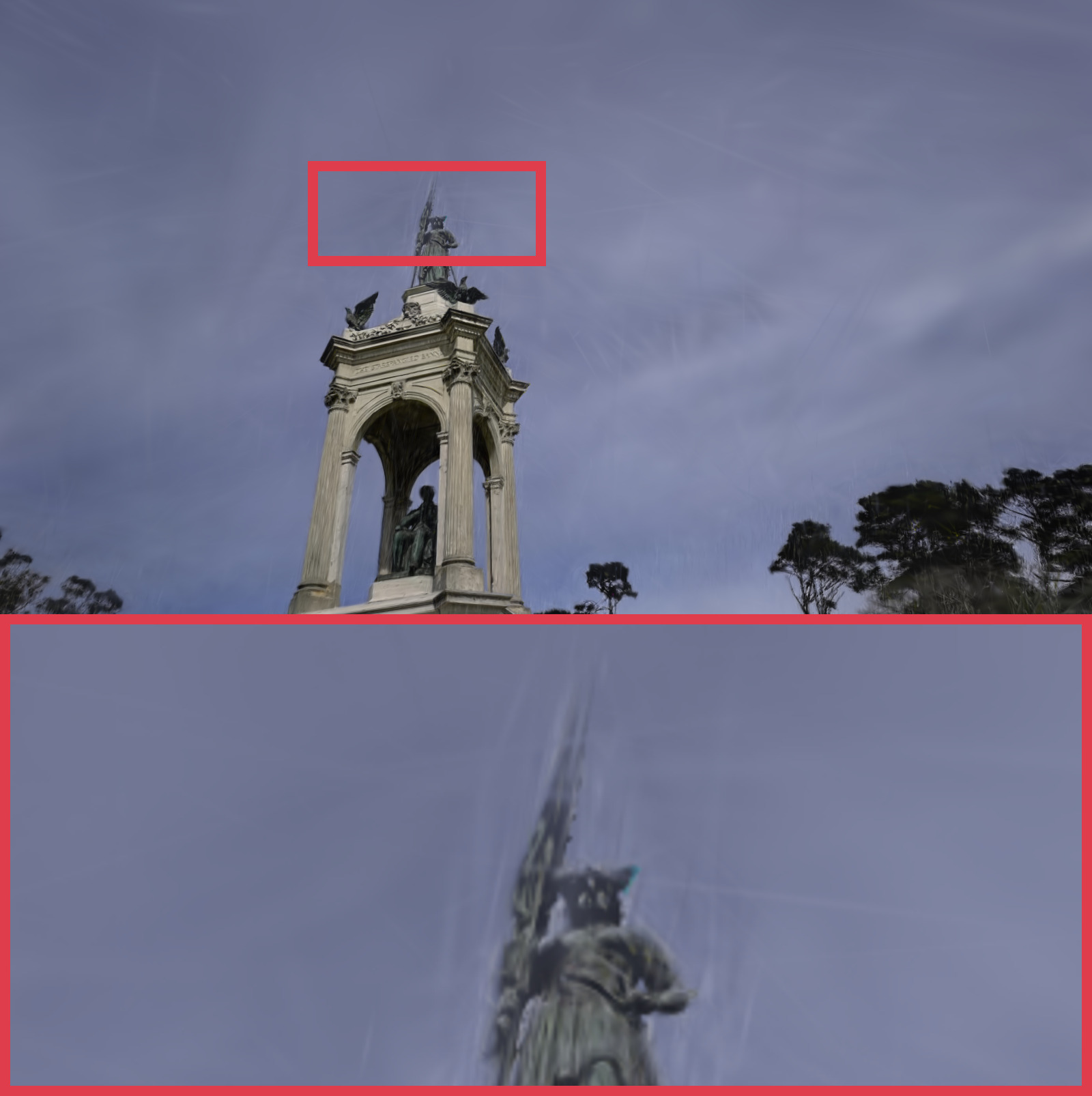} &
    \includegraphics[width=\linewidth]{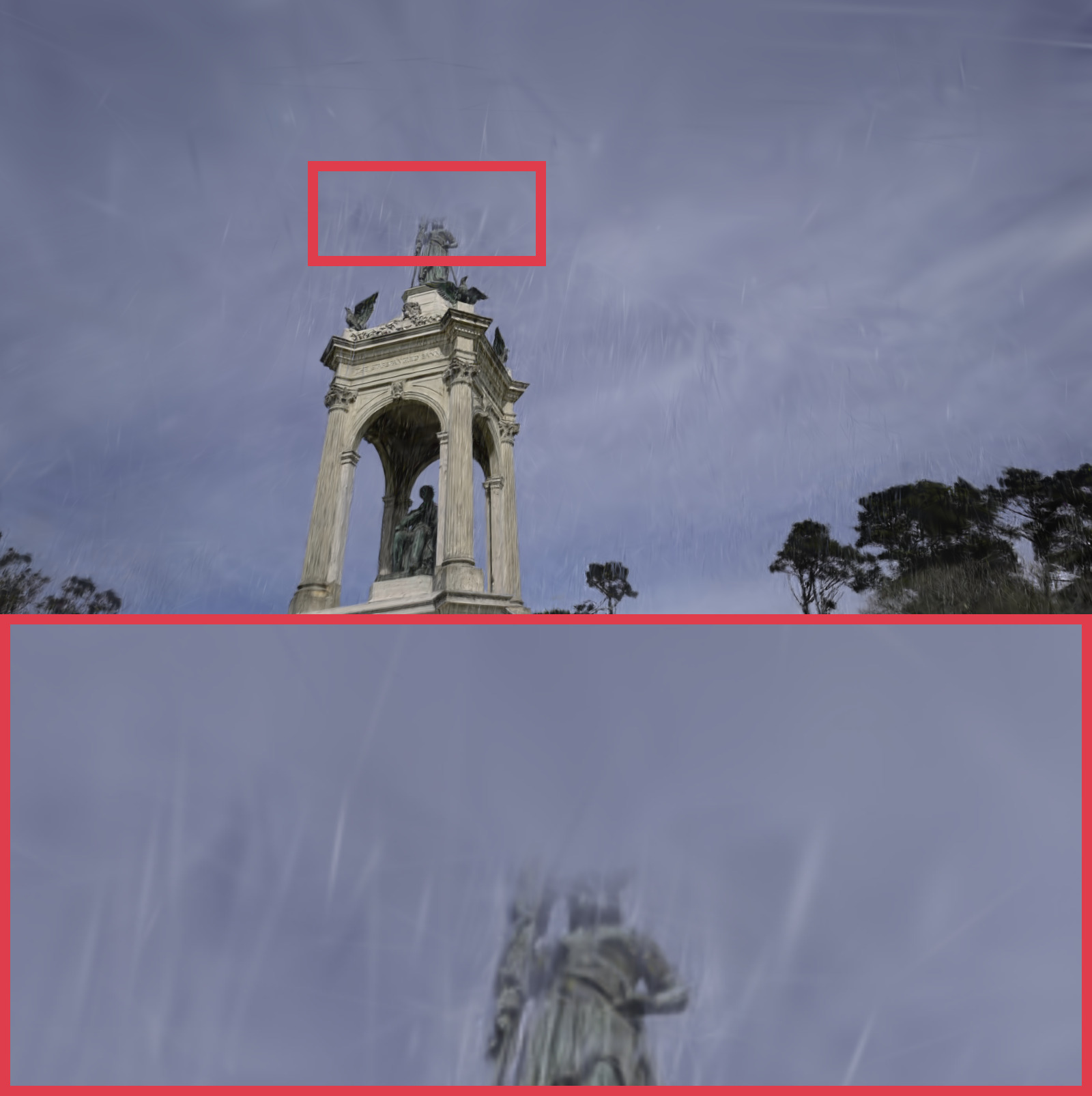} &
    \includegraphics[width=\linewidth]{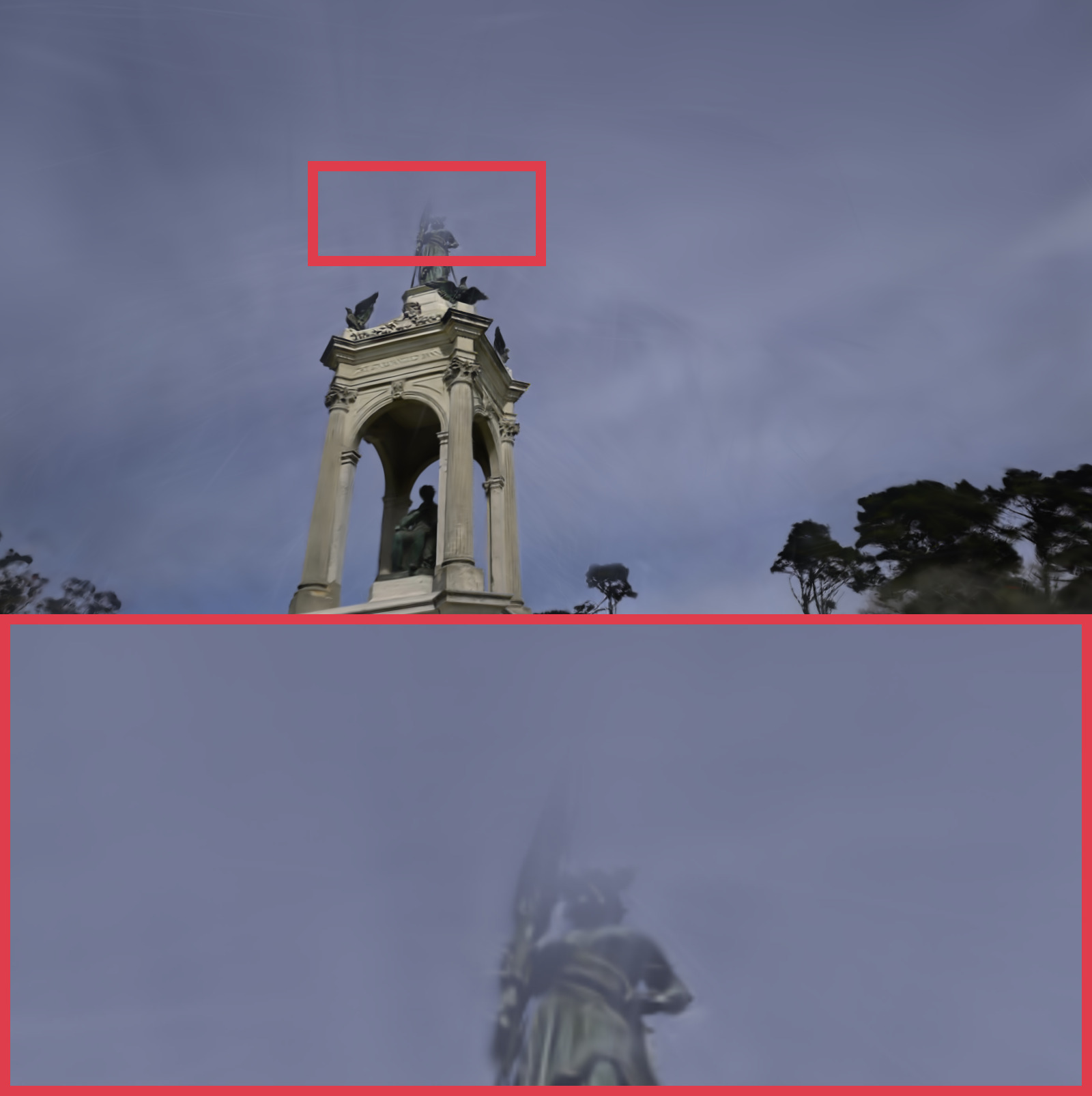} &
    \includegraphics[width=\linewidth]{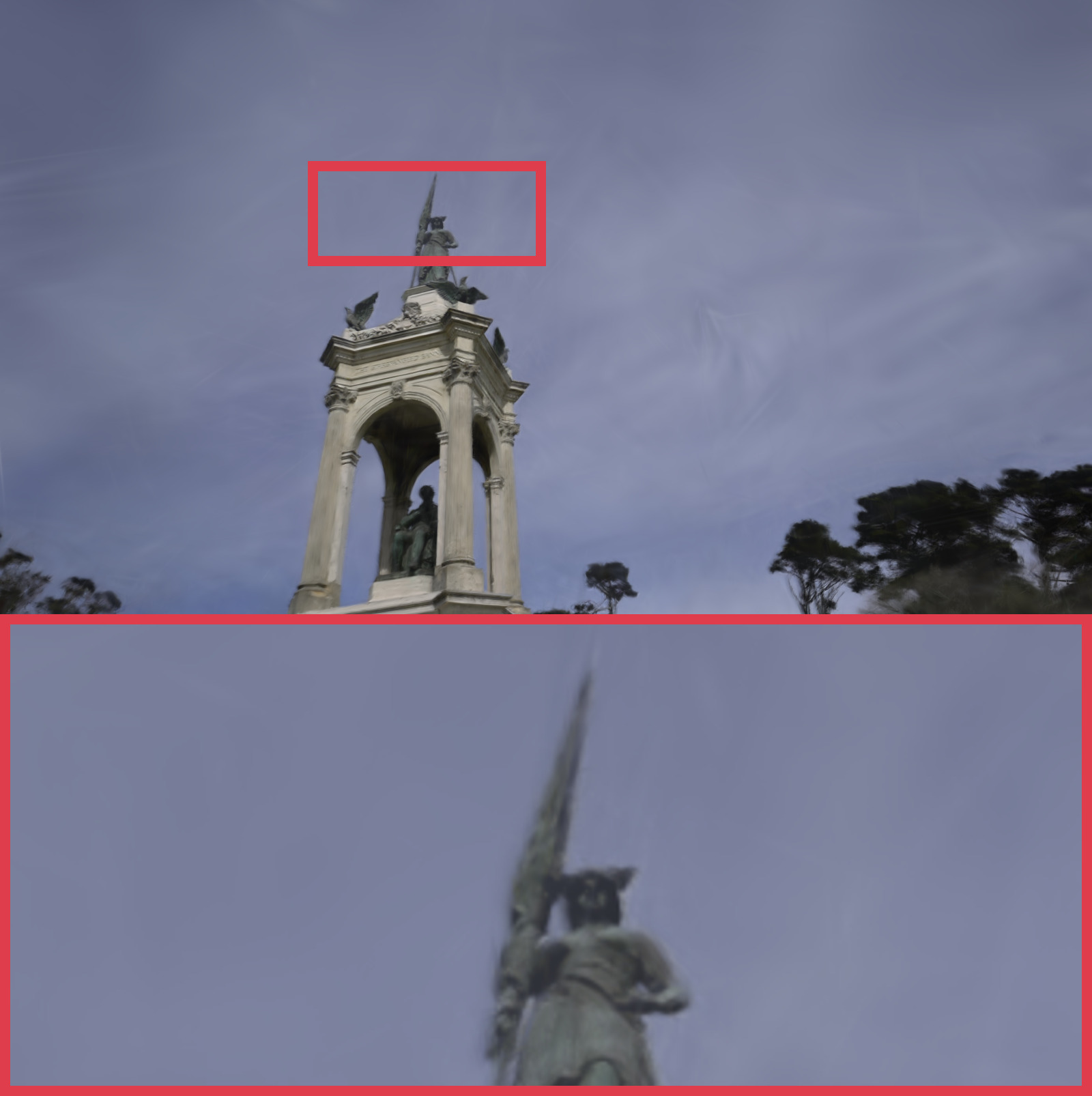} &
    \includegraphics[width=\linewidth]{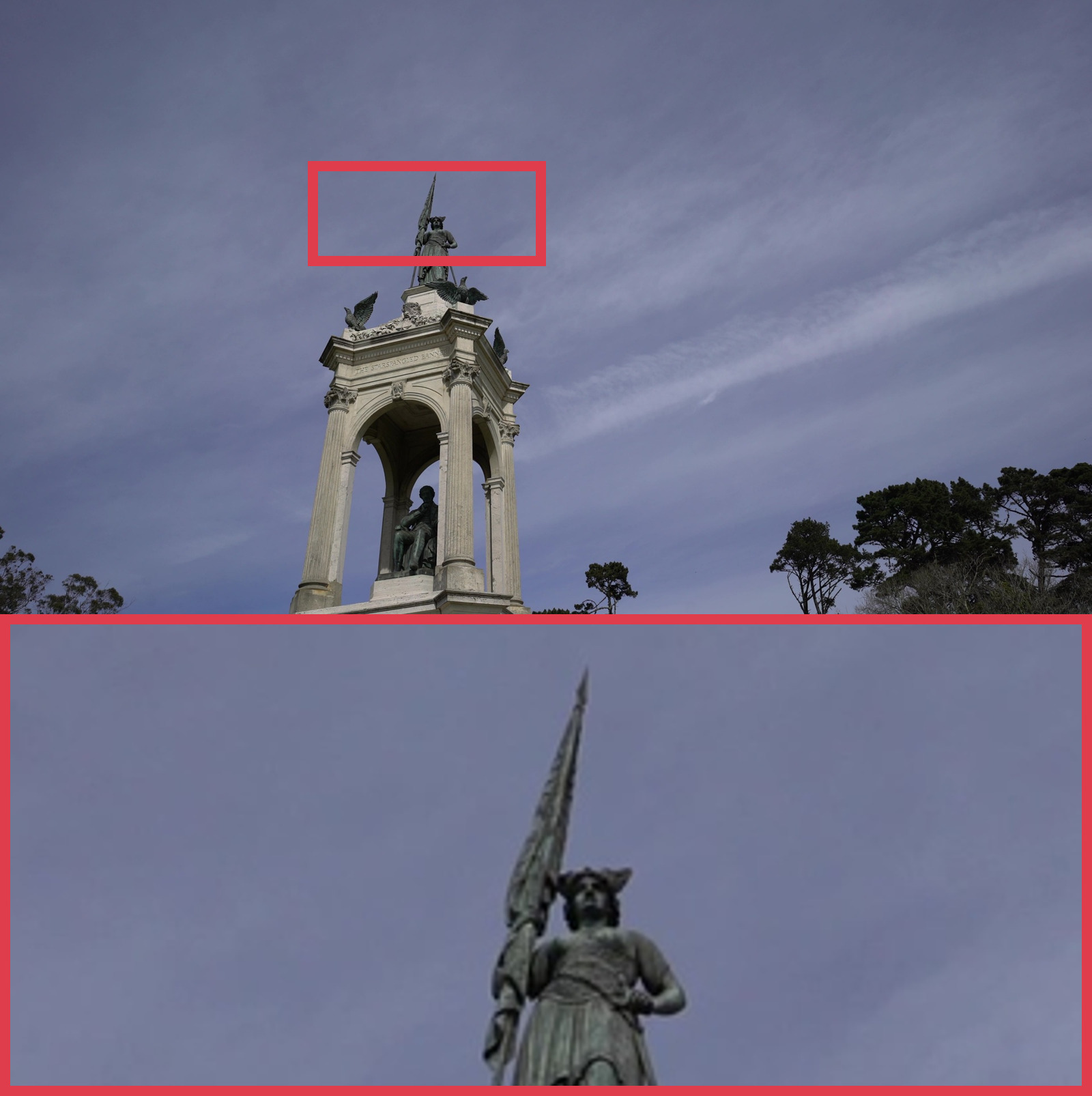}\\[10pt] 
     % 第一行：图像
    \includegraphics[width=\linewidth]{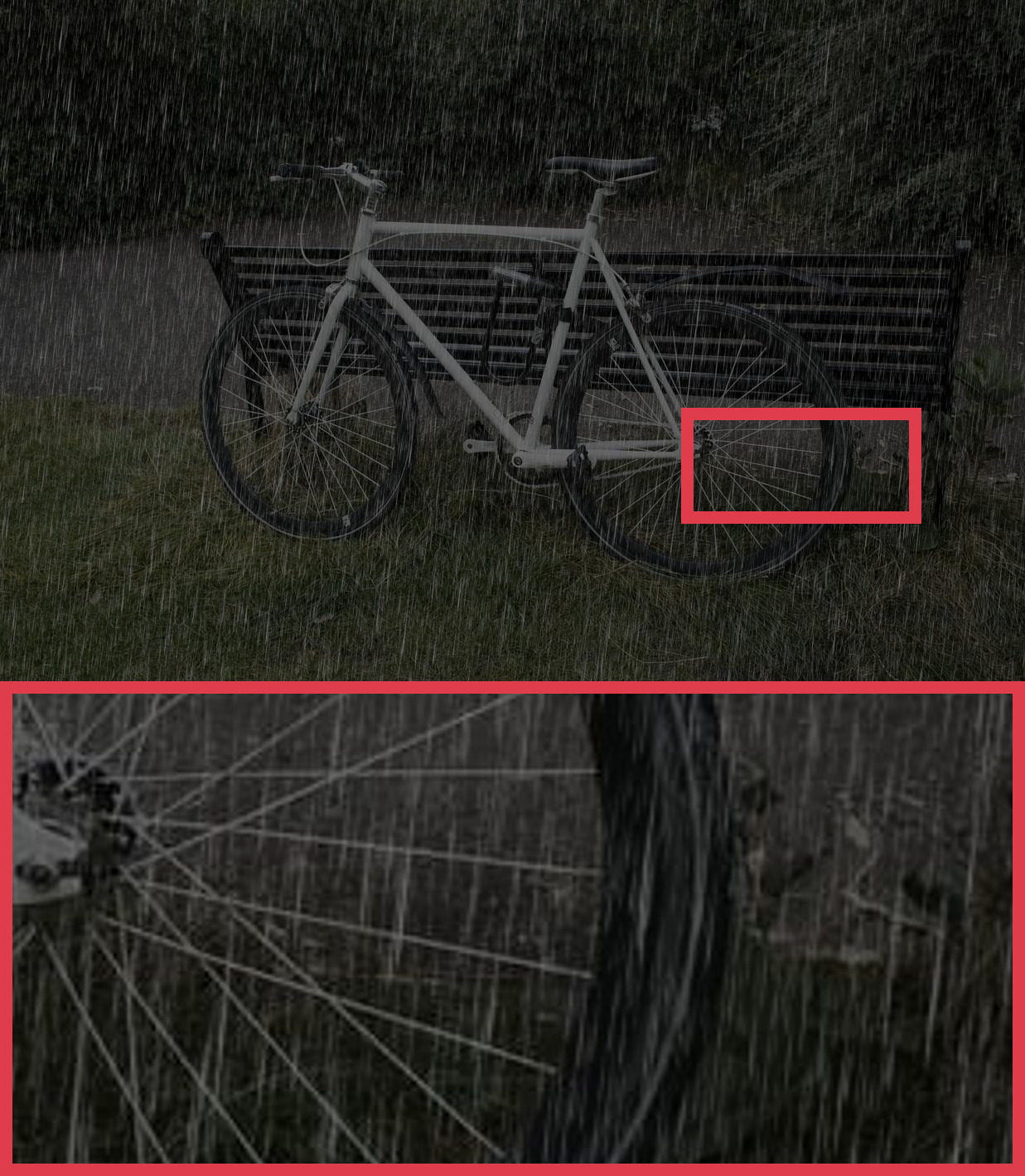} &
    \includegraphics[width=\linewidth]{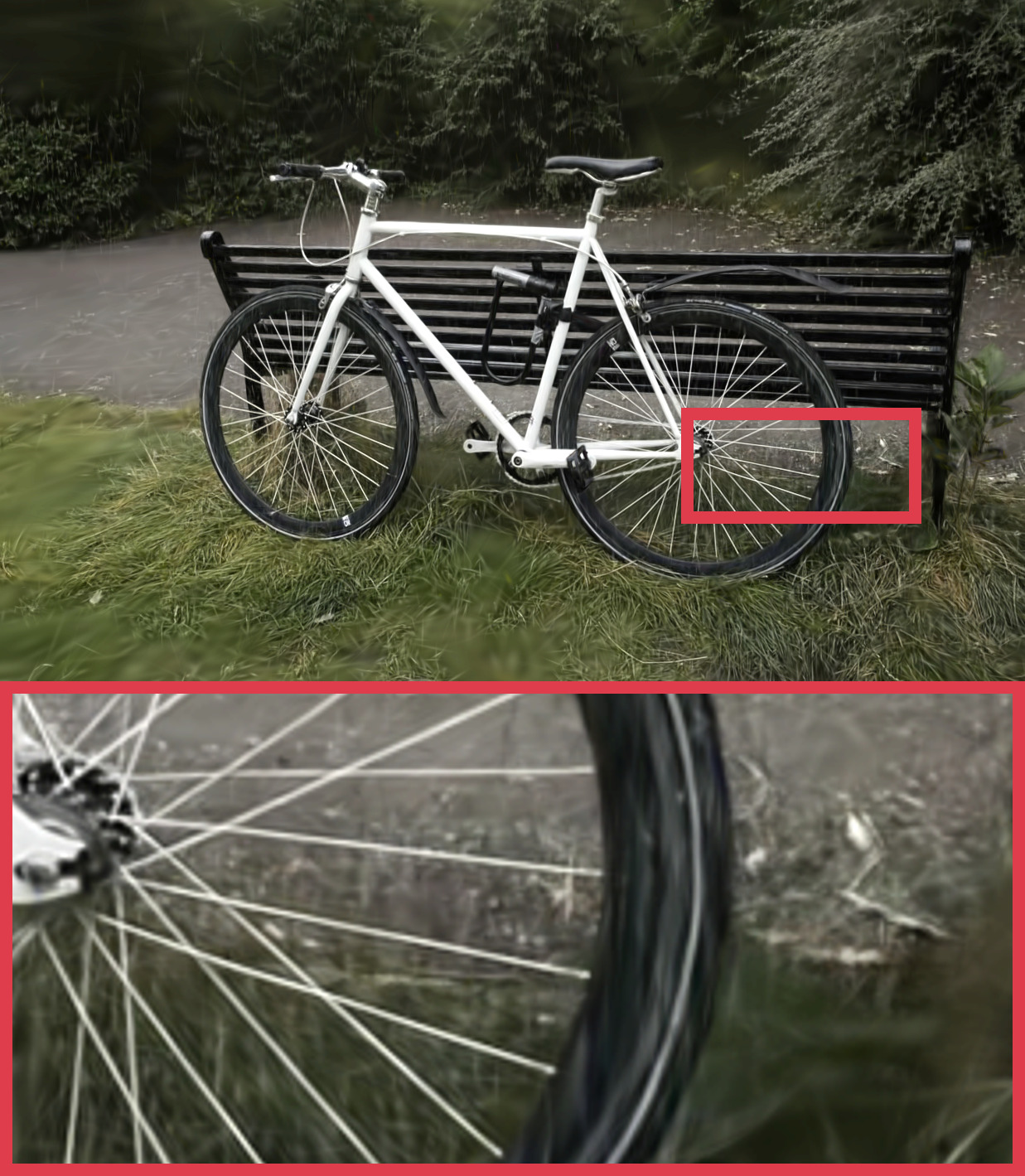} &
    \includegraphics[width=\linewidth]{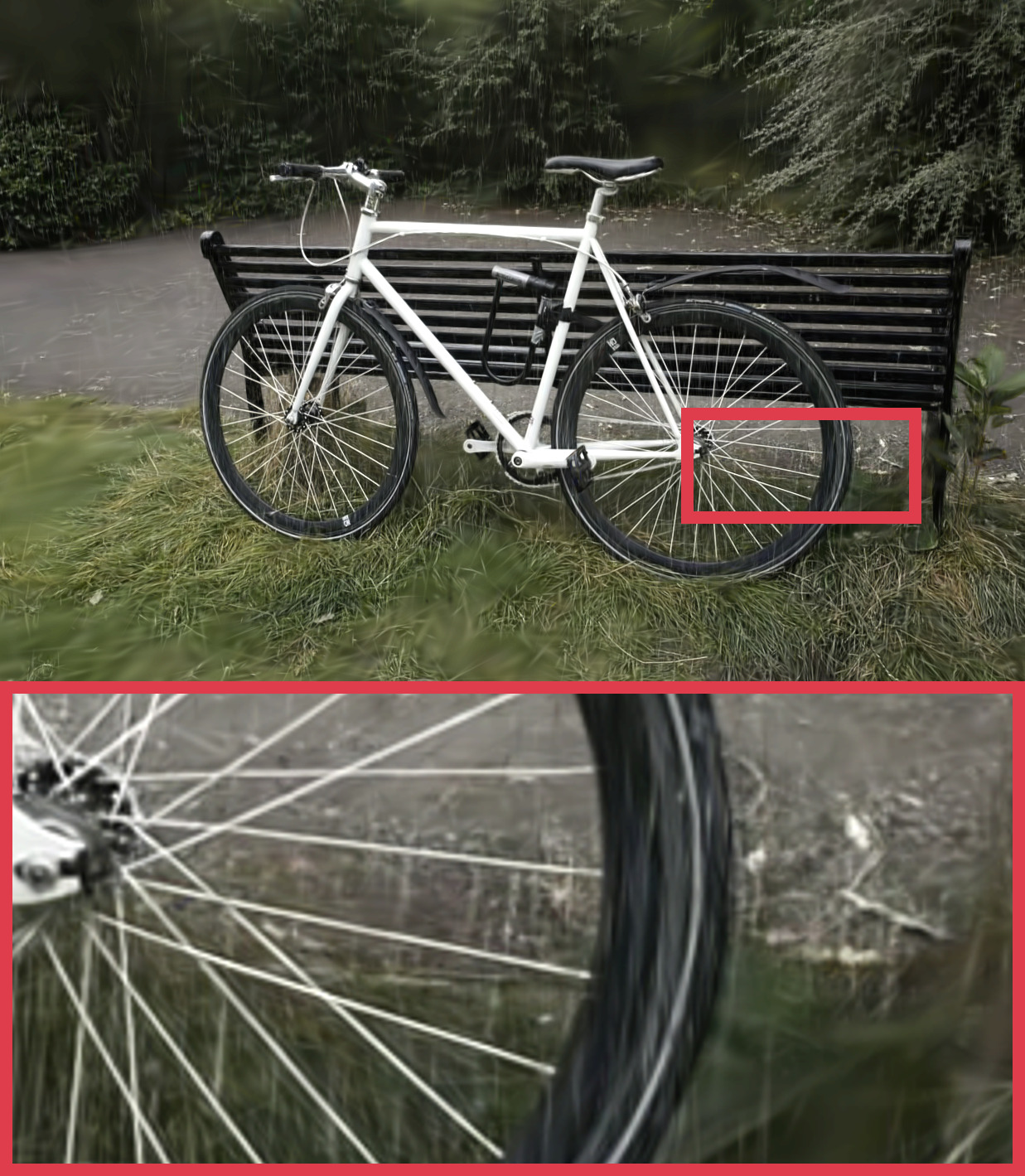} &
    \includegraphics[width=\linewidth]{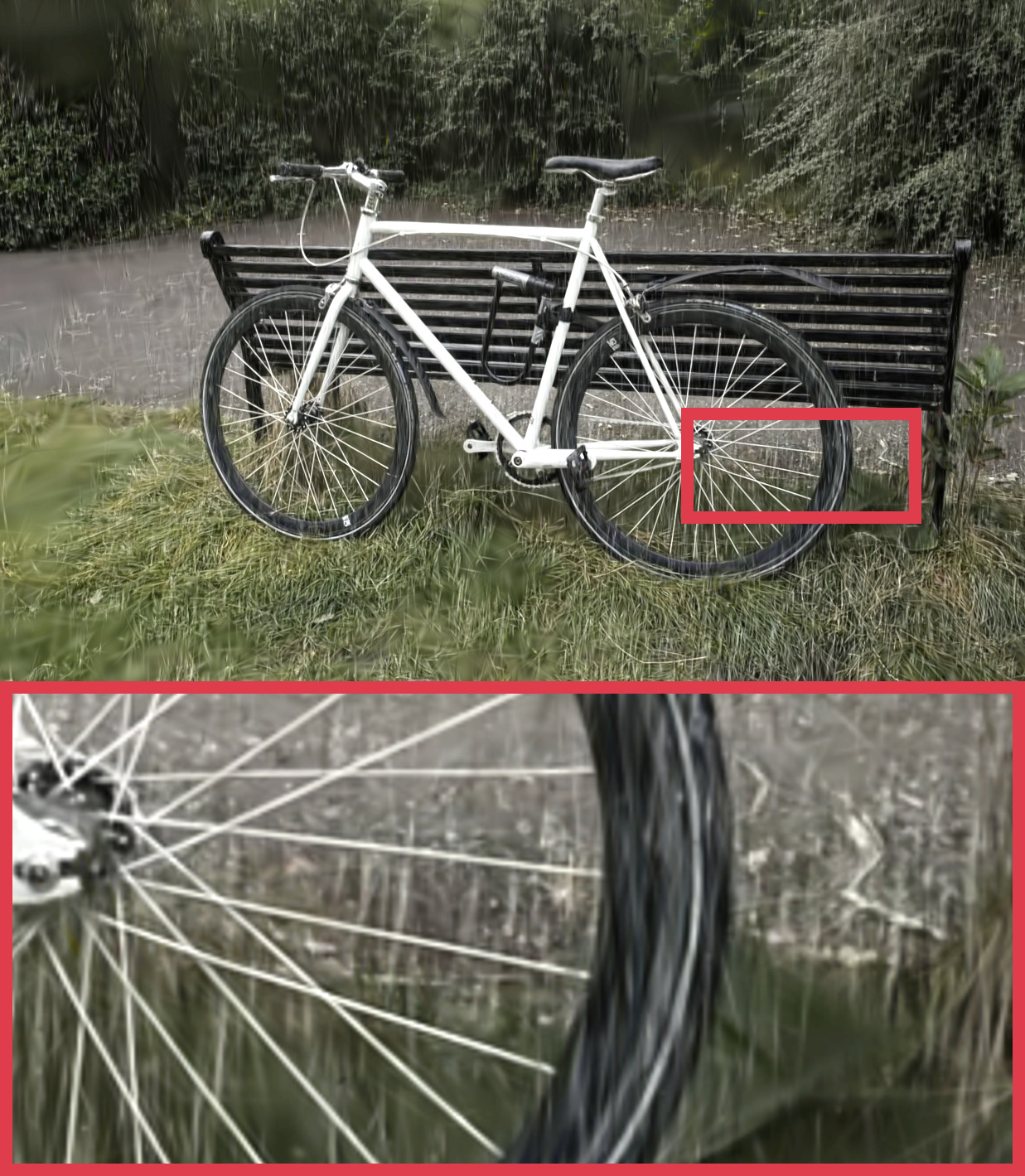} &
    \includegraphics[width=\linewidth]{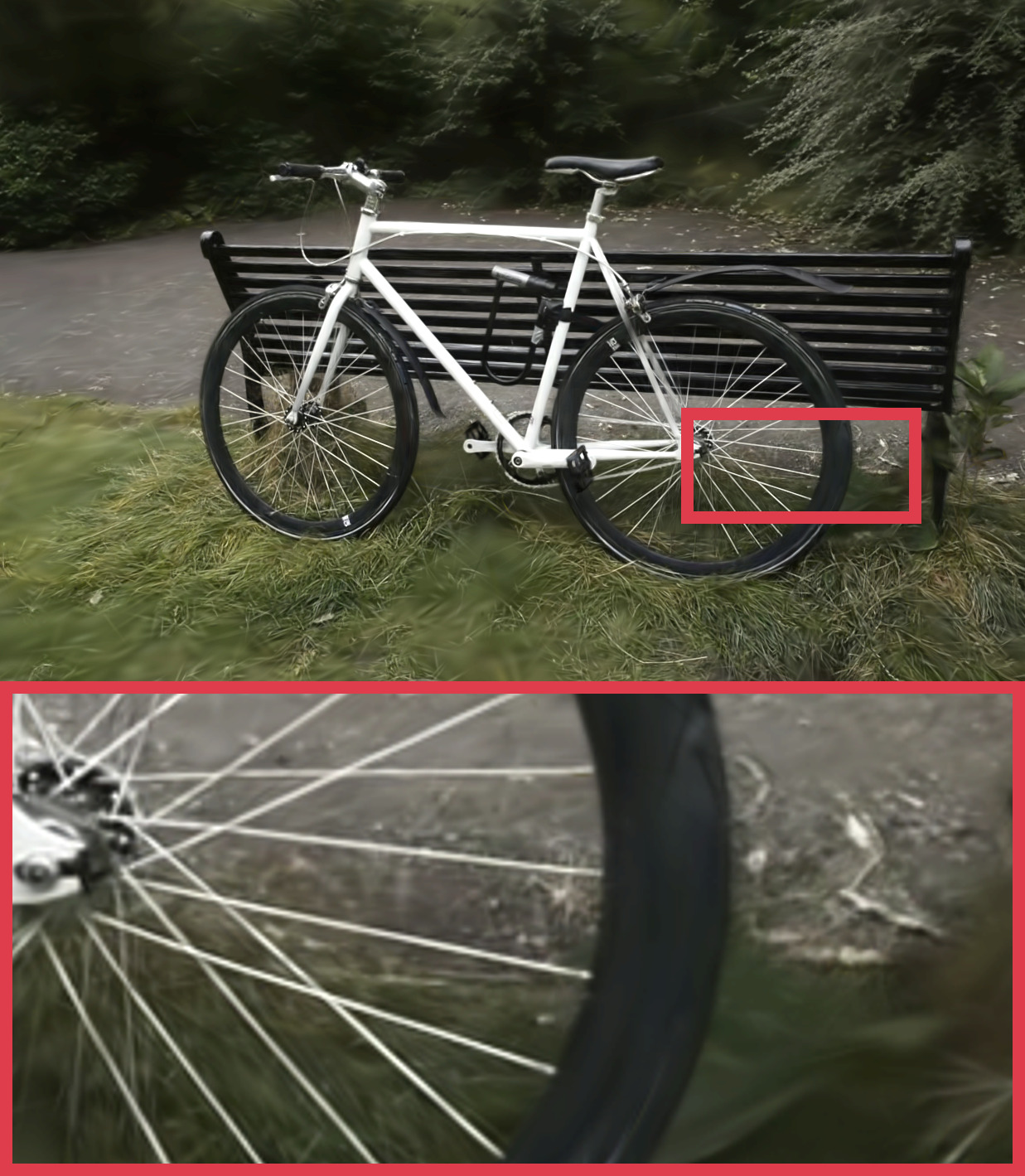} &
    \includegraphics[width=\linewidth]{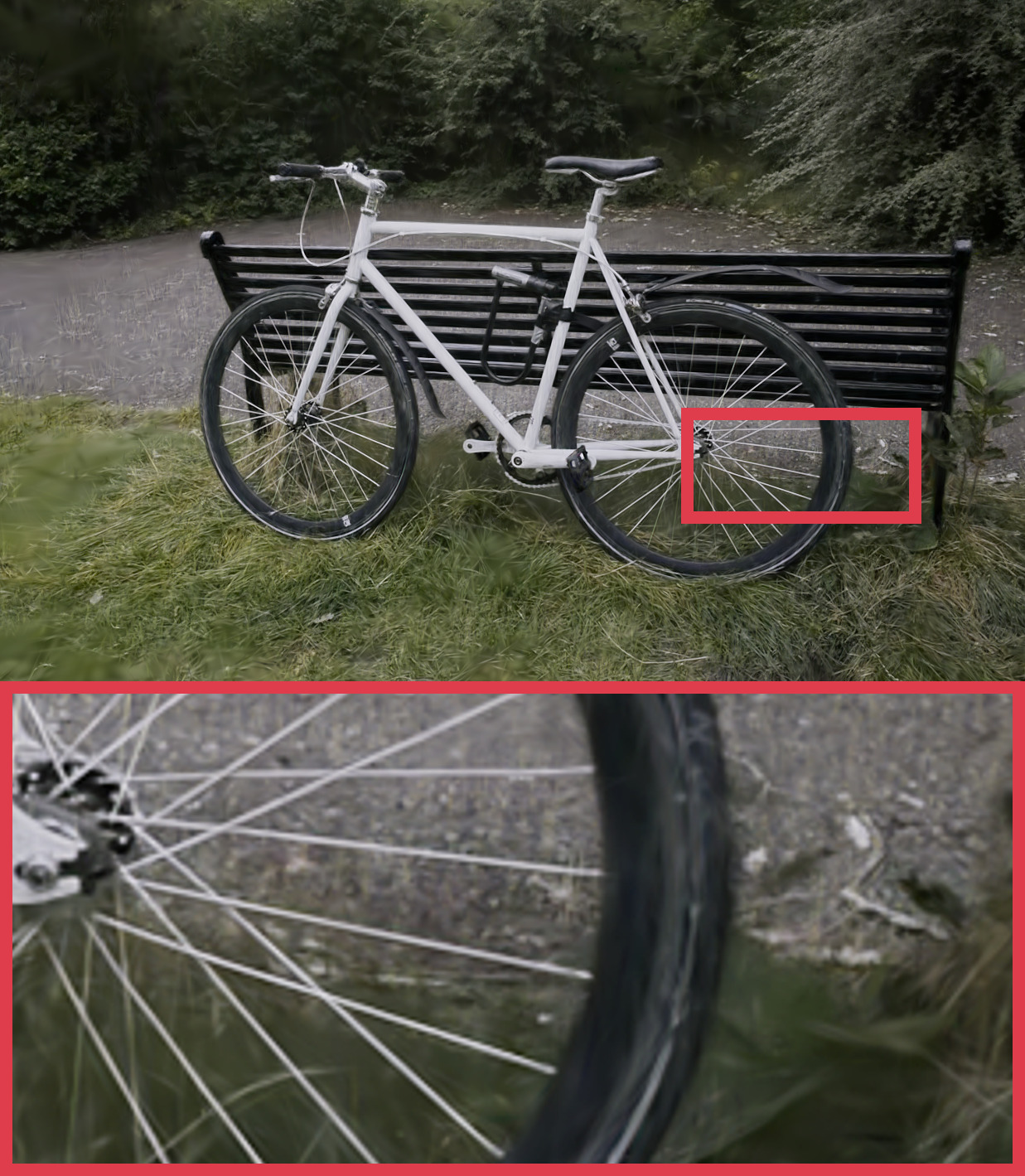} &
    \includegraphics[width=\linewidth]{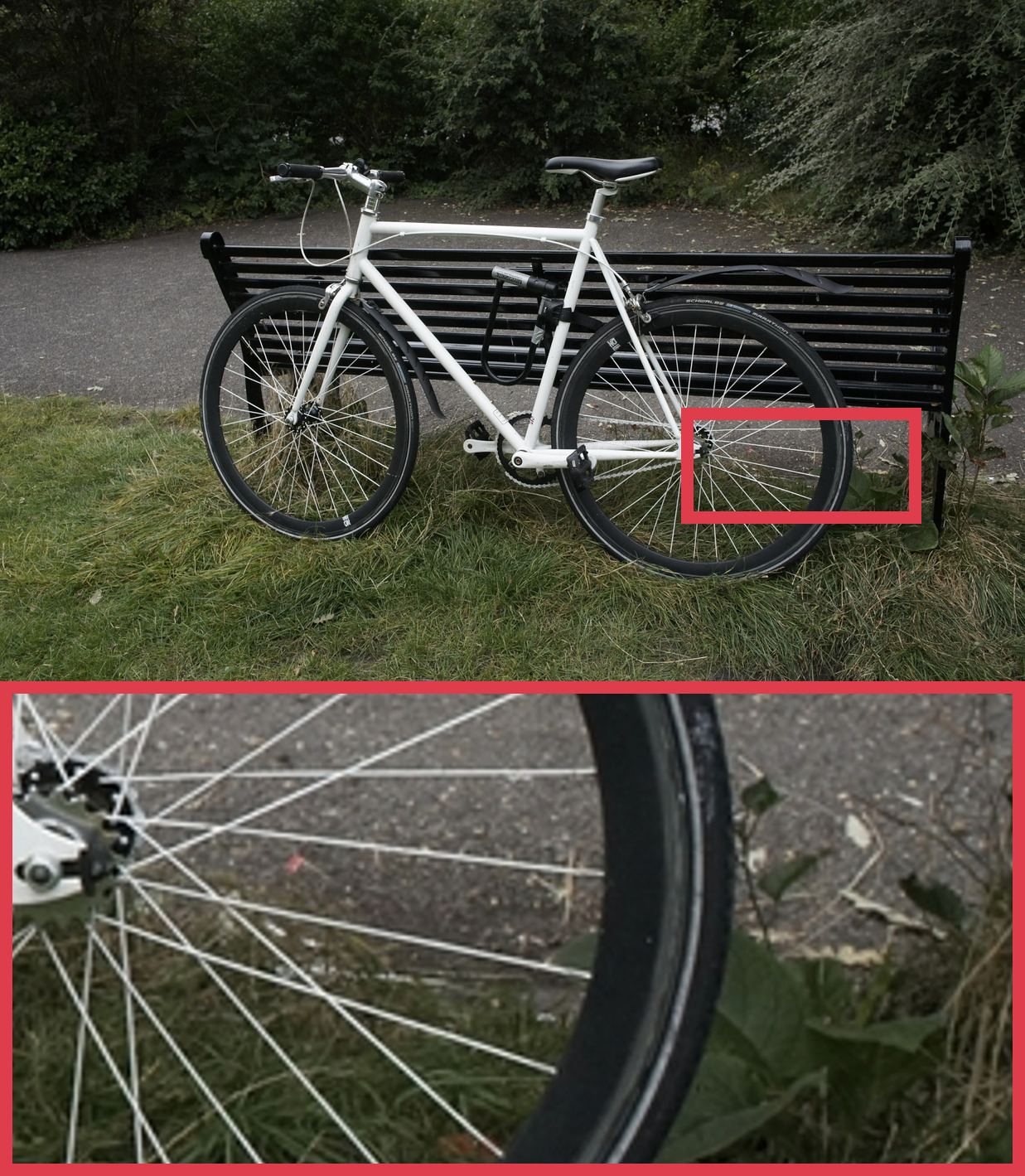}\\[10pt] 
    % 第一行：图像
    % 第二行：标注
    Input & 
    DRSformer* $\dagger$ & 
    NeRD-Rain*  $\dagger$ & 
    RainyScape $\dagger$ & 
    DerainGS  $\dagger$ & 
    Ours & 
    GT \\
    \end{tabular}
    \vspace{-2mm}
    \caption{Qualitative comparison of REVR-GSNet and other baselines on selected rainy scenes from the OmniRain3D dataset.}
    \label{data_show}
    \vspace{-3mm}
\end{figure*}

%----------------------------------------------------------------------------

% \vspace{-2mm}
\section{Experiments}
% \vspace{-1mm}
We first present the experimental settings, then verify the effectiveness of our method and conduct ablation studies. 

\subsection{Experiments Settings}
{\flushleft\textbf{Datasets and metrics}.}
To comprehensively evaluate our method, we conduct experiments on three datasets: synthetic raindrop scenes and real rainy scenes from HydroViews~\cite{liu2024deraings}, as well as rain streak scenes from our OmniRain3D, which includes dim scenarios.
We evaluate the quality of the images rendered from the reconstructed scenes using PSNR~\cite{huynh2008scope}, SSIM~\cite{wang2004image}, and LPIPS~\cite{zhang2018unreasonable}.

{\flushleft\textbf{Baselines}.}
For a comprehensive comparison, we first establish two baseline methods by employing the single image deraining method DRSformer~\cite{chen2023learning} and NeRD-Rain~\cite{chen2024bidirectional} as the preprocessing steps for the input images. Subsequently, we train 3DGS on the preprocessed images and name these two methods DRSformer* and NeRD-Rain*, respectively.
%
% 三个方法加上 端到端的前提
%
In addition, we compare our method with three specialized rainy scene reconstruction approaches: DerainNeRF~\cite{li2024derainnerf}, RainyScape~\cite{lyu2024rainyscape}, and DerainGS~\cite{liu2024deraings}. For fair comparison, all methods are trained for 30,000 iterations using their default parameter settings.
Furthermore, to ensure consistency under dim brightness conditions, all competing methods are preprocessed with a brightness enhancement module~\cite{guo2020zero} prior to training.

{\flushleft\textbf{Implementation details}.} 
The model is implemented using the PyTorch framework and trains end-to-end on an NVIDIA GeForce RTX 3090 GPU.
For RBE and GRE, the number of recursive step $n$ defaults to 4, while the recurrent step $l$ is set to 6.
Adam optimizers are used to train them with an initial learning rate $1e^{-3}$. 
For GPO, we employ the Adam optimizer with learning rates of $1.6e^{-4}$, $5e^{-4}$, and $2.5e^{-3}$ for 3DGS's means, scaling, and SH features.

\subsection{Experimental Results }
% \vspace{-1mm}
{\flushleft\textbf{Evaluations under varying brightness on rain streak scenes}.}
Table~\ref{Tab1_light} and Table~\ref{Tab1_dim_light} showcase the performance comparison of our proposed method against five baselines on four brightness scenes and four dim brightness scenes. 
Our method achieves the best performance on the majority of the scenes and metrics.
In challenging dim scenarios, such as francis and bicycle, REVR-GSNet benefits from the effective recursive brightness enhancement and GS-guided rain elimination and consistently delivers high-quality reconstructions, as shown in Figure~\ref{data_show}.

\begin{figure}[!t]
    \centering
    \small
    \setlength{\tabcolsep}{0.3pt} % 设置列间距
    \begin{tabular}{ 
        >{\centering\arraybackslash}m{0.09\textwidth} 
        >{\centering\arraybackslash}m{0.09\textwidth} 
        >{\centering\arraybackslash}m{0.09\textwidth} 
        >{\centering\arraybackslash}m{0.09\textwidth} 
        >{\centering\arraybackslash}m{0.09\textwidth} 
    }
    \includegraphics[width=\linewidth]{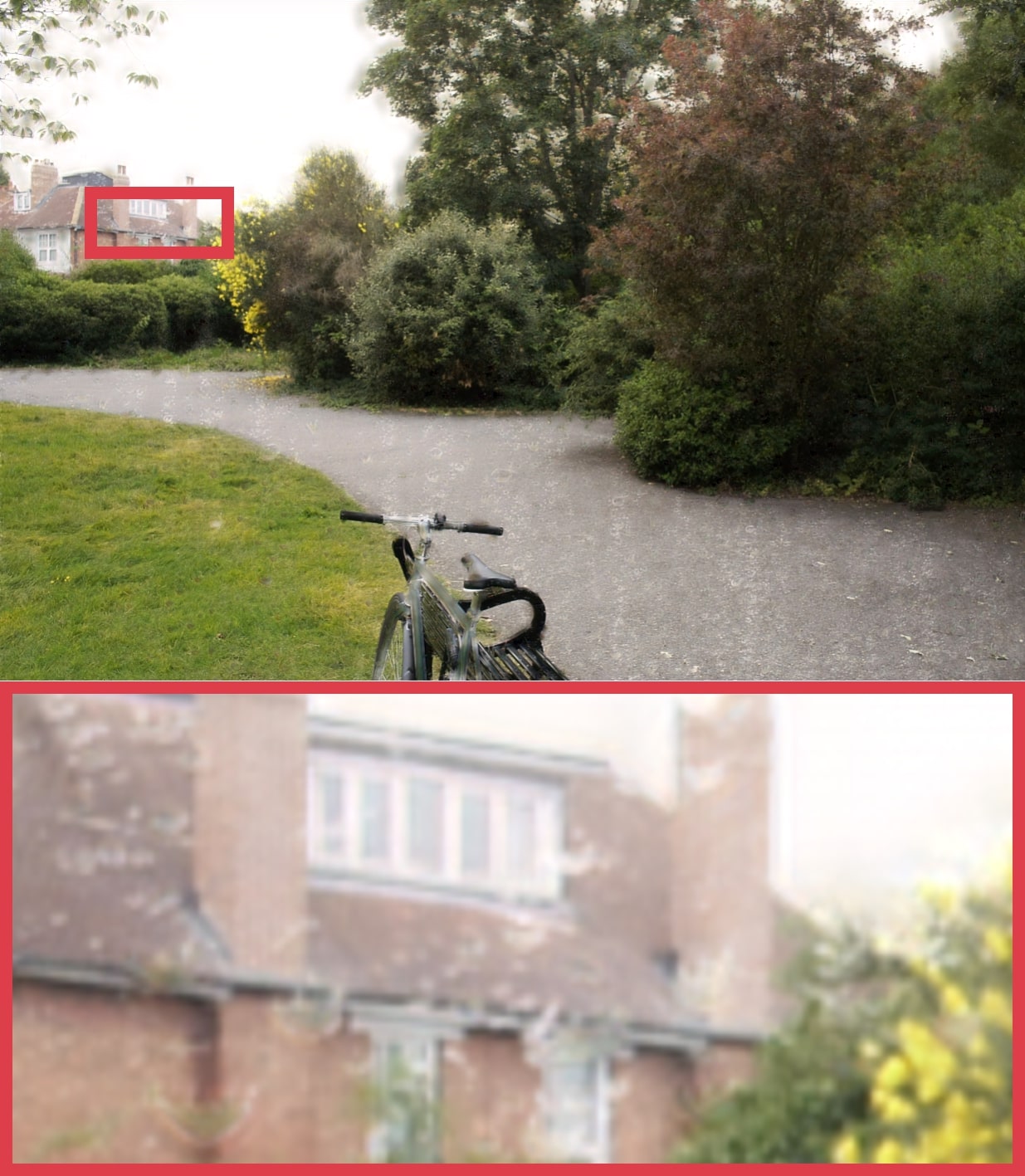} &
    \includegraphics[width=\linewidth]{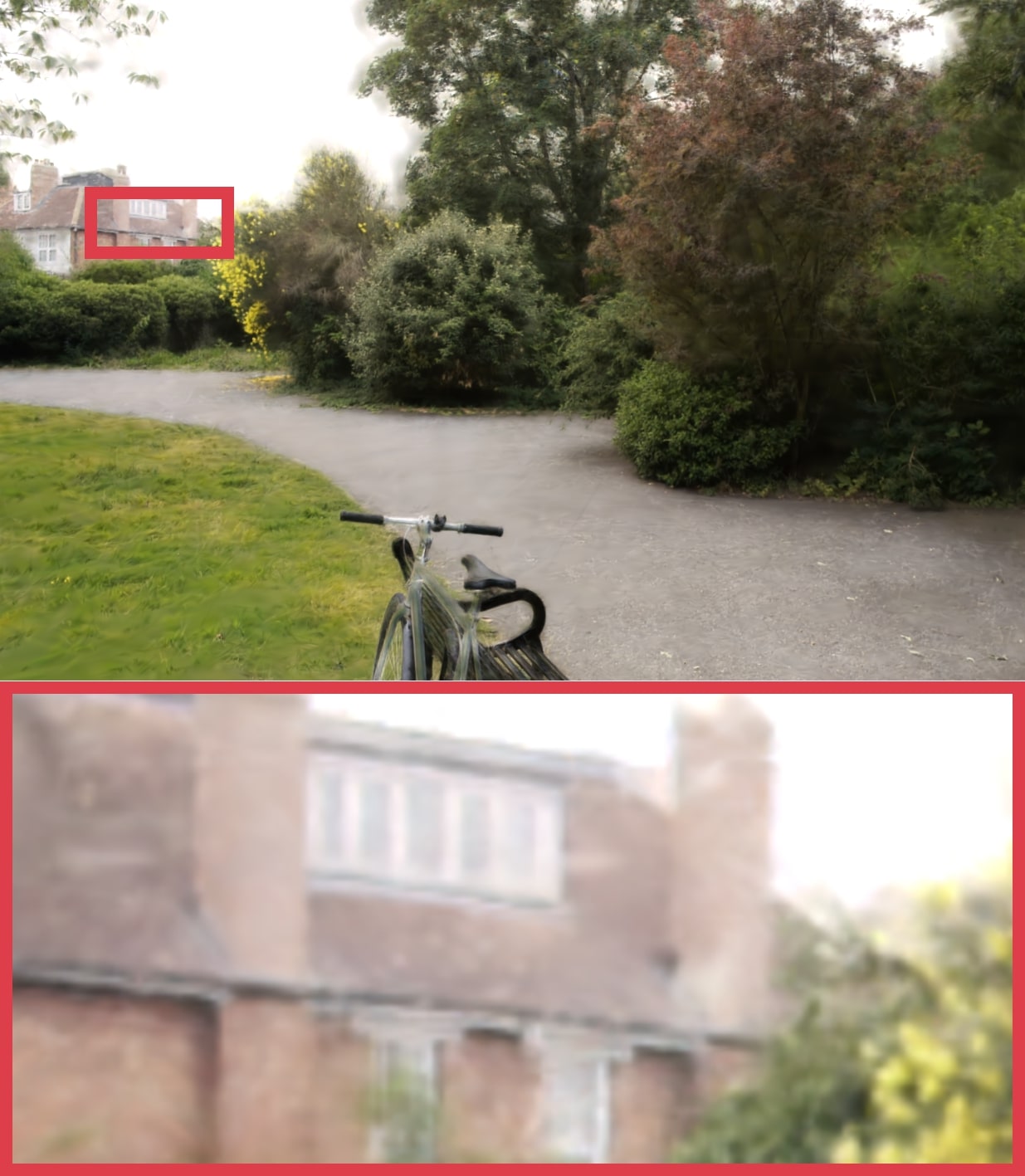} &
    \includegraphics[width=\linewidth]{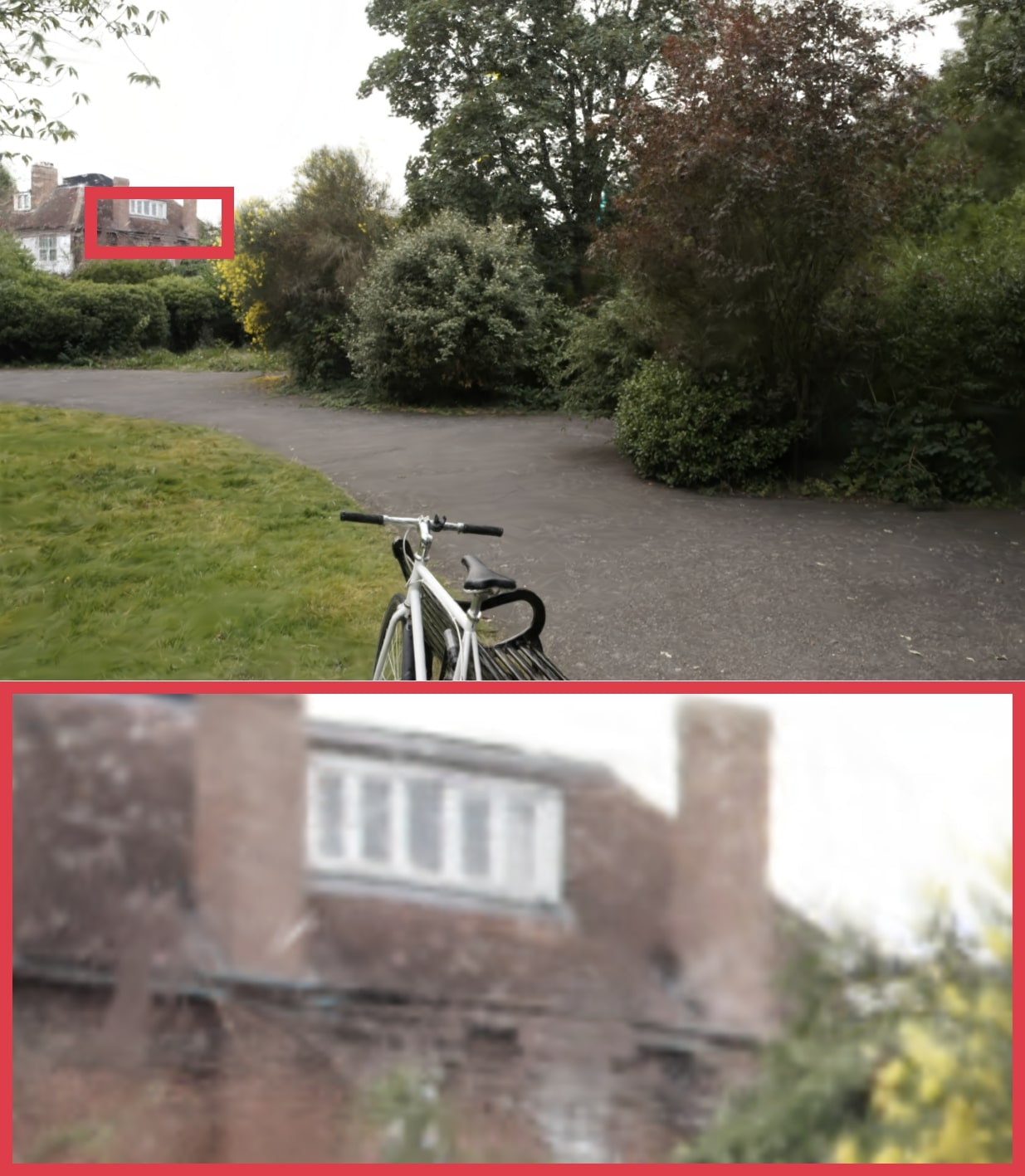} &
    \includegraphics[width=\linewidth]{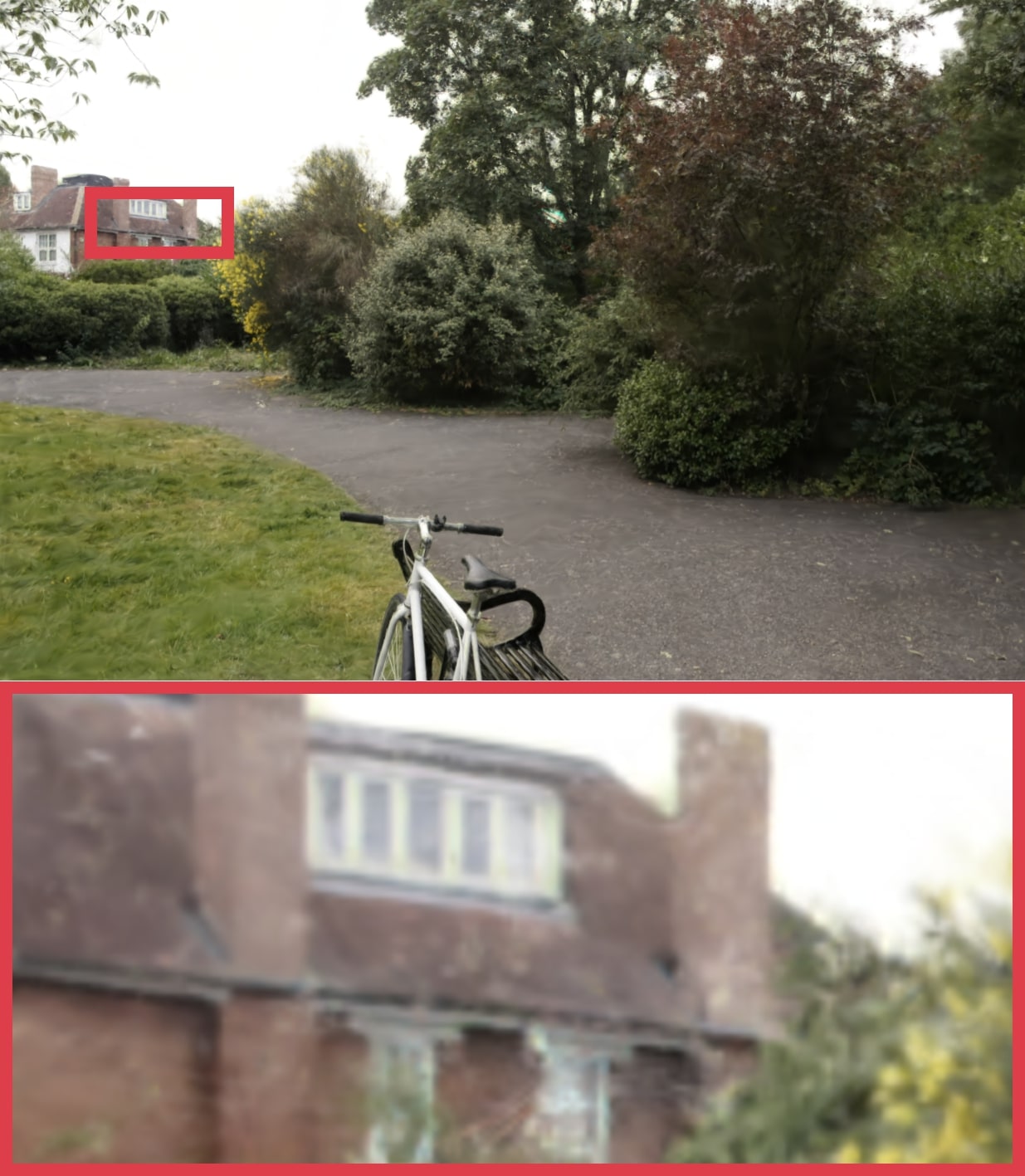} &
    \includegraphics[width=\linewidth]{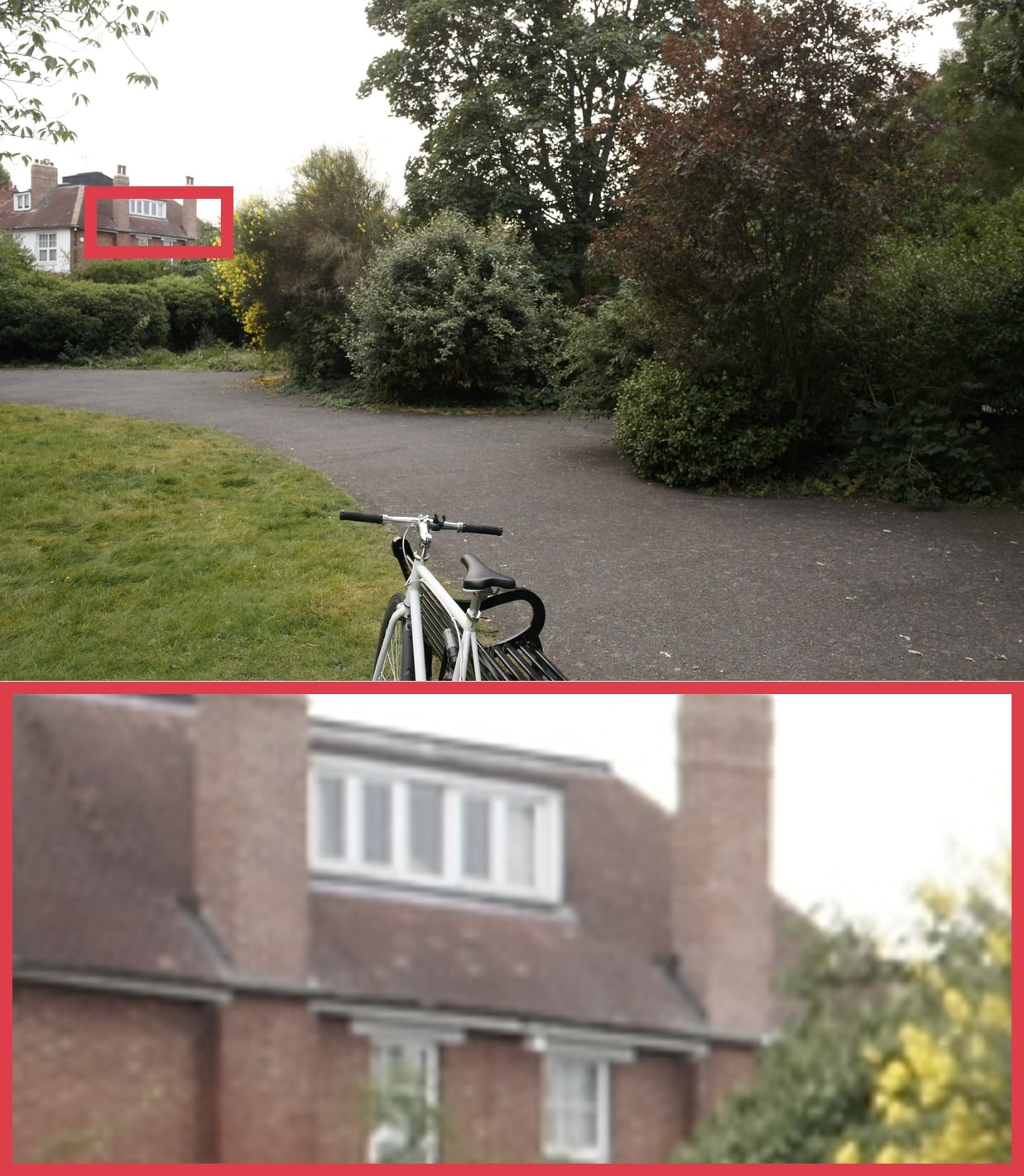}\\[10pt] 
    
    \includegraphics[width=\linewidth]{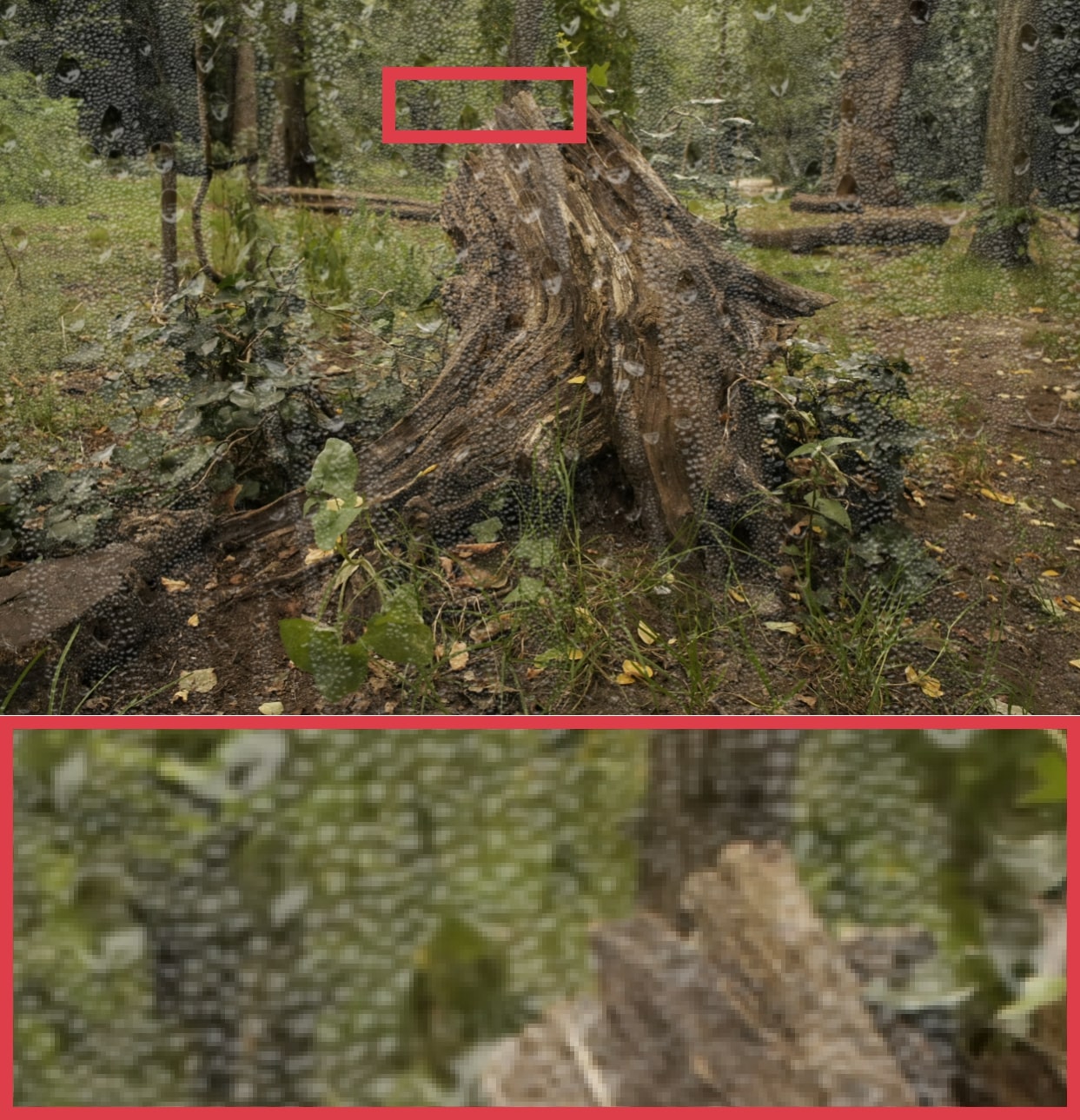} &
    \includegraphics[width=\linewidth]{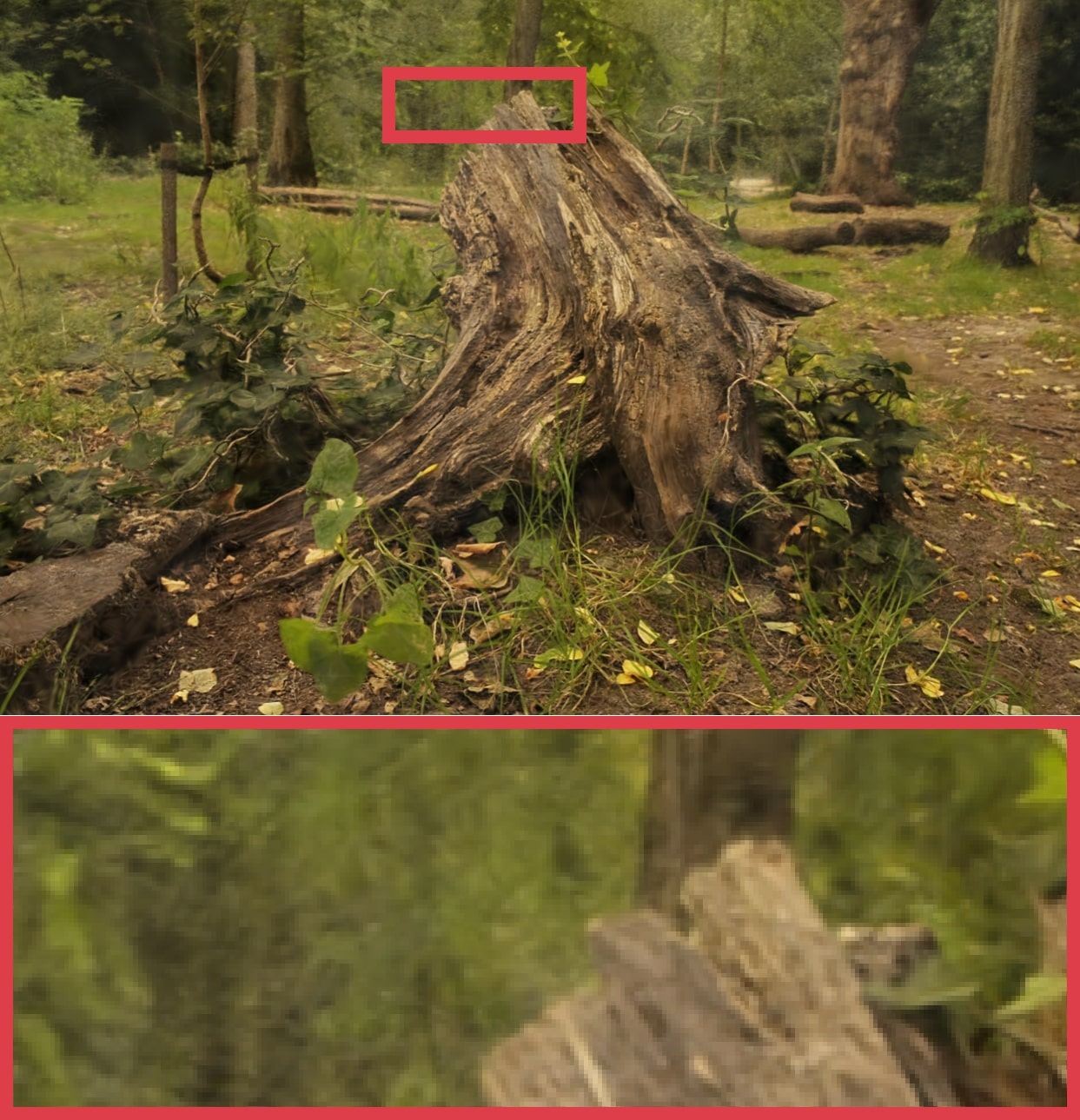} &
    \includegraphics[width=\linewidth]{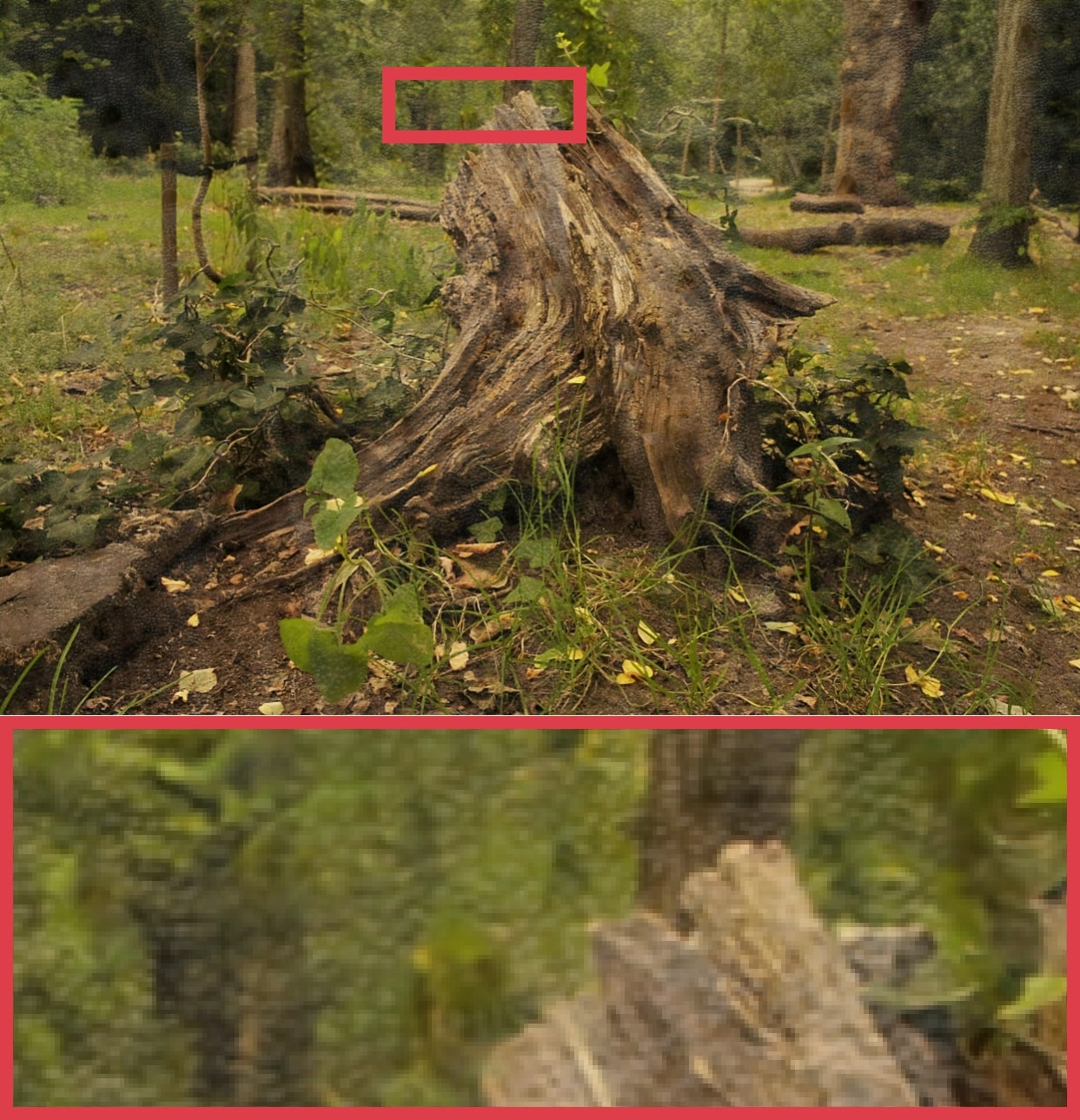} &
    \includegraphics[width=\linewidth]{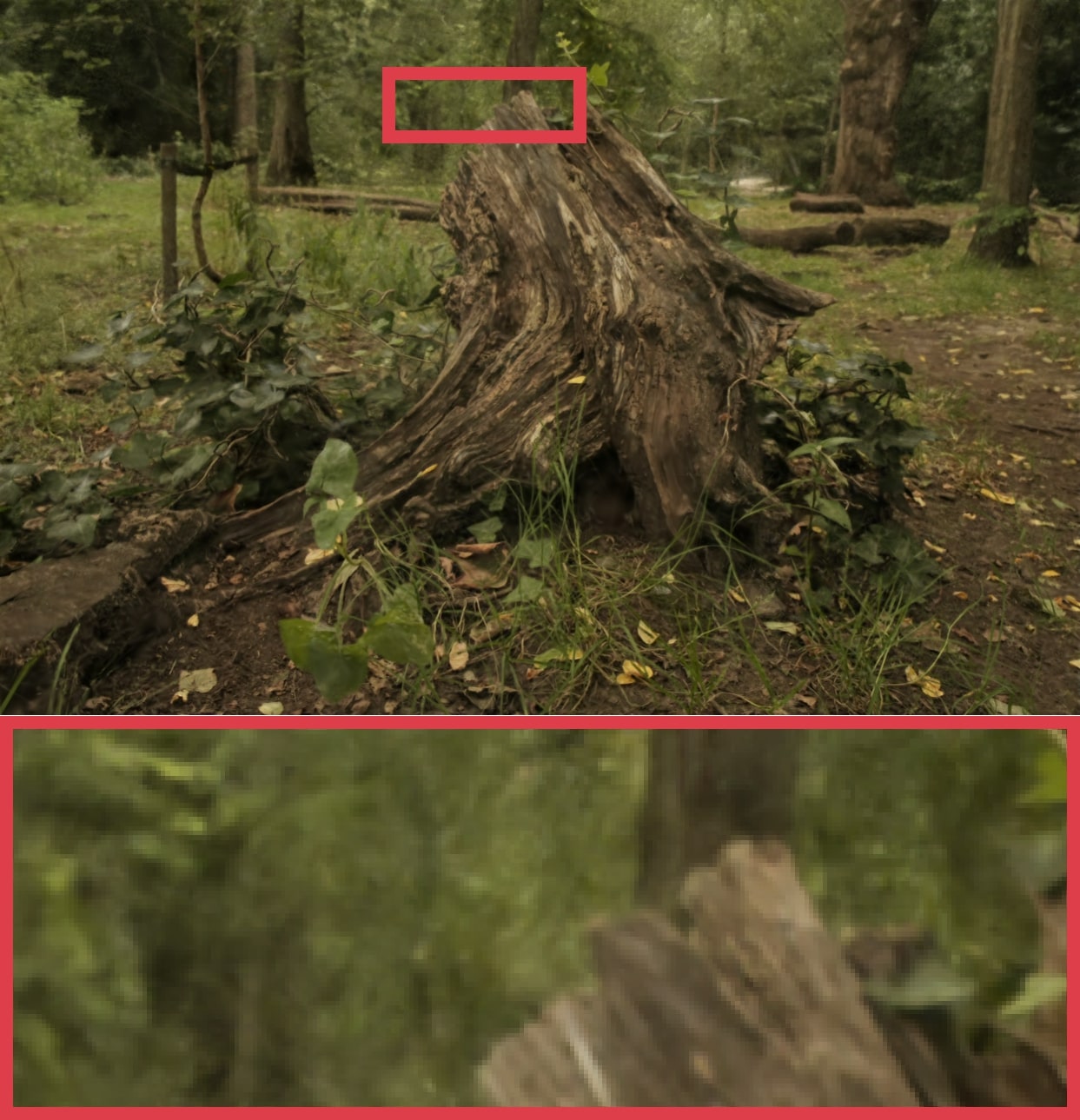} &
    \includegraphics[width=\linewidth]{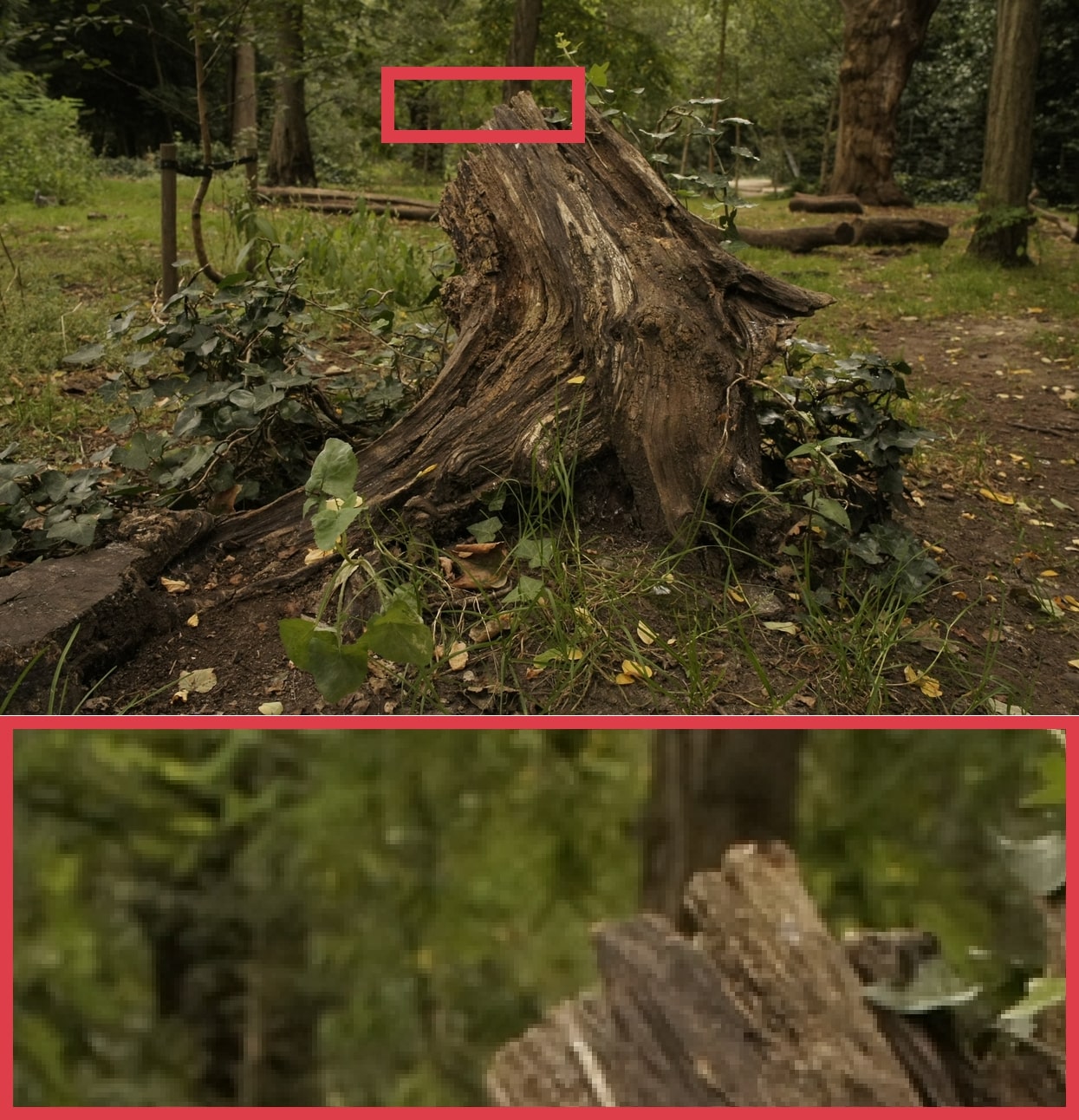}\\[10pt] 
    % 第一行：图像
    Input & 
    DRSformer*& 
    RainyScape & 
    Ours & 
    GT \\
    \end{tabular}
    \vspace{-2mm}
    \caption{The visualization compares the rendering quality of REVR-GSNet and baselines on raindrop scenes.}
    \vspace{-4mm}
    \label{data_show_raindrop}
\end{figure}

{\flushleft\textbf{Evaluations on raindrop scenes}.}
For scenes affected by raindrop, table~\ref{tab_raindrop} presents the quantitative comparison of our method with baselines. REVR-GSNet delivers even higher performance in scene reconstruction.
Figure~\ref{data_show_raindrop} visualize selected reconstruction outcomes, demonstrating that our method achieves rendering quality close to clean scenes.

%----------------------------------------------------------------------------
\begin{table*}[!t]
\renewcommand\arraystretch{1.1}
  \vspace{-2mm}
\resizebox{\linewidth}{!}{
\begin{tabular}{c|cccccccccccc|ccc}
\toprule % 顶部线
\multirow{2}{*}{Scene}& \multicolumn{3}{c}{DRSformer*} & \multicolumn{3}{c}{NeRD-Rain*}
& \multicolumn{3}{c}{DerainNeRF} & \multicolumn{3}{c}{RainyScape} & \multicolumn{3}{|c}{REVR-GSNet (Ours)}  \\
 \cmidrule(r){2-4}\cmidrule(lr){5-7}\cmidrule(lr){8-10}\cmidrule(lr){11-13}\cmidrule(lr){14-16}
& PSNR$\uparrow$ & SSIM$\uparrow$ &  LPIPS$\downarrow$
& PSNR$\uparrow$ & SSIM$\uparrow$ &  LPIPS$\downarrow$
& PSNR$\uparrow$ & SSIM$\uparrow$ &  LPIPS$\downarrow$
& PSNR$\uparrow$ & SSIM$\uparrow$ &  LPIPS$\downarrow$
& PSNR$\uparrow$ & SSIM$\uparrow$ &  LPIPS$\downarrow$\\
\midrule % 中部线
\multicolumn{1}{c|}{Bicycle} &14.68 & \underline{0.516} &0.409 &14.44&0.493& 0.491 &12.72&0.367 &0.595 & \underline{15.08}& 0.510 & \textbf{0.399} & \textbf{16.33} & \textbf{0.553} & \underline{0.400}
\\
\multicolumn{1}{c|}{Garden} & \underline{9.23} & \underline{0.512} & \underline{0.391} &8.40 &0.494 &0.399 &8.52 &0.354 &0.629 &8.49 &0.493 &0.403  & \textbf{10.01} & \textbf{0.519} & \textbf{0.339}
\\
\multicolumn{1}{c|}{Stump} &18.23 &0.601 & 0.303 &19.79 &0.503 &0.336 &18.68 &0.408 &0.596 & \underline{22.59} & \underline{0.639} & \underline{0.284} & \textbf{22.61} & \textbf{0.650} & \textbf{0.258}
\\
\bottomrule
\end{tabular}
}
\caption{Performance comparison of different methods on raindrop scenes from HydroViews dataset.
For each scene, we evaluate all methods across three patterns of different rain effects and report the averaged results.}
\label{tab_raindrop}
\vspace{-3mm}
\end{table*}
%----------------------------------------------------------------------------

%%%%%%%%%%%%%%%%%%%%%%%%%%%%%%%%%%%%%%%%%%%%%%%%%%%%%%%%%%%%%

\begin{figure}[!t]
    \centering
    \small
    \setlength{\tabcolsep}{0.2pt} % 设置列间距
    \begin{tabular}{ 
        >{\centering\arraybackslash}m{0.095\textwidth} 
        >{\centering\arraybackslash}m{0.095\textwidth} 
        >{\centering\arraybackslash}m{0.095\textwidth} 
        >{\centering\arraybackslash}m{0.095\textwidth} 
        >{\centering\arraybackslash}m{0.095\textwidth} 
    }

    \includegraphics[width=\linewidth]{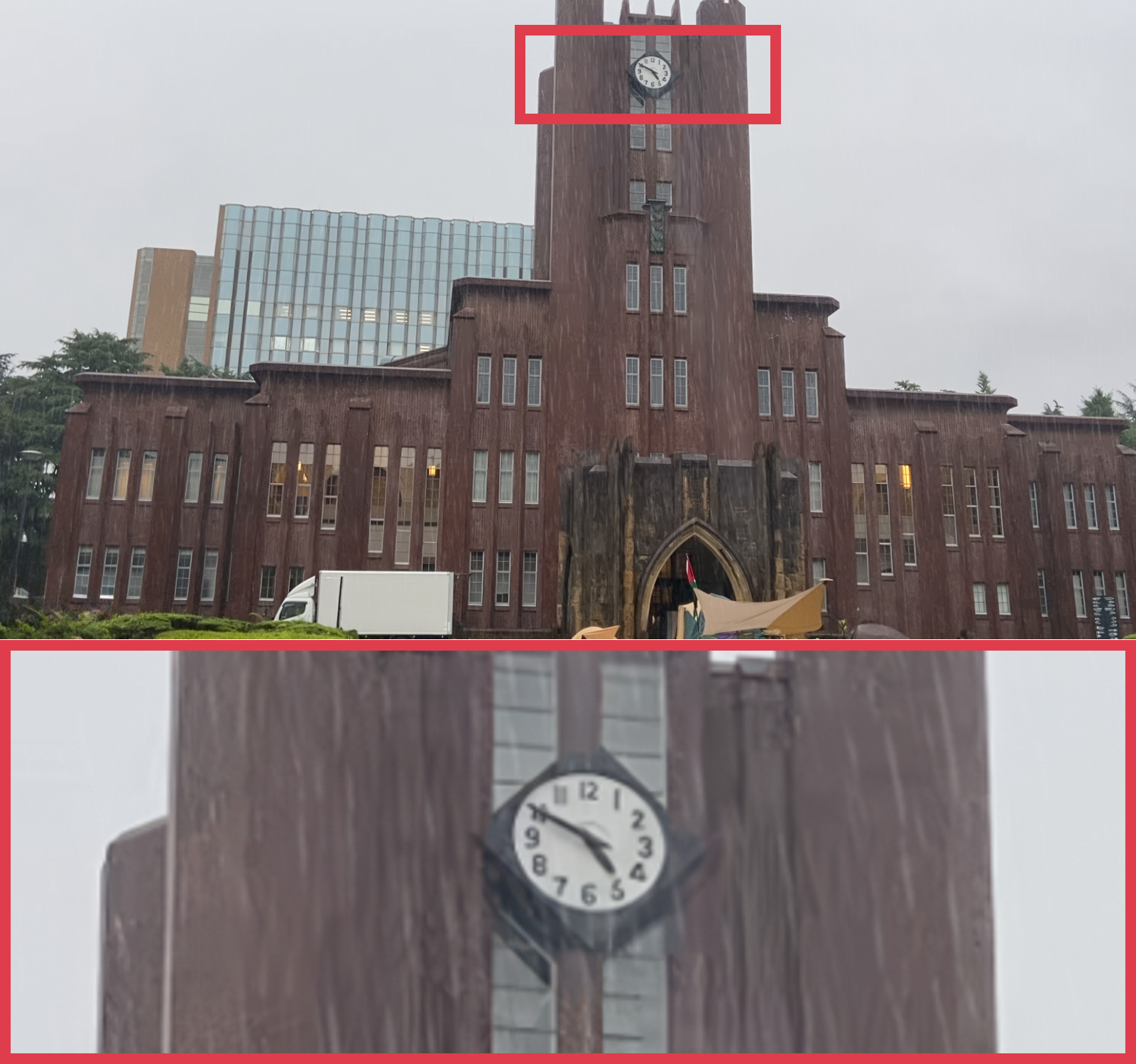} &
    \includegraphics[width=\linewidth]{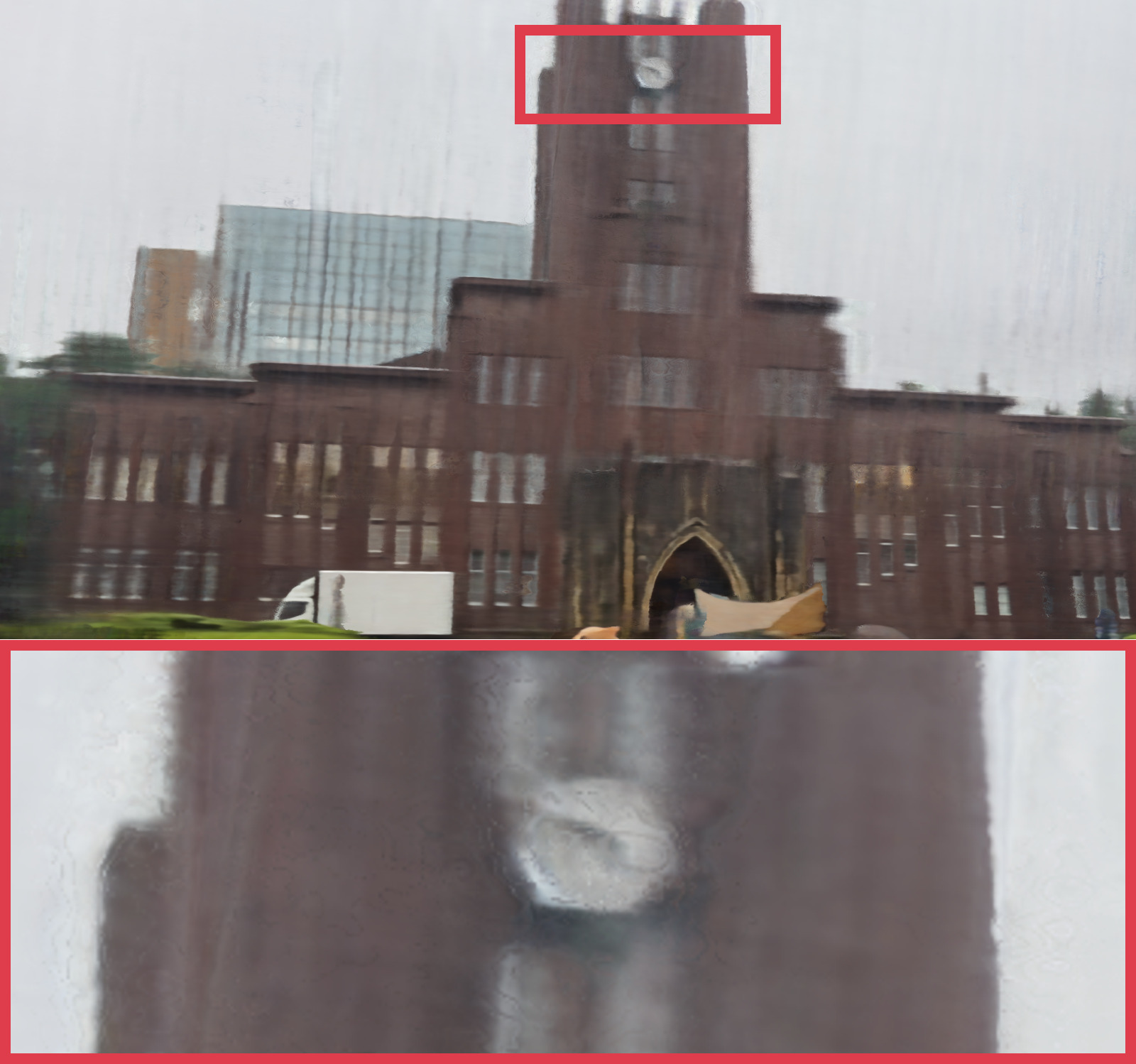} &
    \includegraphics[width=\linewidth]{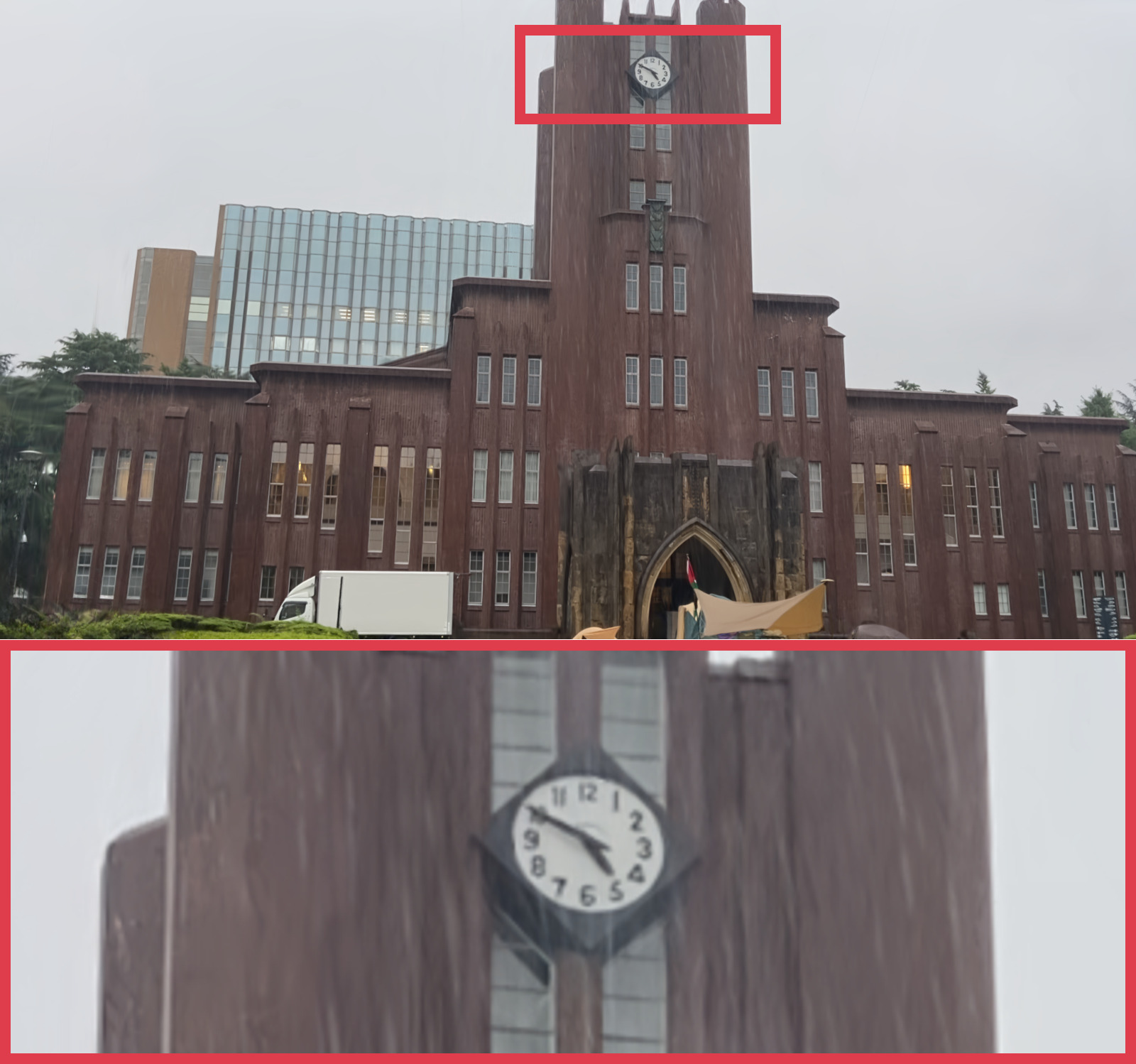} &
    \includegraphics[width=\linewidth]{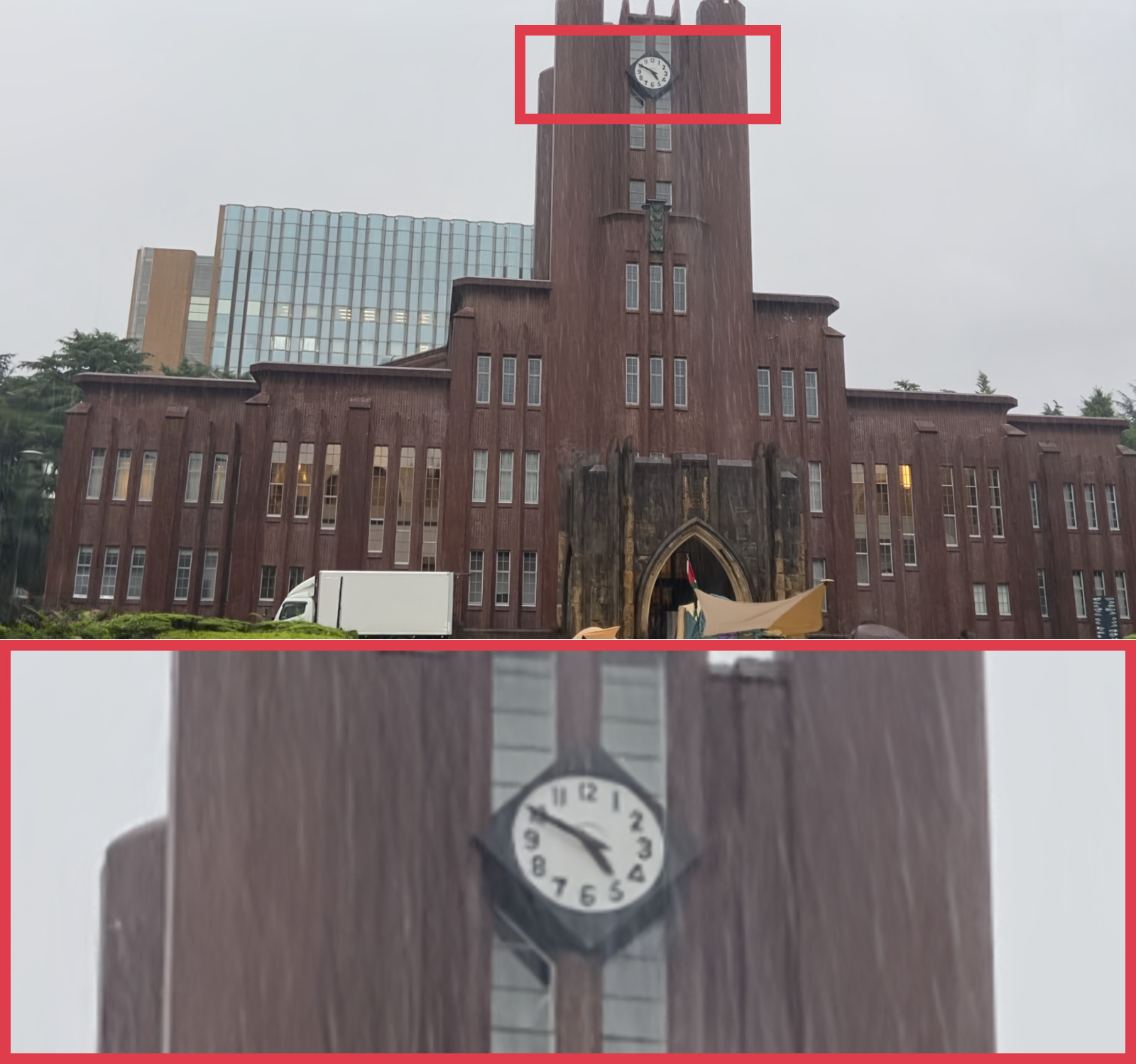} &
    \includegraphics[width=\linewidth]{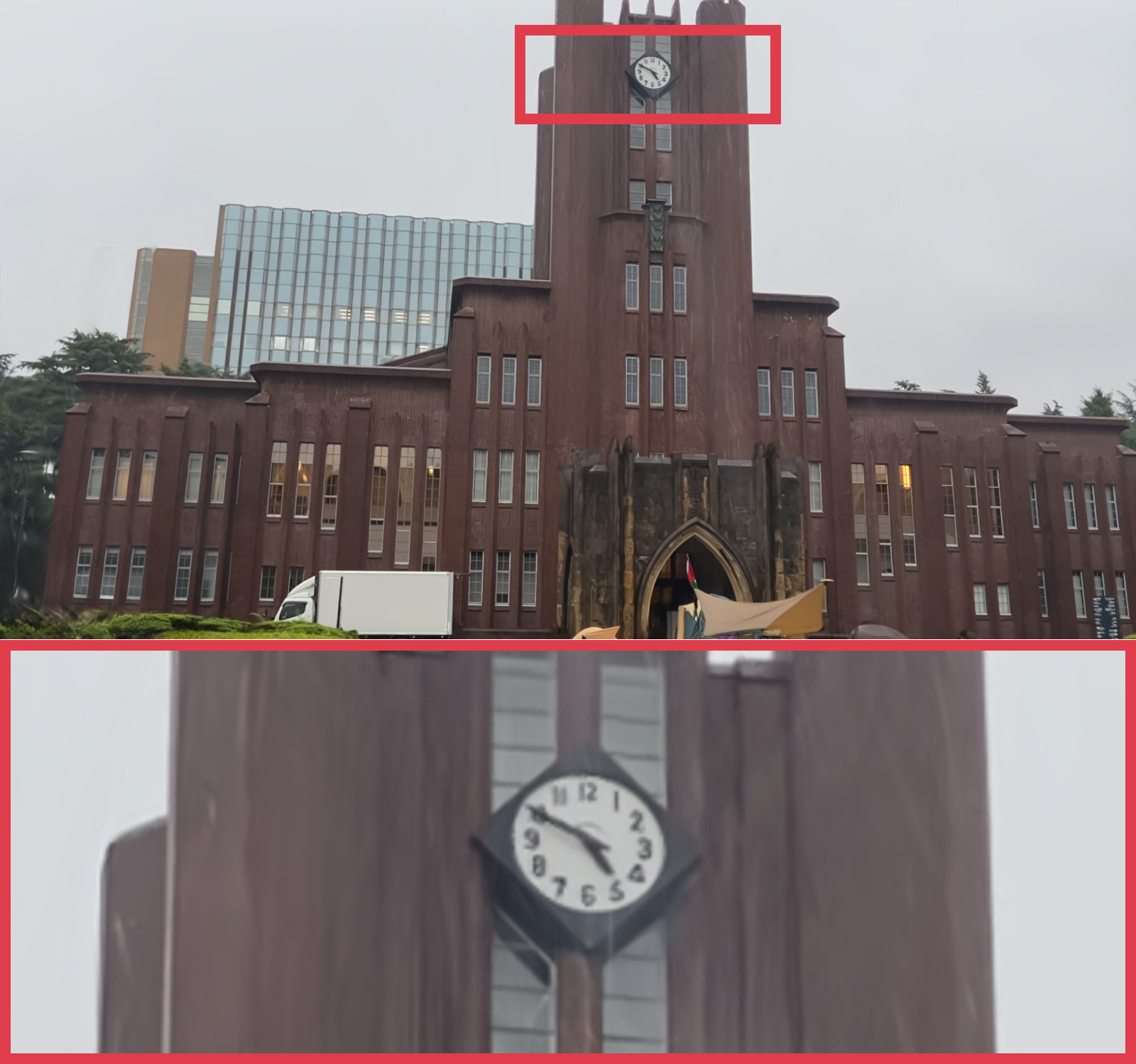}\\[10pt] 
    
    \includegraphics[width=\linewidth]{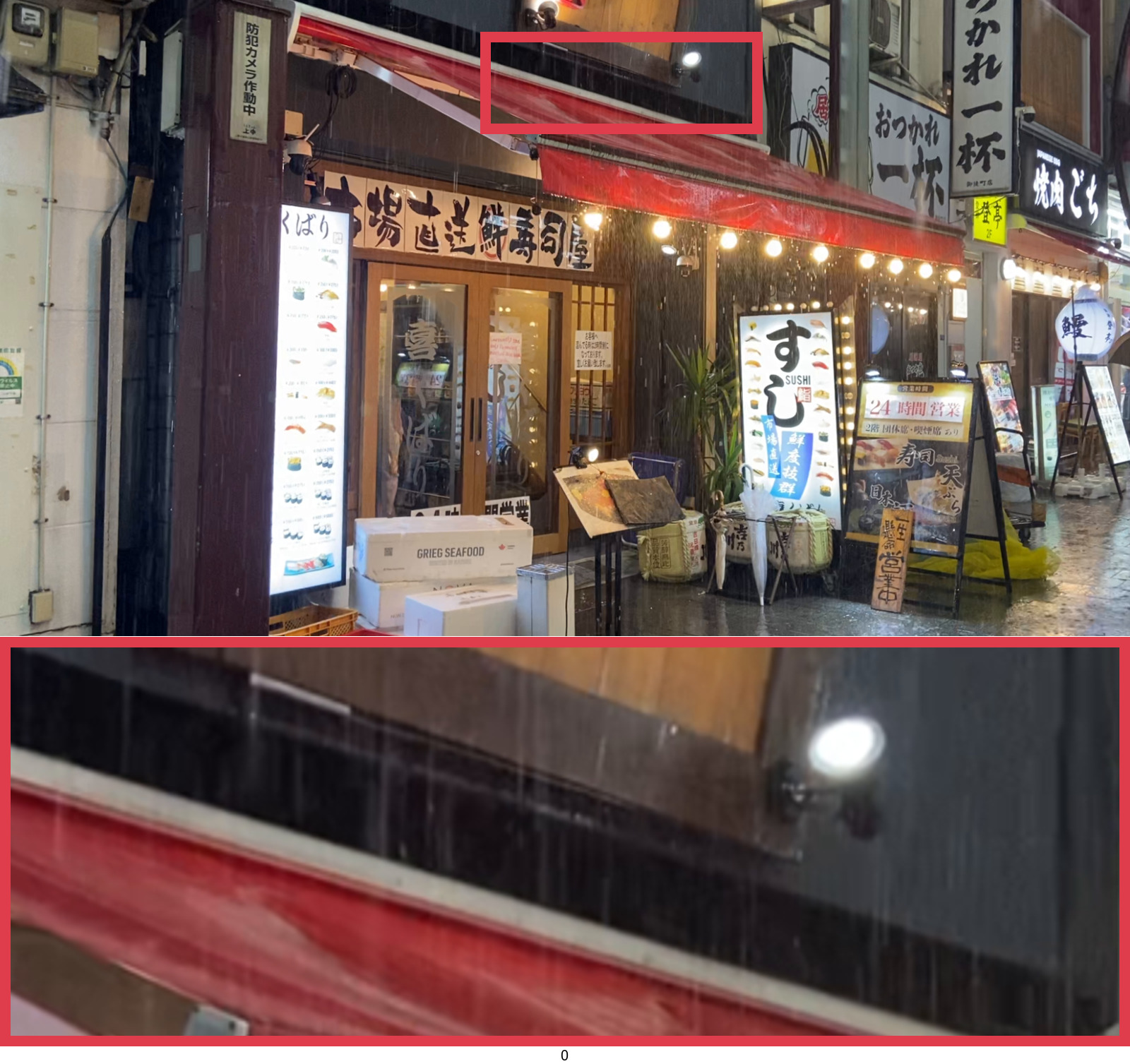} &
    \includegraphics[width=\linewidth]{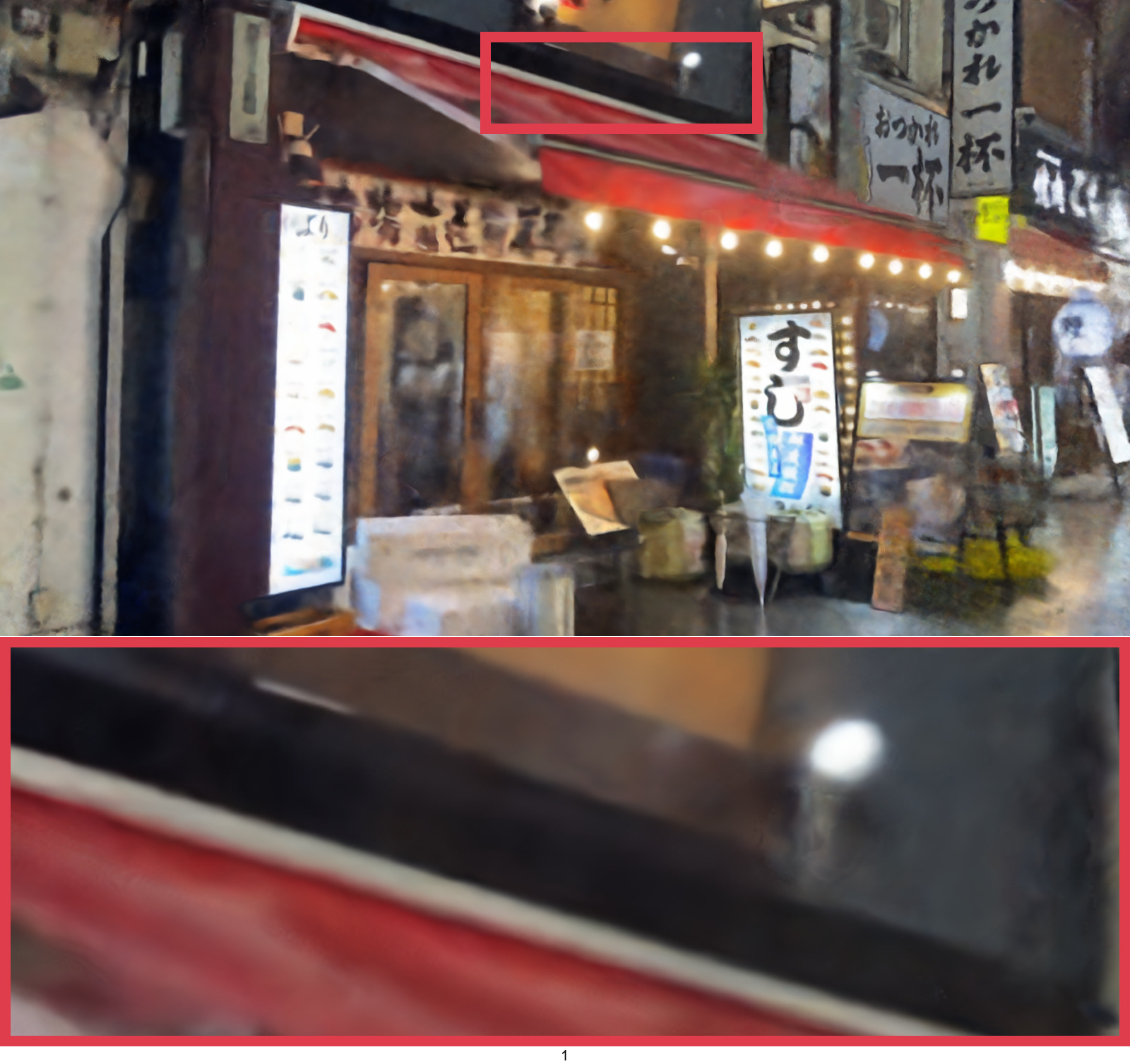} &
    \includegraphics[width=\linewidth]{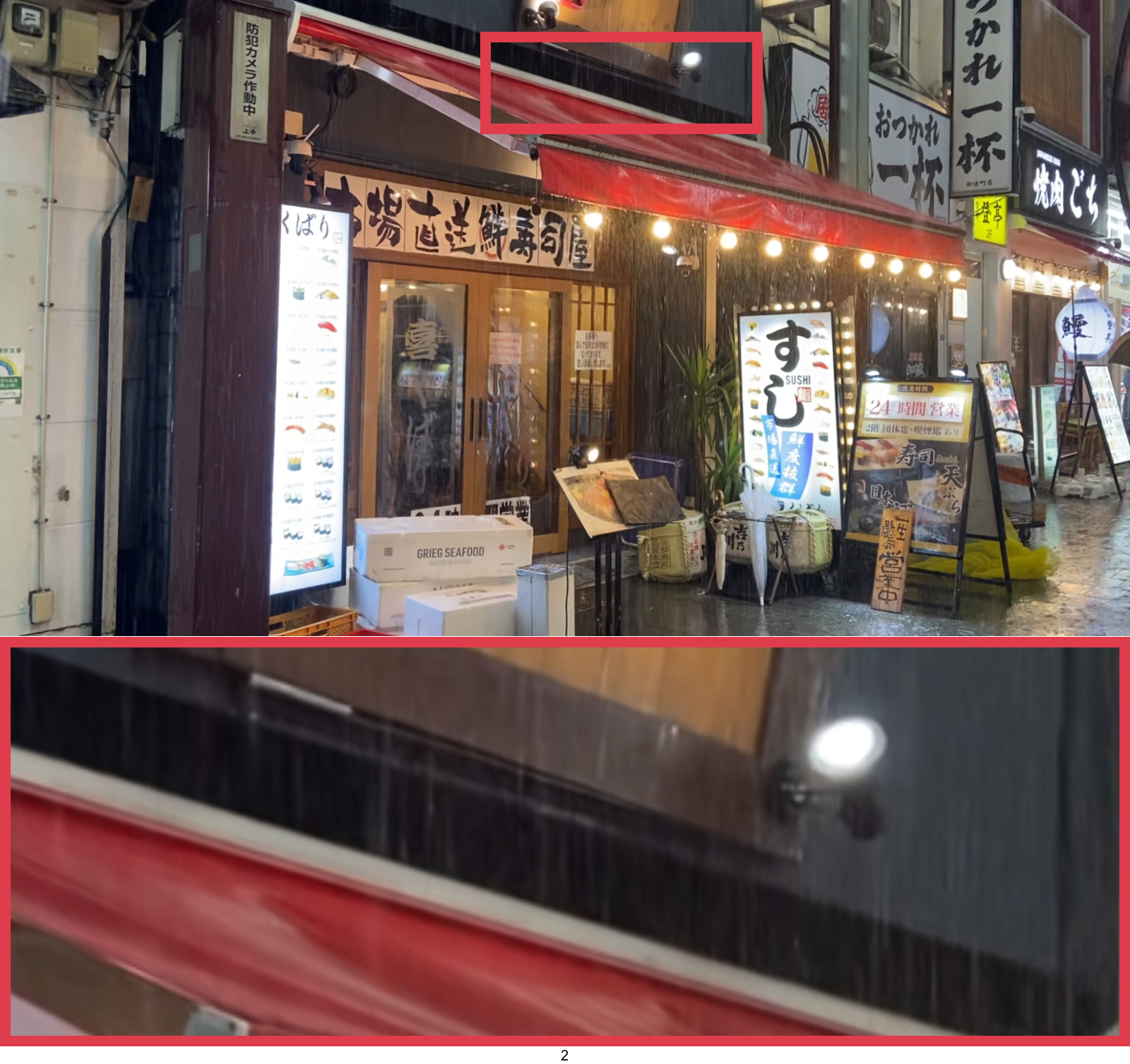} &
    \includegraphics[width=\linewidth]{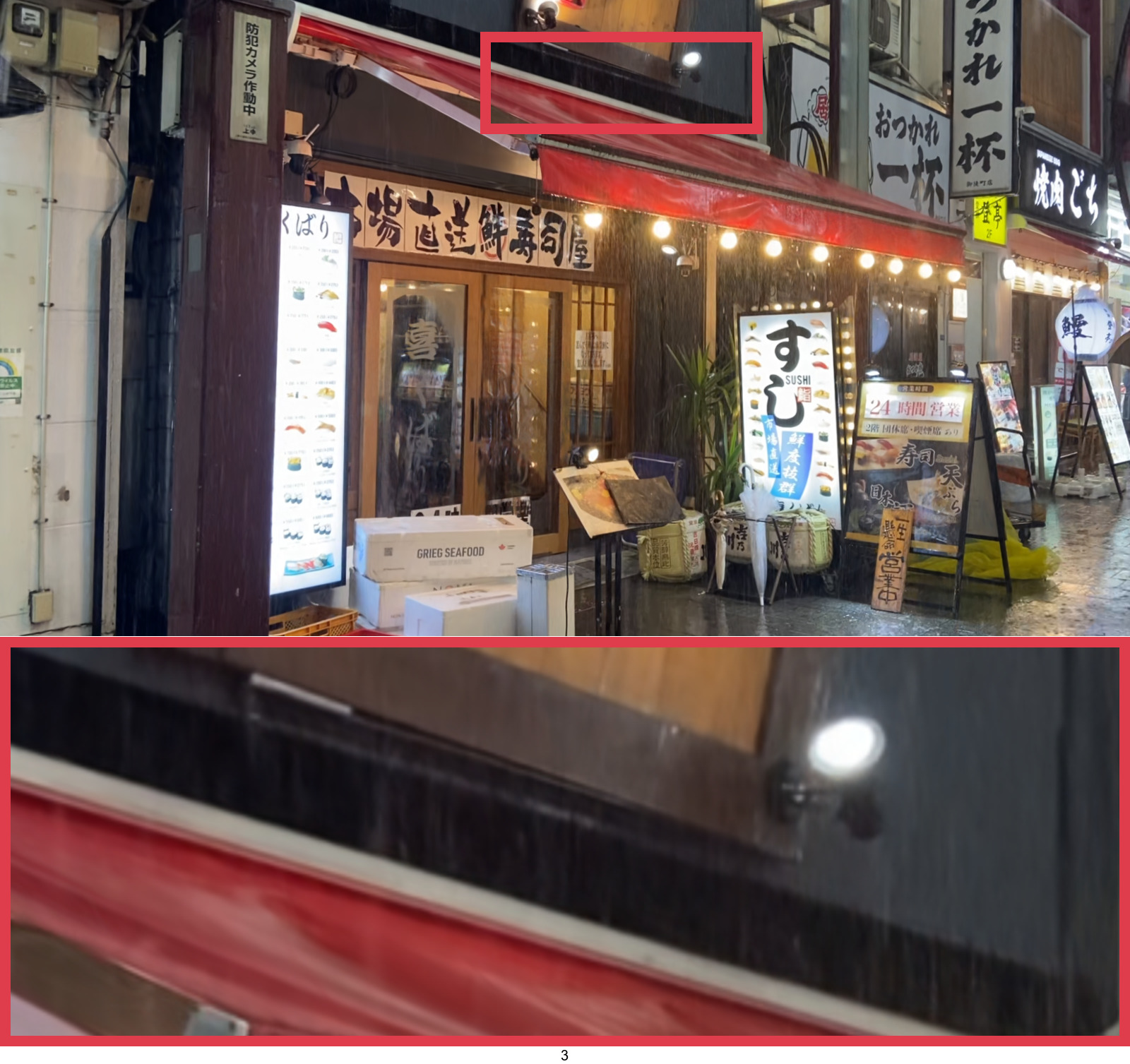} &
    \includegraphics[width=\linewidth]{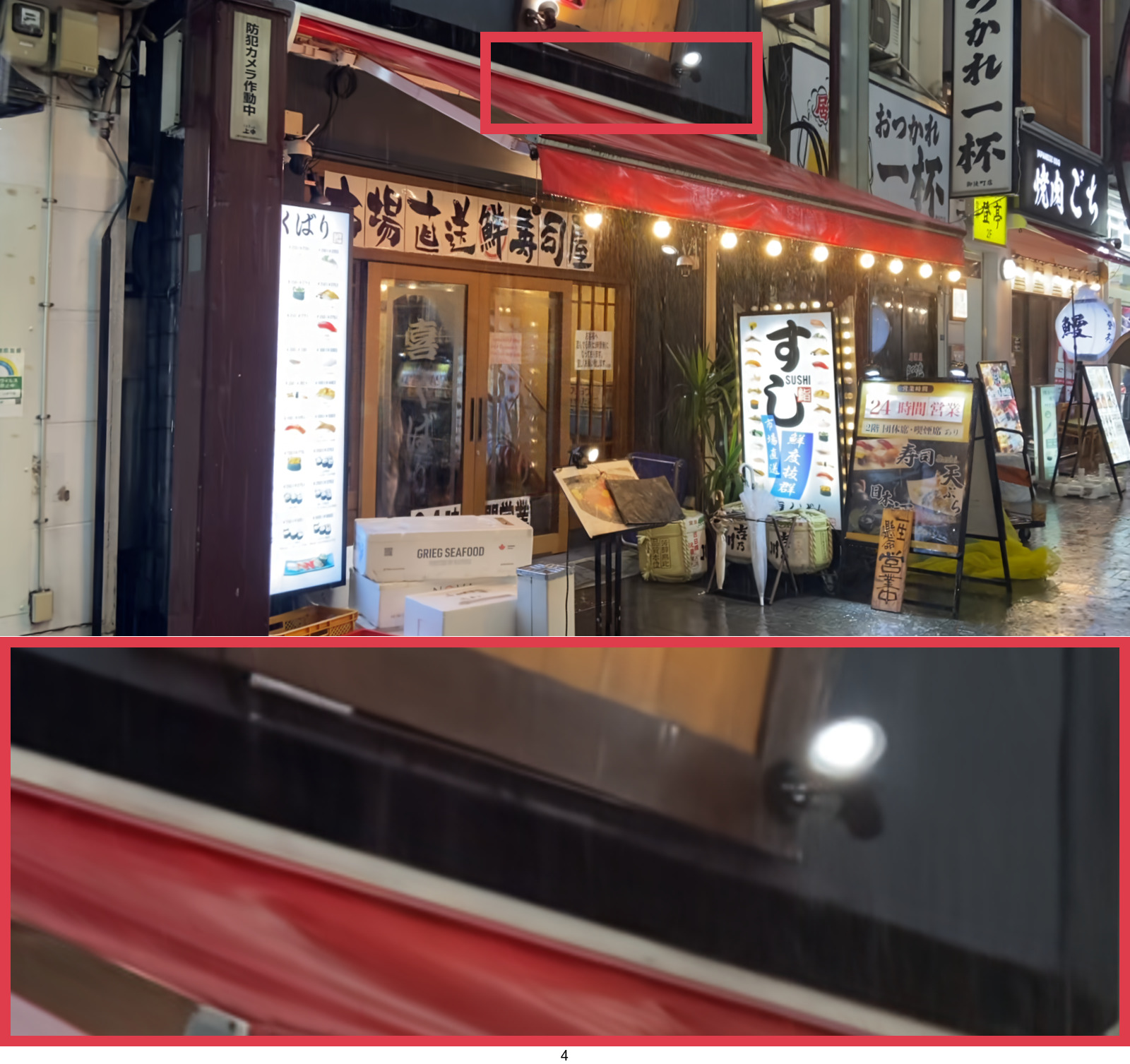}\\[10pt] 
    % 第一行：图像
    Input & 
    DerainNeRF & 
    DRSformer*& 
    RainyScape & 
    Ours \\
    \end{tabular}
    % \vspace{-3mm}
    \caption{The visualization compares the rendering quality of REVR-GSNet and baselines on real-world rainy scenes.}
    % \vspace{-3mm}
    \label{data_show_real_world}
\end{figure}

{\flushleft\textbf{Evaluations on real-world rainy scenes}.} 
To assess the generalization performance of our approach, Figure~\ref{data_show_real_world} showcases a comprehensive evaluation comparing our method with existing baseline methods on real-world scenarios from the HydroViews benchmark.
These selected outdoor scenes exhibit irregular rainfall distributions, presenting challenges for 3D reconstruction tasks.
Experimental results demonstrate that our REVR-GSNet framework outperforms baseline methods in rain removal effectiveness, particularly in addressing the complex atmospheric distortions that conventional methods struggle to process during scene modeling.

\subsection{Ablation Study}

\begin{figure}[!t]
  \centering
  % \vspace{1mm}
  \includegraphics[width=\linewidth]{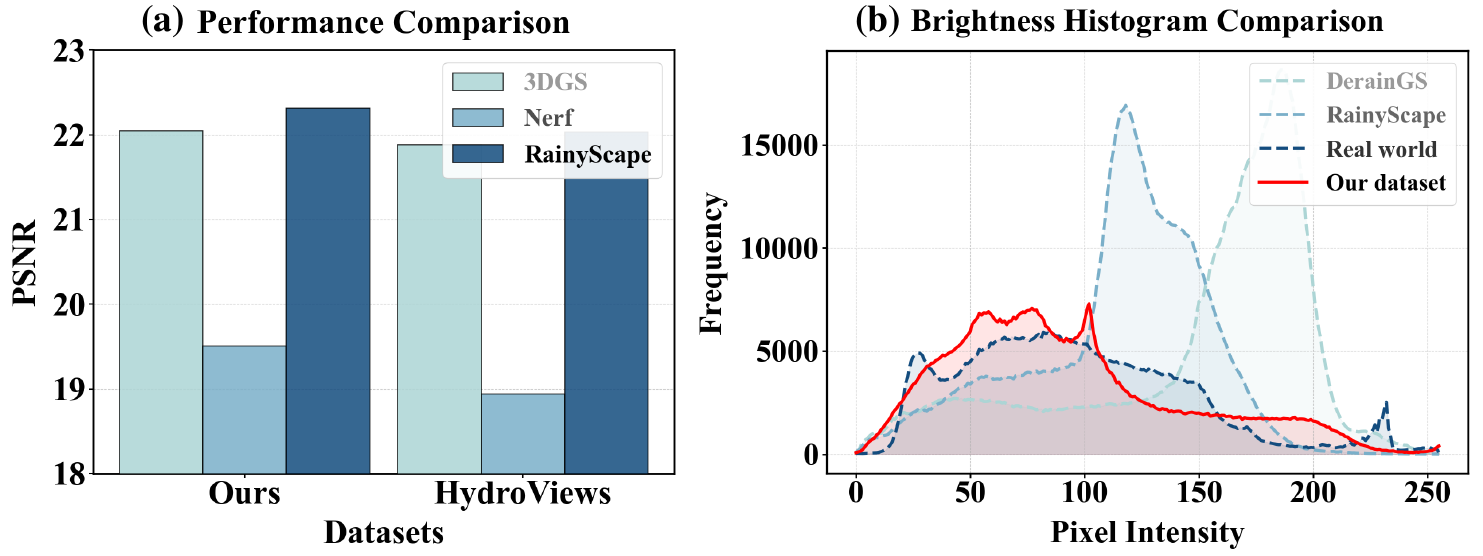}
  \vspace{-6mm}
  \caption{(a) Comparison of scene reconstruction performance of different methods on our dataset and HydroViews dataset. (b) Comparison of brightness histograms between different datasets and real-world rainy images.}
  \label{Fig9}
  \vspace{-4mm}
\end{figure}

{\flushleft\textbf{Effectiveness of our OmniRain3D dataset}.}
To demonstrate the superior quality of our dataset, we evaluate three methods (3DGS, NeRF, RainyScape) on both our dataset and the HydroViews dataset. To ensure fairness, all methods use their default parameter settings, and the datasets share identical background scenes. As shown in Figure~\ref{Fig9}(a), all three methods achieve better reconstruction results on our dataset compared to those on HydroViews.

{\flushleft\textbf{Effectiveness of recursive brightness enhancement}.}
Real-world rainy scences inherently exhibit significant low-brightness characteristics. 
However, existing benchmark datasets ({\itshape e.g.}, RainyScape) predominantly feature brightness patterns that deviate from authentic low-brightness scenarios. 
To quantitatively validate the brightness authenticity of our dataset, we analyze pixel intensity distributions using brightness histograms. 
As shown in Figure~\ref{Fig9}(b), the brightness distribution of our dataset aligns more closely with real-world pixel brightness profiles, confirming its superior authenticity in modeling real-world brightness characteristics.

{\flushleft\textbf{Effectiveness of network components}.}
To validate the effectiveness of our network components, we conduct ablation studies in Table~\ref{tab_ablation}. To ensure fairness, we report the average quantitative results after 30,000 iterations. The results demonstrate that the model with all components achieves the best performance by jointly addressing brightness variations and rain streaks in 3D reconstruction.

%%%%%%%%%%%%%%%%%%%%%%%%%%%%%%%%%%%%%%%%%%%%%%%%%%%%%%%%%%%%%

%%%%%%%%%%%%%%%%%%%%%%%%%%%%%%%%%%%%%%%%%%%%%%%%%%%%%%
\begin{table}[!t]
\footnotesize
\vspace{-2mm}
\resizebox{\linewidth}{!}{
    \begin{tabular}{c|ccc|cc}
        \toprule % 顶部线
        \multirow{2}{*}{Methods}& \multicolumn{3}{c}{Components} & 
        \multicolumn{2}{|c}{Metrics}  \\
          \cmidrule(r){2-4}\cmidrule(lr){5-6}
        & GPO & RBE &  GRE
        & PSNR$\uparrow$ & SSIM$\uparrow$ \\
        \midrule % 中部线
        \multicolumn{1}{l|}{$\text{Ours}_{w/ GPO \& w/o RBE \& w/o GRE}$} &$\checkmark$  && & 19.03 &0.514
        \\
        \multicolumn{1}{l|}{$\text{Ours}_{w/ GPO \& w/ RBE \& w/o GRE}$} &$\checkmark$ &$\checkmark$ & &22.71 &0.615
        \\
        \multicolumn{1}{l|}{$\text{Ours}_{w/ GPO \& w/o RBE \& w/ GRE}$} &$\checkmark$ & & $\checkmark$ &21.64 &0.535 
        \\
        \multicolumn{1}{l|}{Ours} &$\checkmark$ &$\checkmark$ &$\checkmark$ & \textbf{23.88} & \textbf{0.687}
        \\
        \bottomrule
    \end{tabular}
}
% \vspace{-2mm}
\caption{Ablation study of our method variants on HydroViews dataset. 
GPO, RBE, and GRE denote Gaussian Primitives Optimization, Recursive Brightness Enhancement, and GS-guided Rain Elimination, respectively.}
\label{tab_ablation}
\vspace{-4mm}
\end{table}  
%----------------------------------------------------------------------------

\subsection{Discussions with HydroViews dataset}

To better solve the rainy 3D scene reconstruction problem, we propose a new 3D rain model and contribute a high-quality dataset called OmniRain3D.
We note that HydroViews introduces a motion blur method for synthesizing rainy images suitable for 3D scenes.
However, their method has inherent limitations that may not accurately reflect real-world rainy scenarios.
First, the method fails to maintain multi-view consistency.
This is because authentic rain streak should exhibit geometrically corresponding projections across different viewpoints, while a spatial coordination notably absent in HydroViews's scenes.
To address this, we implement perspective-synchronized rendering through physically accurate 3D raindrop modeling in Blender, where rain streak generation and camera parameter alignment with background images ensure cross-view consistency. 
Second, the conventional linear superposition model inadequately represents environmental brightness attenuation, neglecting the nonlinear inverse relationship between rainfall density and scene brightness.
Our method introduces a density-aware brightness attenuation model that maps rainfall characteristics to brightness parameters, enhancing the realism of synthesized rainy images.

\section{Conclusion}
In this paper, we propose a dataset that incorporates perspective heterogeneity and brightness dynamicity to enable more faithful simulation of rain degradation in 3D scenes. 
Based on this dataset, we develop REVR-GSNet, a reconstruction framework that integrates recursive brightness enhancement, Gaussian primitives optimization, and GS-guided rain elimination into a closed-loop architecture. 
This unified approach enables high-quality reconstruction of clean 3D scenes from rain-degraded inputs in an end-to-end manner.
Extensive experiments show that our method effectively addresses brightness variation and rain artifacts in 3D reconstruction.

\bibliography{aaai2026}

@article{tremblay2021rain,
  title={Rain rendering for evaluating and improving robustness to bad weather},
  author={Tremblay, Maxime and Halder, Shirsendu Sukanta and De Charette, Raoul and Lalonde, Jean-Fran{\c{c}}ois},
  journal={IJCV},
  year={2021},
  publisher={Springer}
}

@article{changbo2008real,
  title={Real-time modeling and rendering of raining scenes},
  author={Changbo, Wang and Wang, Zhangye and Zhang, Xin and Huang, Lei and Yang, Zhiliang and Peng, Qunsheng},
  journal={TVC},
  year={2008},
  publisher={Springer}
}

@article{farber1978geometric,
  title={Geometric transformations of pictured space},
  author={Farber, James and Rosinski, Richard R},
  journal={Perception},
  year={1978},
  publisher={SAGE Publications Sage UK: London, England}
}

@inproceedings{choi2025exploiting,
  title={Exploiting Deblurring Networks for Radiance Fields},
  author={Choi, Haeyun and Yang, Heemin and Han, Janghyeok and Cho, Sunghyun},
  booktitle={CVPR},
  year={2025}
}

@article{swinehart1962beer,
  title={The beer-lambert law},
  author={Swinehart, Donald F},
  journal={JCE},
  year={1962},
  publisher={ACS Publications}
}

@article{nia2025exploring,
  title={Exploring the impact of rainfall intensity on the attenuation-rainfall relationship},
  author={Nia, Saeid Esmaeil and Shokri, Ali},
  journal={JQSRT},
  year={2025},
  publisher={Elsevier}
}

@article{chen2023towards,
  title={Towards unified deep image deraining: A survey and a new benchmark. arXiv},
  author={Chen, X and Pan, J and Dong, J and Tang, J},
  journal={arXiv preprint arXiv:2310.03535},
  year={2023}
}

@article{bossu2011rain,
  title={Rain or snow detection in image sequences through use of a histogram of orientation of streaks},
  author={Bossu, J{\'e}r{\'e}mie and Hautiere, Nicolas and Tarel, Jean-Philippe},
  journal={IJCV},
  year={2011},
  publisher={Springer}
}

@inproceedings{brewer2008using,
  title={Using the shape characteristics of rain to identify and remove rain from video},
  author={Brewer, Nathan and Liu, Nianjun},
  booktitle={S+SSPR},
  year={2008},
}

@inproceedings{wei2017should,
  title={Should we encode rain streaks in video as deterministic or stochastic?},
  author={Wei, Wei and Yi, Lixuan and Xie, Qi and Zhao, Qian and Meng, Deyu and Xu, Zongben},
  booktitle={ICCV},
  year={2017}
}

@article{barnum2010analysis,
  title={Analysis of rain and snow in frequency space},
  author={Barnum, Peter C and Narasimhan, Srinivasa and Kanade, Takeo},
  journal={IJCV},
  year={2010},
  publisher={Springer}
}

@article{chen2013rain,
  title={A rain pixel recovery algorithm for videos with highly dynamic scenes},
  author={Chen, Jie and Chau, Lap-Pui},
  journal={IEEE TIP},
  year={2013},
  publisher={IEEE}
}

@inproceedings{chen2013generalized,
  title={A generalized low-rank appearance model for spatio-temporally correlated rain streaks},
  author={Chen, Yi-Lei and Hsu, Chiou-Ting},
  booktitle={Proceedings of the IEEE International Conference on Computer Vision},
  year={2013}
}

@article{liu2009pixel,
  title={Pixel Based Temporal Analysis Using Chromatic Property for Removing Rain from Videos.},
  author={Liu, Peng and Xu, Jing and Liu, Jiafeng and Tang, Xianglong},
  journal={Comput. Inf. Sci.},
  year={2009}
}

@inproceedings{zhang2006rain,
  title={Rain removal in video by combining temporal and chromatic properties},
  author={Zhang, Xiaopeng and Li, Hao and Qi, Yingyi and Leow, Wee Kheng and Ng, Teck Khim},
  booktitle={ICME},
  year={2006},
}

@article{garg2007vision,
  title={Vision and rain},
  author={Garg, Kshitiz and Nayar, Shree K},
  journal={IJCV},
  year={2007},
  publisher={Springer}
}

@inproceedings{garg2004detection,
  title={Detection and removal of rain from videos},
  author={Garg, Kshitiz and Nayar, Shree K},
  booktitle={CVPR},
  year={2004},
}

@article{wang2023multi,
  title={Multi-scale fusion and decomposition network for single image deraining},
  author={Wang, Qiong and Jiang, Kui and Wang, Zheng and Ren, Wenqi and Zhang, Jianhui and Lin, Chia-Wen},
  journal={IEEE TIP},
  year={2023},
  publisher={IEEE}
}

@inproceedings{ran2024rainmer,
  title={Rainmer: Learning Multi-view Representations for Comprehensive Image Deraining and Beyond},
  author={Ran, Wu and Ma, Peirong and He, Zhiquan and Lu, Hong},
  booktitle={ACM MM},
  year={2024}
}

@inproceedings{jiang2023dawn,
  title={Dawn: Direction-aware attention wavelet network for image deraining},
  author={Jiang, Kui and Liu, Wenxuan and Wang, Zheng and Zhong, Xian and Jiang, Junjun and Lin, Chia-Wen},
  booktitle={ACM MM},
  year={2023}
}

@inproceedings{zhang2018density,
  title={Density-aware single image de-raining using a multi-stream dense network},
  author={Zhang, He and Patel, Vishal M},
  booktitle={CVPR},
  year={2018}
}

@article{zhang2019image,
  title={Image de-raining using a conditional generative adversarial network},
  author={Zhang, He and Sindagi, Vishwanath and Patel, Vishal M},
  journal={IEEE TCSVT},
  year={2019},
  publisher={IEEE}
}

@article{zhang2022enhanced,
  title={Enhanced spatio-temporal interaction learning for video deraining: faster and better},
  author={Zhang, Kaihao and Li, Dongxu and Luo, Wenhan and Ren, Wenqi and Liu, Wei},
  journal={IEEE TPAMI},
  year={2022},
  publisher={IEEE}
}

@inproceedings{zhang2018unreasonable,
  title={The unreasonable effectiveness of deep features as a perceptual metric},
  author={Zhang, Richard and Isola, Phillip and Efros, Alexei A and Shechtman, Eli and Wang, Oliver},
  booktitle={CVPR},
  year={2018}
}

@article{wang2004image,
  title={Image quality assessment: from error visibility to structural similarity},
  author={Wang, Zhou and Bovik, Alan C and Sheikh, Hamid R and Simoncelli, Eero P},
  journal={IEEE TIP},
  year={2004},
  publisher={IEEE}
}

@article{huynh2008scope,
  title={Scope of validity of PSNR in image/video quality assessment},
  author={Huynh-Thu, Quan and Ghanbari, Mohammed},
  journal={Electronics Letters},
  year={2008},
  publisher={IET}
}

@inproceedings{schoenberger2016sfm,
    author={Sch\"{o}nberger, Johannes Lutz and Frahm, Jan-Michael},
    title={Structure-from-Motion Revisited},
    booktitle={CVPR},
    year={2016},
}

@inproceedings{wang2020dcsfn,
  title={Dcsfn: Deep cross-scale fusion network for single image rain removal},
  author={Wang, Cong and Xing, Xiaoying and Wu, Yutong and Su, Zhixun and Chen, Junyang},
  booktitle={ACM MM},
  year={2020}
}

@inproceedings{wang2024cascaded,
  title={Cascaded Adversarial Attack: Simultaneously Fooling Rain Removal and Semantic Segmentation Networks},
  author={Wang, Zhiwen and Wu, Yuhui and Wang, Zheng and Wei, Jiwei and Li, Tianyu and Wang, Guoqing and Yang, Yang and Shen, Hengtao},
  booktitle={ACM MM},
  year={2024}
}

@inproceedings{li2022single,
  title={Single image deraining network with rain embedding consistency and layered LSTM},
  author={Li, Yizhou and Monno, Yusuke and Okutomi, Masatoshi},
  booktitle={WACV},
  year={2022}
}

@inproceedings{guo2020zero,
  title={Zero-reference deep curve estimation for low-light image enhancement},
  author={Guo, Chunle and Li, Chongyi and Guo, Jichang and Loy, Chen Change and Hou, Junhui and Kwong, Sam and Cong, Runmin},
  booktitle={CVPR},
  year={2020}
}

@book{hess2013blender,
  title={Blender foundations: The essential guide to learning blender 2.5},
  author={Hess, Roland},
  year={2013},
  publisher={Routledge}
}

@inproceedings{liu2018erase,
  title={Erase or fill? deep joint recurrent rain removal and reconstruction in videos},
  author={Liu, Jiaying and Yang, Wenhan and Yang, Shuai and Guo, Zongming},
  booktitle={CVPR},
  year={2018}
}

@article{shi2024nitedr,
  title={NiteDR: Nighttime Image De-Raining with Cross-View Sensor Cooperative Learning for Dynamic Driving Scenes},
  author={Shi, Cidan and Fang, Lihuang and Wu, Han and Xian, Xiaoyu and Shi, Yukai and Lin, Liang},
  journal={IEEE TMM},
  year={2024},
  publisher={IEEE}
}

@inproceedings{jiang2020multi,
  title={Multi-scale progressive fusion network for single image deraining},
  author={Jiang, Kui and Wang, Zhongyuan and Yi, Peng and Chen, Chen and Huang, Baojin and Luo, Yimin and Ma, Jiayi and Jiang, Junjun},
  booktitle={CVPR},
  year={2020}
}

@inproceedings{wang2024progressive,
  title={Progressive Local and Non-Local Interactive Networks with Deeply Discriminative Training for Image Deraining},
  author={Wang, Cong and Wang, Liyan and Mu, Jie and Yu, Chengjin and Wang, Wei},
  booktitle={ACM MM},
  year={2024}
}

@article{song2024exploringanefficient,
  title={Exploring an efficient frequency-guidance transformer for single image deraining},
  author={Song, Tianyu and Fan, Shumin and Jin, Jiyu and Jin, Guiyue and Fan, Lei},
  journal={SIVP},
  year={2024},
  publisher={Springer}
}

@article{song2024exploring,
  title={Exploring a context-gated network for effective image deraining},
  author={Song, Tianyu and Li, Pengpeng and Fan, Shumin and Jin, Jiyu and Jin, Guiyue and Fan, Lei},
  journal={JVCIR},
  publisher={Elsevier}
}

@inproceedings{zamir2021multi,
  title={Multi-stage progressive image restoration},
  author={Zamir, Syed Waqas and Arora, Aditya and Khan, Salman and Hayat, Munawar and Khan, Fahad Shahbaz and Yang, Ming-Hsuan and Shao, Ling},
  booktitle={CVPR},
  year={2021}
}

@inproceedings{wang2019spatial,
  title={Spatial attentive single-image deraining with a high quality real rain dataset},
  author={Wang, Tianyu and Yang, Xin and Xu, Ke and Chen, Shaozhe and Zhang, Qiang and Lau, Rynson WH},
  booktitle={CVPR},
  year={2019}
}

@article{fu2019lightweight,
  title={Lightweight pyramid networks for image deraining},
  author={Fu, Xueyang and Liang, Borong and Huang, Yue and Ding, Xinghao and Paisley, John},
  journal={IEEE TNNLS},
  year={2019},
  publisher={IEEE}
}

@article{liu2024deraings,
  title={DeRainGS: Gaussian Splatting for Enhanced Scene Reconstruction in Rainy Environments},
  author={Liu, Shuhong and Chen, Xiang and Chen, Hongming and Xu, Quanfeng and Li, Mingrui},
  journal={arXiv preprint arXiv:2408.11540},
  year={2024}
}

@inproceedings{lyu2024rainyscape,
  title={Rainyscape: Unsupervised rainy scene reconstruction using decoupled neural rendering},
  author={Lyu, Xianqiang and Liu, Hui and Hou, Junhui},
  booktitle={ACM MM},
  year={2024}
}

@article{shao2021selective,
  title={Selective generative adversarial network for raindrop removal from a single image},
  author={Shao, Mingwen and Li, Le and Wang, Hong and Meng, Deyu},
  journal={Neurocomputing},
  year={2021},
  publisher={Elsevier}
}

@article{li2024derainnerf,
  title={DerainNeRF: 3D Scene Estimation with Adhesive Waterdrop Removal},
  author={Li, Yunhao and Wu, Jing and Zhao, Lingzhe and Liu, Peidong},
  journal={arXiv preprint arXiv:2403.20013},
  year={2024}
}

@inproceedings{chen2023learning,
  title={Learning a sparse transformer network for effective image deraining},
  author={Chen, Xiang and Li, Hao and Li, Mingqiang and Pan, Jinshan},
  booktitle={CVPR},
  year={2023}
}

@inproceedings{chen2024bidirectional,
  title={Bidirectional multi-scale implicit neural representations for image deraining},
  author={Chen, Xiang and Pan, Jinshan and Dong, Jiangxin},
  booktitle={CVPR},
  year={2024}
}

@article{kang2011automatic,
  title={Automatic single-image-based rain streaks removal via image decomposition},
  author={Kang, Li-Wei and Lin, Chia-Wen and Fu, Yu-Hsiang},
  journal={IEEE TIP},
  year={2011},
  publisher={IEEE}
}

@inproceedings{li2016rain,
  title={Rain streak removal using layer priors},
  author={Li, Yu and Tan, Robby T and Guo, Xiaojie and Lu, Jiangbo and Brown, Michael S},
  booktitle={CVPR},
  year={2016}
}

@inproceedings{lu2025dual,
  title={Dual-View Interaction-Aware Lane Change Prediction for Autonomous Driving},
  author={Lu, Yuhuan and Zhang, Zhen and Bai, Rufan and Liu, Han and Wang, Wei},
  booktitle={AAAI},
  year={2025}
}

@inproceedings{hu2024daldet,
  title={DALDet: Depth-aware learning based object detection for autonomous driving},
  author={Hu, Ke and Cao, Tongbo and Li, Yuan and Chen, Song and Kang, Yi},
  booktitle={AAAI},
  year={2024}
}

@inproceedings{liang2025diffusion,
  title={Diffusion Models for Robotics},
  author={Liang, Jessica E},
  booktitle={AAAI},
  year={2025}
}

\end{document}